\documentclass[tablecaption=bottom,wcp]{jmlr} 

\usepackage{graphicx}
\usepackage[utf8]{inputenc} 
\usepackage[T1]{fontenc}    
\usepackage{mathtools}
\usepackage{bm}
\usepackage{breakcites}
\usepackage{booktabs}
\usepackage{amsfonts}
\usepackage{nicefrac}
\usepackage{microtype}
\usepackage{xcolor}
\usepackage{xspace}
\usepackage{multirow}
\usepackage{multicol}
\usepackage{pgfplots}
\usepackage{caption}
\usepackage{enumitem}
\usepackage{algorithm}
\usepackage{algorithmic}
\usepackage{wrapfig}

\def\eqref#1{equation~\ref{#1}}

\def\1{\bm{1}}

\def\vmu{{\bm{\mu}}}
\def\vtheta{{\bm{\theta}}}

\def\vb{{\bm{b}}}

\def\vf{{\bm{f}}}
\def\vg{{\bm{g}}}

\def\vp{{\bm{p}}}

\def\vs{{\bm{s}}}

\def\vu{{\bm{u}}}
\def\vv{{\bm{v}}}
\def\vw{{\bm{w}}}
\def\vx{{\bm{x}}}
\def\vy{{\bm{y}}}
\def\vz{{\bm{z}}}

\def\mI{{\bm{I}}}

\def\mS{{\bm{S}}}

\def\mW{{\bm{W}}}
\def\mX{{\bm{X}}}

\def\mSigma{{\bm{\Sigma}}}

\DeclareMathAlphabet{\mathsfit}{\encodingdefault}{\sfdefault}{m}{sl}
\SetMathAlphabet{\mathsfit}{bold}{\encodingdefault}{\sfdefault}{bx}{n}

\def\gD{{\mathcal{D}}}

\def\gL{{\mathcal{L}}}

\def\gN{{\mathcal{N}}}

\def\gX{{\mathcal{X}}}
\def\gY{{\mathcal{Y}}}

\newcommand{\E}{\mathbb{E}}

\newcommand{\KL}{D_{\mathrm{KL}}}

\jmlrproceedings{AABI 2023}{5th Symposium on Advances in Approximate Bayesian Inference, 2023}

\title[Function-Space Regularization for Deep Bayesian Classification]{Function-Space Regularization for Deep Bayesian Classification}

\author{
\Name{Jihao Andreas Lin\nametag{\thanks{Equal contribution. $\dagger$ Work done as a M.Sc. student at TU Darmstadt.}}}$^\dagger$ \Email{jal232@cam.ac.uk} \\ \addr University of Cambridge \quad Max Planck Institute for Intelligent Systems \\
\Name{Joe Watson}{\normalfont\textsuperscript{*}} \Email{joe.watson@tu-darmstadt.de} \\
\Name{Pascal Klink} \Email{pascal.klink@tu-darmstadt.de} \\
\Name{Jan Peters} \Email{jan.peters@tu-darmstadt.de} \\
\addr Technical University of Darmstadt
}

\begin{document}

\maketitle

\begin{abstract}
Bayesian deep learning approaches assume model parameters to be latent random variables and infer posterior distributions to quantify uncertainty, increase safety and trust, and prevent overconfident and unpredictable behavior.
However, weight-space priors are model-specific, can be difficult to interpret and are hard to specify.
Instead, we apply a Dirichlet prior in predictive space and perform approximate function-space variational inference.
To this end, we interpret conventional categorical predictions from stochastic neural network classifiers as samples from an implicit Dirichlet distribution.
By adapting the inference, the same function-space prior can be combined with different models without affecting model architecture or size.
We illustrate the flexibility and efficacy of such a prior with toy experiments and demonstrate scalability, improved uncertainty quantification and adversarial robustness with large-scale image classification experiments.
\end{abstract}


\section{Introduction}
\label{sec:introduction}
Deep learning has enabled powerful classification models capable of working with complex data modalities and scaling to large data sets \citep{NIPS2012_c399862d,GoodBengCour16}.
The aim of \emph{Bayesian} neural networks (BNNs) is to provide these complex models with priors for regularization, generalization and uncertainty quantification (UQ) useful in  prediction tasks \citep{Gal2016Uncertainty,wilson2020bayesian,fortuin2021priors,ABDAR2021}.
Predictive uncertainty is crucial for machine learning systems in real-world settings, as it provides a degree of safety \citep{ijcai2017-661}, trust \citep{Lim_Lee_Hsu_Wong_2019}, sample efficiency \citep{deisenroth2011pilco,pmlr-v70-gal17a} and human-in-the-loop cooperation \citep{oatml2019bdlb}.
In this work, we leverage function-space variational inference
\footnote{We use \emph{function-space} VI instead of \emph{functional} VI to avoid the overloaded term with different connotations.}
\citep{Sun19} (fVI) to implement regularization for classification tasks.
Function-space priors can explicitly affect the predictive distribution and do not depend on the particular model parameterization, whereas weight-space priors are implicit and model-specific.
Given any stochastic neural network capable of producing multiple predictions, such as Monte Carlo dropout \citep{Gal16} or deep ensembles \citep{NIPS2017_7219}, we estimate a Dirichlet predictive distribution from several categorical outputs via maximum likelihood.
This approach retains the same mean prediction of conventional deep learning classifiers, while also capturing the information contained in the variance of the outputs.
The Dirichlet predictive distribution can then be used to specify a function-space prior to regularize classification.
Various prior work which uses the Dirichlet distribution and function-space regularization \citep{malinin2018predictive,Malinin2020Ensemble,joo2020being,sensoy2018evidential, Sensoy_2021_WACV} can be viewed as a special case of fVI.
We demonstrate that our method improves uncertainty quantification and adversarial robustness across a range of popular models and datasets for both small- and large-scale inference. 
\begin{figure}[t]
    \centering
    \begin{tabular}{l c c c c}
        & Weight-space prior & fVI, uniform prior & fVI, GP prior & fVI, RF prior \\
        \rotatebox[origin=l]{90}{\hspace{1.75em}MAP} &
        \includegraphics[width=.20\textwidth]{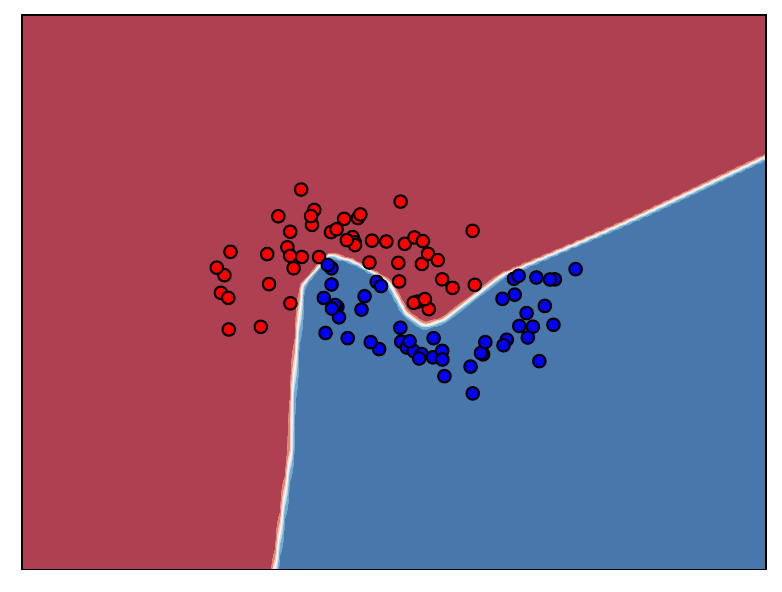} &
        \includegraphics[width=.20\textwidth]{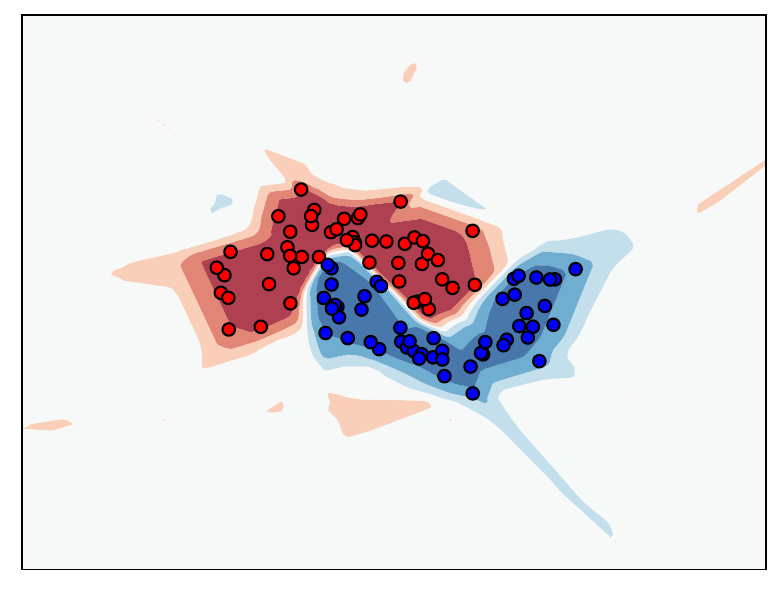} &
        \includegraphics[width=.20\textwidth]{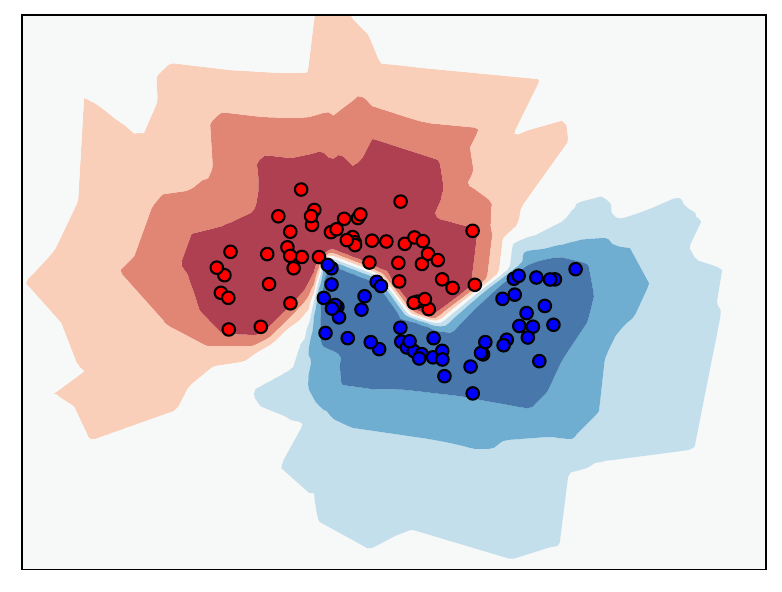} &
        \includegraphics[width=.20\textwidth]{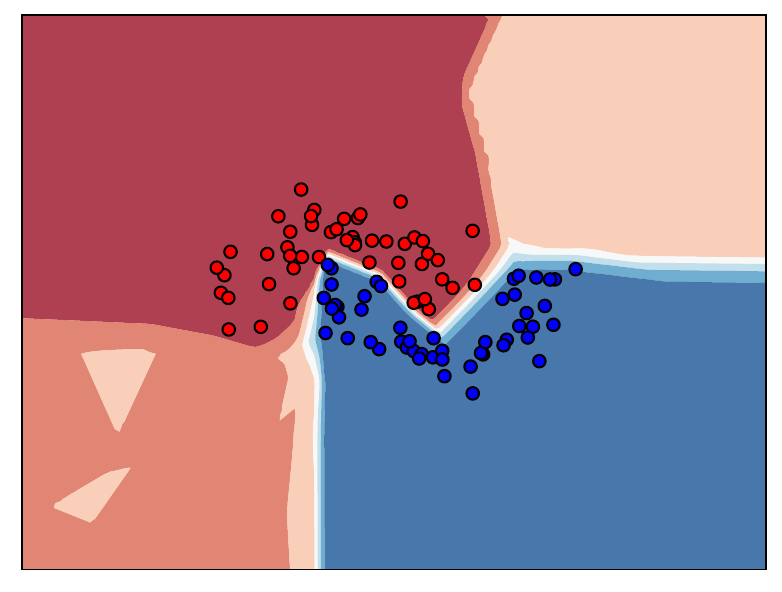} \\
        \rotatebox[origin=l]{90}{\hspace{0.3em}MC Dropout} &
        \includegraphics[width=.20\textwidth]{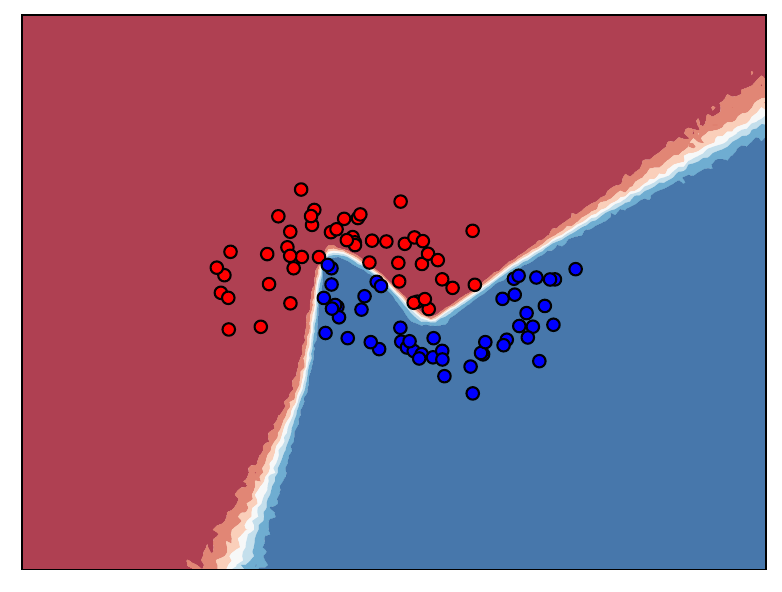} &
        \includegraphics[width=.20\textwidth]{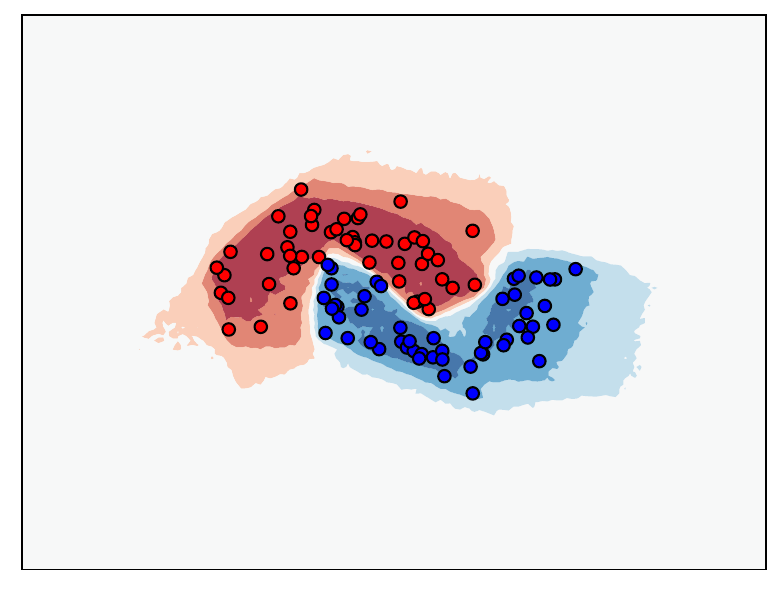} &
        \includegraphics[width=.20\textwidth]{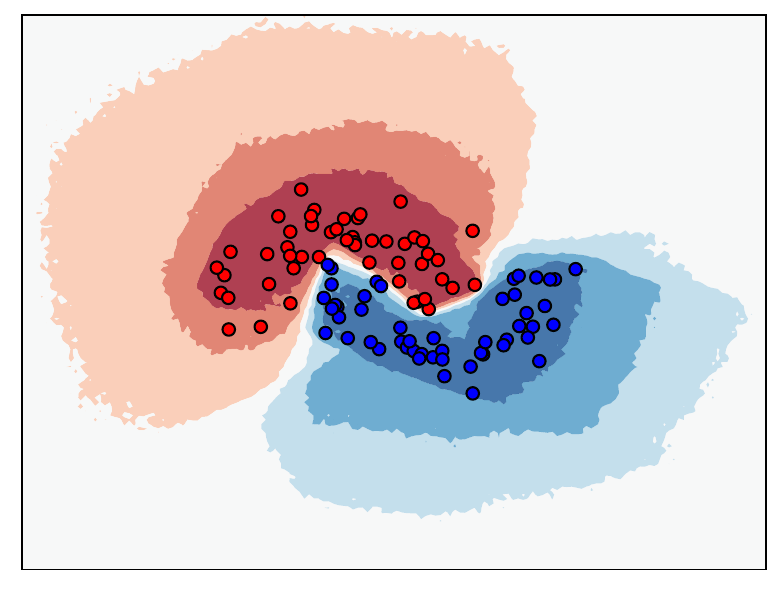} &
        \includegraphics[width=.20\textwidth]{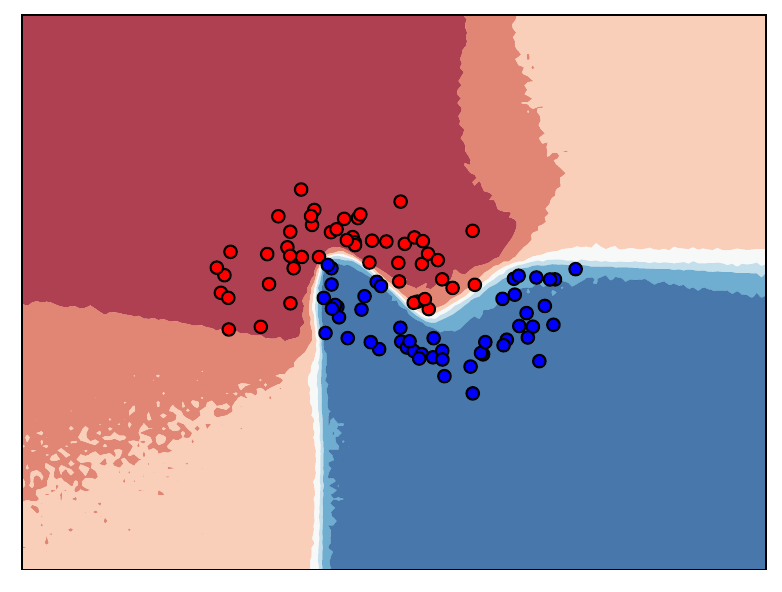} \\
        \rotatebox[origin=l]{90}{\hspace{0.6em}Ensemble} &
        \includegraphics[width=.20\textwidth]{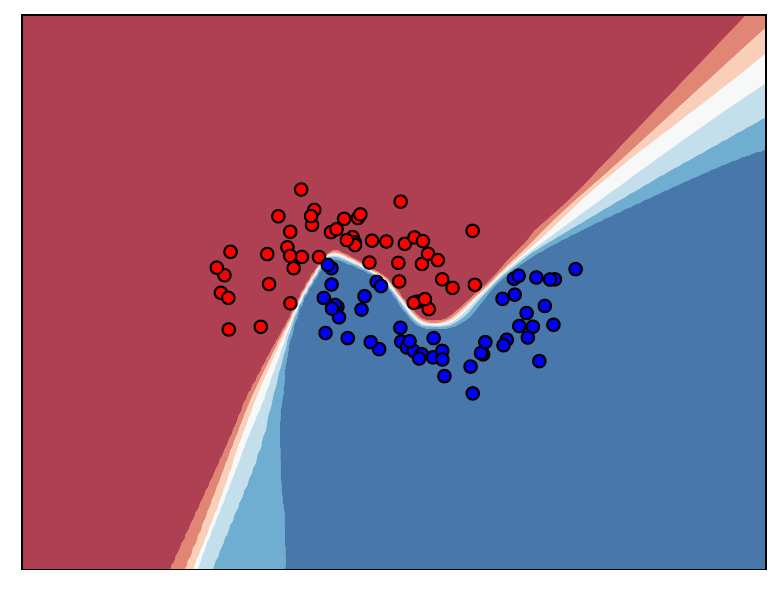} &
        \includegraphics[width=.20\textwidth]{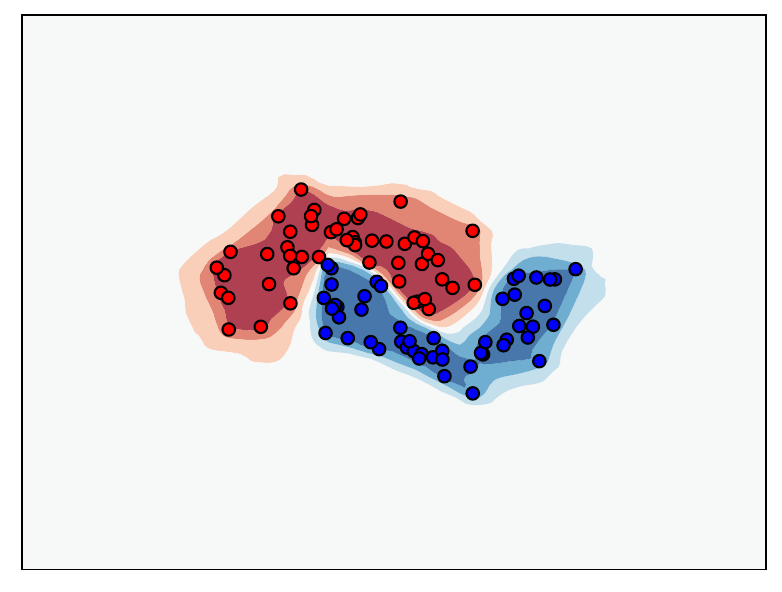} &
        \includegraphics[width=.20\textwidth]{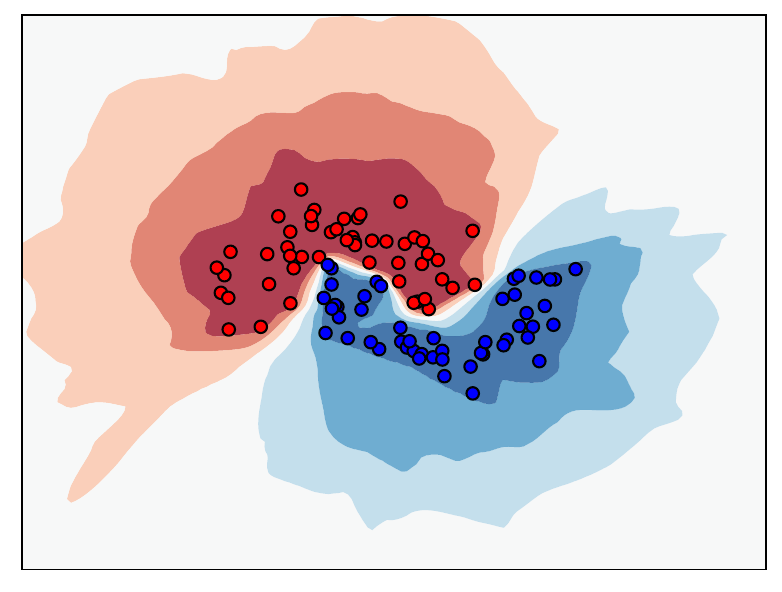} &
        \includegraphics[width=.20\textwidth]{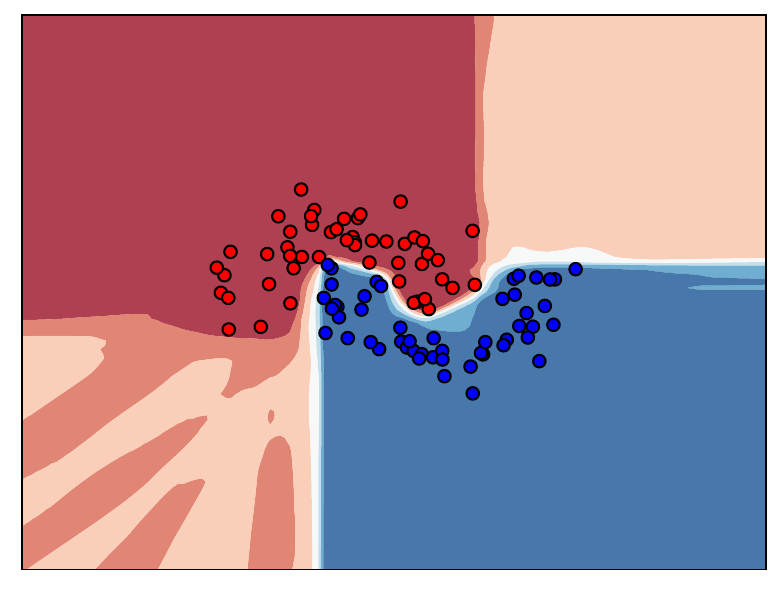}
    \end{tabular}
    \caption{Toy classification problem using the Two Moons dataset.
    The leftmost column shows the undesirable overconfidence of standard weight-space priors outside of the data distribution.
    The second column illustrates how our function-space inference approach combined with a uniform Dirichlet prior
    trapolation behavior and instead
    adequately increases model uncertainty outside of the observed data.
    The third and fourth column demonstrate that our approach can also be combined with
    priors based on (trained) Gaussian processes (GP) or random forests (RF). 
    }
    \label{fig:two_moons}
\end{figure}
\section{Dirichlet Function Priors}
\label{sec:dirichlet_function_priors}
%
\label{sec:classification}
Let $\gD = \{(\vx_n, \vy_n)\}_{n=1}^N$ be the training data consisting of $N$ observed pairs of input data $\vx_n \in \gX$ and corresponding $K$-dimensional, one-hot class label vectors $\vy_n \in \gY$.
A neural network $\bm{\phi}$ with weights $\vw$ defines a deterministic function $\vf$ which maps an input $\vx \in \mathcal{X}$ to an element $\vf_{\vx} \in \Delta^{K-1}$, where $\Delta^{K-1}$ denotes the $K{-}1$ simplex.
More precisely, we write $\vf_{\vx} = \bm{\sigma}\left(\bm{\phi}\left(\vx; \vw\right)\right)$ and $\vy \sim \mathrm{Cat}\left(\cdot | \vf_{\vx}\right)$, where $\bm{\sigma}$ is the softmax function.
In conventional maximum likelihood (ML) training, the weights $\vw$ are optimized by maximizing
\begin{align}
     \log \textstyle\prod_{\mathcal{D}} \mathrm{Cat}\left(\vy | \vf_{\vx} \right)
     = \sum_{\mathcal{D}} \sum_{k=1}^K y_k \log {f_{\vx}}_k,
\end{align}
where $\prod_{\mathcal{D}}$ and $\sum_{\mathcal{D}}$ denote $\prod_{(\vx, \vy) \in \mathcal{D}}$ and $\sum_{(\vx, \vy) \in \mathcal{D}}$, and $\bm{\phi}$, $\bm{\sigma}$, and $\vw$ are implicit in $\vf$.

In Bayesian deep learning, $\vw$ becomes a random variable and the goal is to estimate its posterior weight distribution.
In general, exact inference is intractable and various approximations employ different parameterizations.
In this paper, we assume that samples from a weight distribution $p(\vw)$ are available but an explicit density is not.
This makes our method particularly generic and compatible with most BNNs and stochastic models.

\paragraph{Dirichlet Posterior Predictive}
Bayesian neural networks and stochastic deep learning models for classification typically make predictions by first sampling from a weight distribution $p(\vw)$, then predicting a softmax output for each weight sample, and finally averaging those predictions to produce a posterior categorical predictive.

\begin{wrapfigure}{r}{.43\textwidth}
    \centering
    \vspace{-7.5pt}
    \begin{tabular}{c@{\hspace{.3cm}}c@{\hspace{.3cm}}c}
        \includegraphics[width=.12\textwidth]{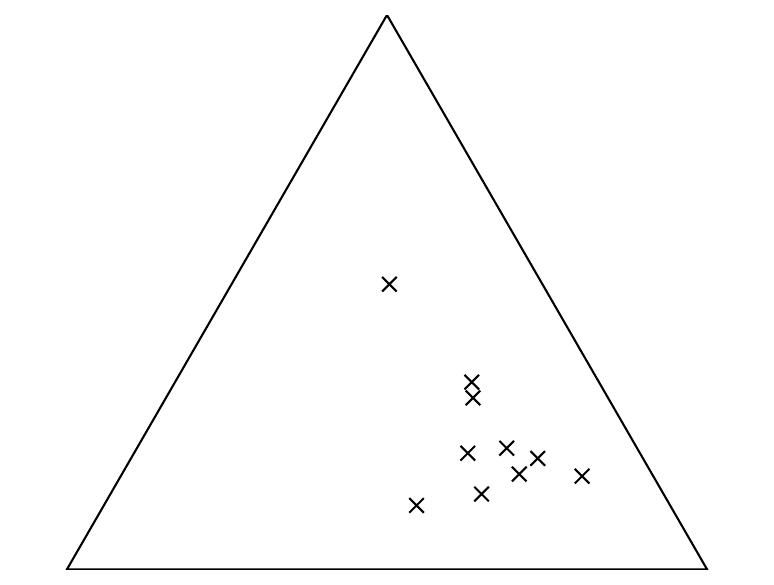} &
        \includegraphics[width=.12\textwidth]{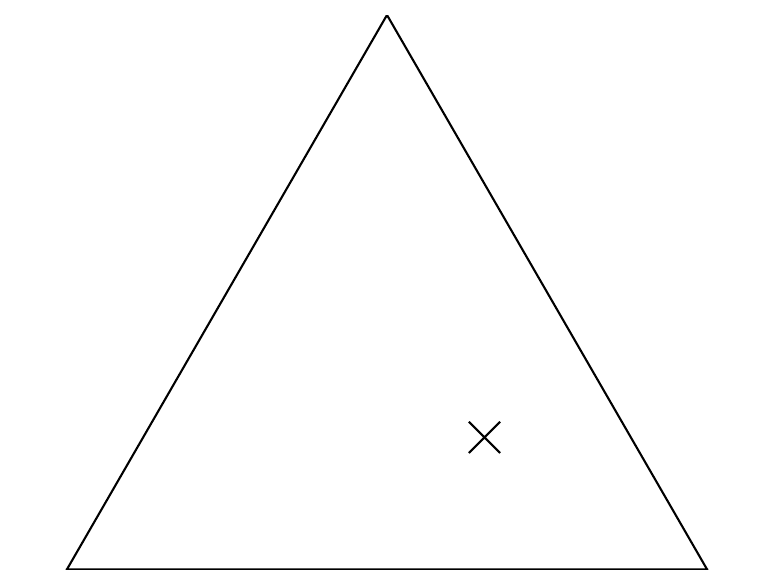} &
        \includegraphics[width=.12\textwidth]{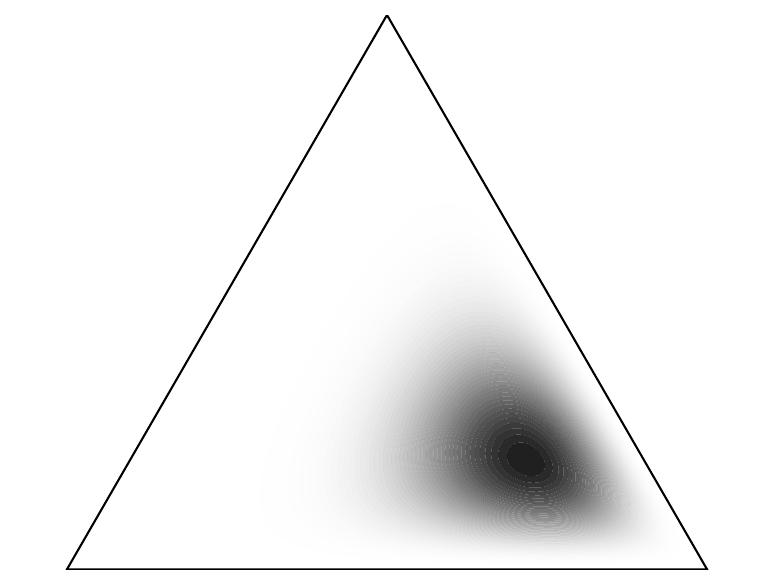}
    \end{tabular}
    \vspace{-8.5pt}
    \caption{Treating model predictions as samples from a simplex (left), the mean reduction (middle) discards information which is present in the variance.
    A Dirichlet distribution (right) fitted to the same samples can capture the uncertainty with its density function.}
    \vspace{-12.5pt}
    \label{fig:dirichlet}
\end{wrapfigure}
However, taking the average \emph{throws away} the epistemic uncertainty of the classifier.
Instead, we interpret categorical predictions as samples from a \emph{Dirichlet} distribution $p(\vf_{\vx})$, which allows us to leverage those samples to estimate a Dirichlet distribution over probability vectors instead of computing the average.
Figure \ref{fig:dirichlet} shows how the Dirichlet density captures the variance of the samples.
We assume that, given any input $\vx$, the model predicts a corresponding Dirichlet distribution $p(\vf_{\vx})$, which is induced by the weight distribution $p(\vw)$.

\paragraph{Implicit Stochastic Processes}
The model's capability to predict a $K$-dimensional Dirichlet distribution $p(\vf_{\vx})$ for any $\vx \in \gX$ implicitly defines a stochastic process whose state space is the $K-1$ simplex $\Delta^{K-1}$ and whose index set is $\gX$ \citep{pmlr-v97-ma19b}.
This stochastic process, despite using the Dirichlet distribution, is \emph{not} a Dirichlet process \citep{Teh2010}.
A Dirichlet process with index set $\gX$ requires that any finite subset $\{\vx_1, \dots, \vx_L\} \subset \gX$ follows a joint $L$-dimensional Dirichlet, whereas our implicit stochastic process defines a $K$-dimensional Dirichlet for each $\vx \in \gX$.
For us, the finite collection $\{\vx_1, \dots, \vx_L\}$ would produce an element from $\Delta^{K-1}$ to the power of $L$ and the whole implicit stochastic process could be rigorously defined as a random variable from $\Delta^{K-1}$ to the power of $\gX$ (see Appendix~\ref{app:vip}).

\paragraph{Function-Space Regularization}
\label{sec:fvi}
To apply regularization in function space, we use the function-space evidence lower bound objective (fELBO) \citep{Sun19},
\begin{align} \label{eq:felbo}
\mathcal{L}(\bm{\theta}) = \E_{\,\vf \sim q}
\left[ \log p(\mathcal{D} | \vf) \right]
- \KL[{q(\vf |\bm{\theta}) \mid\mid p(\vf)}],
\end{align}
which resembles the conventional evidence lower bound objective (ELBO) \citep{jordanvi, hoffman13a}.
To compute the likelihood term, we stay faithful to the backbone model and use $M$ samples to estimate the expected categorical log-likelihood,
\begin{align}
\E_{\vf\sim q}
\left[ \log p(\mathcal{D} | \vf) \right]
&\approx \textstyle\frac{1}{M} \sum_{n,m=1}^{N, M}
\log p\left(\vy_n,\vf^{(m)}_{\vx_n}\right),
\end{align}
which is identical to the likelihood term in conventional ELBO optimization for BNNs.
The novelty of our approach manifests in the KL term, which requires computing a function-space KL divergence (fKL) between stochastic processes.
\citet{Sun19} derived this divergence as the supremum over regular KL divergences evaluated at all possible finite sets $\mX \subset \gX$,
\begin{align} \label{eq:fkl}
\KL[q\mid\mid p]
&=
\textstyle\sup_{\mX \subset \gX, |\mX| < \infty}
\KL[q(\vf_\mX |\bm{\theta})\mid\mid p(\vf_\mX)].
\end{align}
However, the supremum is generally intractable because there are infinitely many possible finite measurement sets.
A tractable approximation \citep{Sun19,bruinsma2021the} replaces the supremum with an expectation,
\begin{align}
\KL[q\mid\mid p]
&\approx \;
\E_{\mS} \;
\KL[q(\vf_\mS |\vtheta)\mid\mid p(\vf_\mS)],
\end{align}
where $\mS \subset \mathcal{X}$ is a randomly sampled, finite measurement set of size $L$,
which contains all the points that the stochastic processes are conditioned on. 
For us,
this is the training data, but we can improve it further by adding unlabeled data (see Section \ref{sec:experiments}).
\begin{figure*}[t]
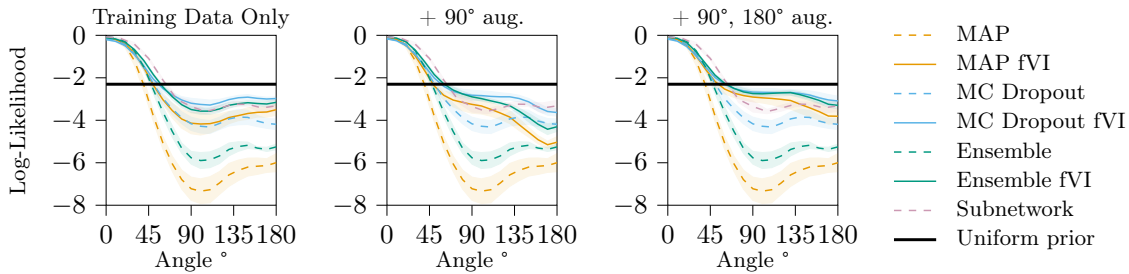

    \centering
    \resizebox{\textwidth}{!}{%
        \input{figures/figure_MNIST_index_set_llh}
        \input{figures/figure_MNIST_index_set_index90_llh}
        \input{figures/figure_MNIST_index_set_index90_180_llh}
        \raisebox{0.3cm}{\begin{tikzpicture}

\newenvironment{customlegend}[1][]{%
    \begingroup
    \csname pgfplots@init@cleared@structures\endcsname
    \pgfplotsset{#1}%
}{%
    \csname pgfplots@createlegend\endcsname
    \endgroup
}%

\def\addlegendimage{\csname pgfplots@addlegendimage\endcsname}

\definecolor{color0}{rgb}{0.9,0.6,0}
\definecolor{color1}{rgb}{0.35,0.7,0.9}
\definecolor{color2}{rgb}{0,0.6,0.5}
\definecolor{color3}{rgb}{0.8,0.6,0.7}

\begin{customlegend}[legend entries={MAP, MAP fVI, MC Dropout, MC Dropout fVI, Ensemble, Ensemble fVI, Subnetwork, Uniform prior}, legend columns={1}, legend cell align=left, legend style={draw=none, font=\footnotesize, column sep=.2cm}]
\addlegendimage{color0,semithick,dashed}
\addlegendimage{color0,semithick}
\addlegendimage{color1,semithick,dashed}
\addlegendimage{color1,semithick}
\addlegendimage{color2,semithick,dashed}
\addlegendimage{color2,semithick}
\addlegendimage{color3,semithick,dashed}
\addlegendimage{black,very thick}
\end{customlegend}
\end{tikzpicture}}}
    \caption{Comparison of rotated MNIST log-likelihood for models trained with fVI using different measurement sets.
    Colored lines denote the mean and shaded areas denote two standard deviations over 10 seeds.
    The subnetwork results are taken from \cite{pmlr-v139-daxberger21a}, who used a ResNet-18 rather than a MLP.
    }
    \label{fig:mnist_index_sets}
\end{figure*}

\paragraph{Dirichlet and KL Divergence Estimation}
\label{sec:estimation}
Assuming $\vf^{(m)}_{\vx} \sim \mathrm{Dir}(\cdot | \bm{\alpha}_\vx)$, we compute a maximum likelihood estimate (MLE) of $\bm{\alpha}_\vx$ using $M$ samples $\vf^{(m)}_{\vx}$.
To this end, we consider $\bm{\alpha}_\vx$ in terms of two separate but dependent parameters: the Dirichlet mean $\bar{\bm{\alpha}}_\vx = \bm{\alpha}_\vx / z_\vx$ and the Dirichlet precision $z_\vx$, where $\bar{\bm{\alpha}}_\vx$ are akin to categorical class probabilities and $z_\vx$ can be interpreted as a confidence score.
By matching the first moment of the empirical distribution of $\vf_\vx$, we obtain $\bar{\bm{\alpha}}_\vx \approx \textstyle\frac{1}{M} \textstyle\sum_{m=1}^M \vf^{(m)}_{\vx}$.
To estimate $z_\vx$, we fix $\bar{\bm{\alpha}}_\vx$ and employ a fast, iterative, quasi-Newton algorithm \citep{minka2000estimating}
using $M$ predictive samples
$\vf_\vx^{(1:M)} {\,=\,} \{\vf^{(1)}_\vx,\dots,\vf^{(M)}_\vx\}$,
\begin{align}
    \left(z^{(t+1)}_\vx\right)^{-1}
    &= \left(z^{(t)}_\vx\right)^{-1}
    + \left(z^{(t)}_\vx\right)^{-2}
    \frac{\partial_{z_\vx}\gL(z^{(t)}_\vx)}{\partial^2_{z_\vx}\gL(z^{(t)}_\vx)}, &&
    \gL(z^{(t)}_\vx) &= \gL_{\mathrm{Dir}}\left(\vf_\vx^{(1:M)}, \bm{\alpha}_\vx^{(t)}\right),
\end{align}
where $z_\vx^{(t)}$ and $\bm{\alpha}_\vx^{(t)} = \bar{\bm{\alpha}}_\vx / z_\vx^{(t)}$ are the Dirichlet precision and concentration at iteration $t$, and $\gL_{\mathrm{Dir}}$ is the Dirichlet log-likelihood $\log \prod_{m=1}^M \mathrm{Dir}\left(\vf_\vx^{(m)} \Big| \bm{\alpha}_\vx\right)$.
With $\bar{\bm{\alpha}}_\vx$ and $z_\vx$ estimated,
$\bm{\alpha}_\vx = \bar{\bm{\alpha}}_\vx / z_\vx$, $q(\vf_{\vx} | \bm{\theta}) = \mathrm{Dir}\left( \vf_{\vx} | \bm{\alpha}_\vx \right)$ and we compute the KL divergence as
\begin{align*}
    \KL[q \mid\mid p]
    &{\,\approx\,}
    \frac{1}{M}
    \sum_{l,m=1}^{L,M}
    \left( \log q\left(\vf_{\vs_l}^{(m)} \Big| \bm{\theta}\right)
    {-}
    \log p\left(\vf_{\vs_l}^{(m)}\right) \right),
    \label{eq:fkl_dir}
\end{align*}
where $\vf_{\vs_l}^{(m)}$ is the $m$-th prediction of the model evaluated at the $l$-th measurement item $\vs_l \in \mS$, and $\log q(\vf_{\vs_l}^{(m)} | \bm{\theta})$ and $\log p(\vf_{\vs_l}^{(m)})$ are the log-likelihood of $\vf_{\vs_l}^{(m)}$ under the variational Dirichlet posterior and Dirichlet prior respectively.
Further details about optimization and prior specification are discussed in Appendix~\ref{sec:opt_and_prior_spec}.

\section{Experiments}
\label{sec:experiments}
In this section, we present an empirical evaluation of our proposed approach, comparing the performance of several models against their conventional training procedure.
We used feedforward multilayer perceptrons (MLPs) and convolutional neural networks (CNNs).
Metrics include classification accuracy, log-likelihood (LLH) and expected calibration error (ECE) \citep{ece}, which estimates the calibration of accuracy versus confidence through binning the predicted class probabilities.
Appendix~\ref{app:implementation_details} contains additional details.
\begin{figure*}[t]
    \centering
    \resizebox{\textwidth}{!}{
    \begin{tabular}{l@{\hspace{0cm}}l@{\hspace{0cm}}l}
\begin{tikzpicture}

\definecolor{color0}{rgb}{0.9,0.6,0}
\definecolor{color1}{rgb}{0.35,0.7,0.9}
\definecolor{color2}{rgb}{0.95,0.9,0.25}
\definecolor{color3}{rgb}{0,0.6,0.5}
\definecolor{color4}{rgb}{0.8,0.6,0.7}

\begin{axis}[
height=3.5cm,
tick align=outside,
tick pos=left,
width=4cm,
x grid style={white!69.0196078431373!black},
xtick=\empty,
xmin=0, xmax=5,
xtick style={color=black},
y grid style={white!69.0196078431373!black},
ylabel style={align=center},
ylabel={\footnotesize{CIFAR10}\\\bigskip\\\footnotesize{Accuracy}},
ymin=40, ymax=100,
ytick style={color=black}
]
\path [draw=color0, fill=color0, opacity=0.1]
(axis cs:0,94.6108142283478)
--(axis cs:0,94.0271862599334)
--(axis cs:1,87.1499125311944)
--(axis cs:2,81.1647710596095)
--(axis cs:3,75.1365252202497)
--(axis cs:4,67.6240329557773)
--(axis cs:5,55.9403220110066)
--(axis cs:5,58.6249398298138)
--(axis cs:5,58.6249398298138)
--(axis cs:4,70.1027050202969)
--(axis cs:3,76.8085278803362)
--(axis cs:2,82.362386533164)
--(axis cs:1,88.1480870415595)
--(axis cs:0,94.6108142283478)
--cycle;

\path [draw=color0, fill=color0, opacity=0.1]
(axis cs:0,94.8836985198507)
--(axis cs:0,93.912302151536)
--(axis cs:1,87.3048083212329)
--(axis cs:2,81.0045292324371)
--(axis cs:3,74.795365259401)
--(axis cs:4,67.6787951253886)
--(axis cs:5,55.9341927893019)
--(axis cs:5,58.1822287194871)
--(axis cs:5,58.1822287194871)
--(axis cs:4,69.8678357339864)
--(axis cs:3,76.8227392938217)
--(axis cs:2,82.3156810336762)
--(axis cs:1,88.0214032266186)
--(axis cs:0,94.8836985198507)
--cycle;

\path [draw=color1, fill=color1, opacity=0.1]
(axis cs:0,94.5882495714158)
--(axis cs:0,94.0617489027053)
--(axis cs:1,87.7747586812721)
--(axis cs:2,81.2880830887978)
--(axis cs:3,74.6034900570646)
--(axis cs:4,66.2739794743415)
--(axis cs:5,54.1050725425429)
--(axis cs:5,57.3355585609728)
--(axis cs:5,57.3355585609728)
--(axis cs:4,68.9145459162834)
--(axis cs:3,77.0137737368807)
--(axis cs:2,82.9738111372764)
--(axis cs:1,88.6416643533959)
--(axis cs:0,94.5882495714158)
--cycle;

\path [draw=color1, fill=color1, opacity=0.1]
(axis cs:0,93.5920076248097)
--(axis cs:0,93.1739942672802)
--(axis cs:1,86.4647547258041)
--(axis cs:2,79.7259146645449)
--(axis cs:3,73.2510157972194)
--(axis cs:4,65.1843116580904)
--(axis cs:5,53.6752501467997)
--(axis cs:5,56.057380010671)
--(axis cs:5,56.057380010671)
--(axis cs:4,67.4837928951323)
--(axis cs:3,75.4701420396947)
--(axis cs:2,81.5599801108457)
--(axis cs:1,87.5542973982193)
--(axis cs:0,93.5920076248097)
--cycle;

\path [draw=color2, fill=color2, opacity=0.1]
(axis cs:0,95.2787371173643)
--(axis cs:0,94.8212644085146)
--(axis cs:1,87.553928197944)
--(axis cs:2,81.1640589961036)
--(axis cs:3,75.1643293454438)
--(axis cs:4,67.916010065066)
--(axis cs:5,55.9284201577479)
--(axis cs:5,58.553473000211)
--(axis cs:5,58.553473000211)
--(axis cs:4,69.9376749324925)
--(axis cs:3,76.6883043215484)
--(axis cs:2,82.4693073979393)
--(axis cs:1,88.3962808474662)
--(axis cs:0,95.2787371173643)
--cycle;

\path [draw=color2, fill=color2, opacity=0.1]
(axis cs:0,93.9428125283337)
--(axis cs:0,93.5131887534046)
--(axis cs:1,86.9939450100284)
--(axis cs:2,80.8845027372059)
--(axis cs:3,74.7965902827717)
--(axis cs:4,66.9999374495415)
--(axis cs:5,55.2088550643279)
--(axis cs:5,59.634724800174)
--(axis cs:5,59.634724800174)
--(axis cs:4,70.6835357560249)
--(axis cs:3,77.3909875370525)
--(axis cs:2,82.6160221651379)
--(axis cs:1,87.8680545016903)
--(axis cs:0,93.9428125283337)
--cycle;

\path [draw=gray, fill=gray, opacity=0.1]
(axis cs:0,93.9882497580299)
--(axis cs:0,93.3697504250756)
--(axis cs:1,87.2911024108317)
--(axis cs:2,81.8388752407553)
--(axis cs:3,76.3976178973349)
--(axis cs:4,69.0756218882409)
--(axis cs:5,57.4336280761399)
--(axis cs:5,59.6164764465652)
--(axis cs:5,59.6164764465652)
--(axis cs:4,70.7552191762123)
--(axis cs:3,77.3991182476846)
--(axis cs:2,82.8073344760416)
--(axis cs:1,87.9158449158285)
--(axis cs:0,93.9882497580299)
--cycle;

\path [draw=gray, fill=gray, opacity=0.1]
(axis cs:0,94.1397338279577)
--(axis cs:0,93.674269162765)
--(axis cs:1,87.4319530300184)
--(axis cs:2,81.8439769882792)
--(axis cs:3,75.8757348237261)
--(axis cs:4,68.5885462706163)
--(axis cs:5,57.0240079672054)
--(axis cs:5,59.4311495034977)
--(axis cs:5,59.4311495034977)
--(axis cs:4,70.795139032118)
--(axis cs:3,77.6635260405317)
--(axis cs:2,82.9191791396504)
--(axis cs:1,88.067731113048)
--(axis cs:0,94.1397338279577)
--cycle;

\path [draw=color3, fill=color3, opacity=0.1]
(axis cs:0,95.5419428413185)
--(axis cs:0,95.0480580742089)
--(axis cs:1,89.1591161740497)
--(axis cs:2,83.4179268581381)
--(axis cs:3,77.7619151083534)
--(axis cs:4,70.3514358577428)
--(axis cs:5,58.3401878130104)
--(axis cs:5,60.3227607953881)
--(axis cs:5,60.3227607953881)
--(axis cs:4,71.7405654850307)
--(axis cs:3,78.6626118691857)
--(axis cs:2,84.1195487277994)
--(axis cs:1,89.5728830324933)
--(axis cs:0,95.5419428413185)
--cycle;

\path [draw=color3, fill=color3, opacity=0.1]
(axis cs:0,95.4378262637647)
--(axis cs:0,95.0881747127978)
--(axis cs:1,89.2225227040233)
--(axis cs:2,83.5129903237881)
--(axis cs:3,77.9655794179121)
--(axis cs:4,70.6899097034003)
--(axis cs:5,58.7146494840801)
--(axis cs:5,60.7366139436543)
--(axis cs:5,60.7366139436543)
--(axis cs:4,72.1529308727716)
--(axis cs:3,79.0165235483964)
--(axis cs:2,84.3586954672275)
--(axis cs:1,89.6645317393361)
--(axis cs:0,95.4378262637647)
--cycle;

\path [draw=color4, fill=color4, opacity=0.1]
(axis cs:0,91)
--(axis cs:0,91)
--(axis cs:1,81)
--(axis cs:2,77)
--(axis cs:3,66)
--(axis cs:4,62)
--(axis cs:5,59)
--(axis cs:5,59)
--(axis cs:5,59)
--(axis cs:4,66)
--(axis cs:3,70)
--(axis cs:2,77)
--(axis cs:1,85)
--(axis cs:0,91)
--cycle;

\addplot [semithick, color0, dashed]
table {%
0 94.3190002441406
1 87.6489997863769
2 81.7635787963867
3 75.972526550293
4 68.8633689880371
5 57.2826309204102
};
\addplot [semithick, color0]
table {%
0 94.3980003356934
1 87.6631057739258
2 81.6601051330566
3 75.8090522766113
4 68.7733154296875
5 57.0582107543945
};
\addplot [semithick, color1, dashed]
table {%
0 94.3249992370606
1 88.208211517334
2 82.1309471130371
3 75.8086318969727
4 67.5942626953125
5 55.7203155517578
};
\addplot [semithick, color1]
table {%
0 93.3830009460449
1 87.0095260620117
2 80.6429473876953
3 74.360578918457
4 66.3340522766113
5 54.8663150787354
};
\addplot [semithick, color2, dashed]
table {%
0 95.0500007629394
1 87.9751045227051
2 81.8166831970215
3 75.9263168334961
4 68.9268424987793
5 57.2409465789795
};
\addplot [semithick, color2]
table {%
0 93.7280006408691
1 87.4309997558594
2 81.7502624511719
3 76.0937889099121
4 68.8417366027832
5 57.421789932251
};
\addplot [semithick, gray, dashed]
table {%
0 93.6790000915527
1 87.6034736633301
2 82.3231048583984
3 76.8983680725098
4 69.9154205322266
5 58.5250522613525
};
\addplot [semithick, gray]
table {%
0 93.9070014953613
1 87.7498420715332
2 82.3815780639648
3 76.7696304321289
4 69.6918426513672
5 58.2275787353516
};
\addplot [semithick, color3, dashed]
table {%
0 95.2950004577637
1 89.3659996032715
2 83.7687377929688
3 78.2122634887695
4 71.0460006713867
5 59.3314743041992
};
\addplot [semithick, color3]
table {%
0 95.2630004882812
1 89.4435272216797
2 83.9358428955078
3 78.4910514831543
4 71.4214202880859
5 59.7256317138672
};
\addplot [semithick, color4, dashed]
table {%
0 91
1 83
2 77
3 68
4 64
5 59
};
\end{axis}

\end{tikzpicture} &
\begin{tikzpicture}

\definecolor{color0}{rgb}{0.9,0.6,0}
\definecolor{color1}{rgb}{0.35,0.7,0.9}
\definecolor{color2}{rgb}{0.95,0.9,0.25}
\definecolor{color3}{rgb}{0,0.6,0.5}
\definecolor{color4}{rgb}{0.8,0.6,0.7}

\begin{axis}[
height=3.5cm,
tick align=outside,
tick pos=left,
width=4cm,
x grid style={white!69.0196078431373!black},
xtick=\empty,
xmin=0, xmax=5,
xtick style={color=black},
y grid style={white!69.0196078431373!black},
ylabel={\footnotesize{Log-Likelihood}},
ymin=-4, ymax=0,
ytick style={color=black}
]
\path [draw=color0, fill=color0, opacity=0.1]
(axis cs:0,-0.207512508560266)
--(axis cs:0,-0.225273216431343)
--(axis cs:1,-0.545712976932892)
--(axis cs:2,-0.837752319252921)
--(axis cs:3,-1.1656004164273)
--(axis cs:4,-1.61821419492043)
--(axis cs:5,-2.35496679014621)
--(axis cs:5,-2.05334109315328)
--(axis cs:5,-2.05334109315328)
--(axis cs:4,-1.42855392200694)
--(axis cs:3,-1.06879324458183)
--(axis cs:2,-0.763911517049252)
--(axis cs:1,-0.492924841348524)
--(axis cs:0,-0.207512508560266)
--cycle;

\path [draw=color0, fill=color0, opacity=0.1]
(axis cs:0,-0.236197830703572)
--(axis cs:0,-0.258495559744928)
--(axis cs:1,-0.491717658395984)
--(axis cs:2,-0.714165072099431)
--(axis cs:3,-0.943246233352625)
--(axis cs:4,-1.20550087215164)
--(axis cs:5,-1.65501593998642)
--(axis cs:5,-1.53897848866524)
--(axis cs:5,-1.53897848866524)
--(axis cs:4,-1.11016769182749)
--(axis cs:3,-0.857749877368793)
--(axis cs:2,-0.65873862095508)
--(axis cs:1,-0.461712108115287)
--(axis cs:0,-0.236197830703572)
--cycle;

\path [draw=color1, fill=color1, opacity=0.1]
(axis cs:0,-0.167124000800952)
--(axis cs:0,-0.18026119838821)
--(axis cs:1,-0.415498332687885)
--(axis cs:2,-0.673819808896183)
--(axis cs:3,-1.01470769396123)
--(axis cs:4,-1.4780014925916)
--(axis cs:5,-2.22769618612646)
--(axis cs:5,-1.96169155784273)
--(axis cs:5,-1.96169155784273)
--(axis cs:4,-1.24000415668782)
--(axis cs:3,-0.836601749145725)
--(axis cs:2,-0.578617247552648)
--(axis cs:1,-0.374434867152542)
--(axis cs:0,-0.167124000800952)
--cycle;

\path [draw=color1, fill=color1, opacity=0.1]
(axis cs:0,-0.246434859782082)
--(axis cs:0,-0.254817362620603)
--(axis cs:1,-0.462182908389294)
--(axis cs:2,-0.672746041275424)
--(axis cs:3,-0.895212853535796)
--(axis cs:4,-1.17421969483009)
--(axis cs:5,-1.61525172327526)
--(axis cs:5,-1.50836437796217)
--(axis cs:5,-1.50836437796217)
--(axis cs:4,-1.07348006976951)
--(axis cs:3,-0.799179803791128)
--(axis cs:2,-0.601762404902396)
--(axis cs:1,-0.425208357403285)
--(axis cs:0,-0.246434859782082)
--cycle;

\path [draw=color2, fill=color2, opacity=0.1]
(axis cs:0,-0.205391417118734)
--(axis cs:0,-0.224236597166522)
--(axis cs:1,-0.598482550923733)
--(axis cs:2,-0.979542709579852)
--(axis cs:3,-1.39111633038007)
--(axis cs:4,-1.88447591425285)
--(axis cs:5,-2.72791978899149)
--(axis cs:5,-2.48768673956772)
--(axis cs:5,-2.48768673956772)
--(axis cs:4,-1.69557022891824)
--(axis cs:3,-1.24014697136309)
--(axis cs:2,-0.885323307503336)
--(axis cs:1,-0.554660932145985)
--(axis cs:0,-0.205391417118734)
--cycle;

\path [draw=color2, fill=color2, opacity=0.1]
(axis cs:0,-0.270132783524127)
--(axis cs:0,-0.28131306584738)
--(axis cs:1,-0.510712730945622)
--(axis cs:2,-0.726723741592232)
--(axis cs:3,-0.954611574672485)
--(axis cs:4,-1.25205018422765)
--(axis cs:5,-1.729841136856)
--(axis cs:5,-1.49601486741557)
--(axis cs:5,-1.49601486741557)
--(axis cs:4,-1.08991796840756)
--(axis cs:3,-0.848167223713296)
--(axis cs:2,-0.660046694705978)
--(axis cs:1,-0.476235889498892)
--(axis cs:0,-0.270132783524127)
--cycle;

\path [draw=gray, fill=gray, opacity=0.1]
(axis cs:0,-0.316445479721158)
--(axis cs:0,-0.339115704898749)
--(axis cs:1,-0.744534582163776)
--(axis cs:2,-1.10855153427465)
--(axis cs:3,-1.5340750446319)
--(axis cs:4,-2.14470696602952)
--(axis cs:5,-3.1911919582765)
--(axis cs:5,-2.80207048140293)
--(axis cs:5,-2.80207048140293)
--(axis cs:4,-1.89291839832084)
--(axis cs:3,-1.38753210192058)
--(axis cs:2,-1.00163411677723)
--(axis cs:1,-0.671615018625299)
--(axis cs:0,-0.316445479721158)
--cycle;

\path [draw=gray, fill=gray, opacity=0.1]
(axis cs:0,-0.255512188460022)
--(axis cs:0,-0.274924785450064)
--(axis cs:1,-0.479240489258617)
--(axis cs:2,-0.669362679288552)
--(axis cs:3,-0.881463770200524)
--(axis cs:4,-1.14610865683964)
--(axis cs:5,-1.58413672831999)
--(axis cs:5,-1.4748526525898)
--(axis cs:5,-1.4748526525898)
--(axis cs:4,-1.058041958264)
--(axis cs:3,-0.812850161664849)
--(axis cs:2,-0.629847917323992)
--(axis cs:1,-0.457747657902336)
--(axis cs:0,-0.255512188460022)
--cycle;

\path [draw=color3, fill=color3, opacity=0.1]
(axis cs:0,-0.143297958571428)
--(axis cs:0,-0.152798936254461)
--(axis cs:1,-0.353251536140419)
--(axis cs:2,-0.555707132442149)
--(axis cs:3,-0.780645275580325)
--(axis cs:4,-1.06655176863937)
--(axis cs:5,-1.56623520588887)
--(axis cs:5,-1.45777418821871)
--(axis cs:5,-1.45777418821871)
--(axis cs:4,-0.998101796104743)
--(axis cs:3,-0.732580098071682)
--(axis cs:2,-0.527781078787823)
--(axis cs:1,-0.338864758807262)
--(axis cs:0,-0.143297958571428)
--cycle;

\path [draw=color3, fill=color3, opacity=0.1]
(axis cs:0,-0.206886588331722)
--(axis cs:0,-0.212307902352788)
--(axis cs:1,-0.38964636386453)
--(axis cs:2,-0.55976897940216)
--(axis cs:3,-0.736341258587666)
--(axis cs:4,-0.966342597547088)
--(axis cs:5,-1.36960624558302)
--(axis cs:5,-1.28432668813712)
--(axis cs:5,-1.28432668813712)
--(axis cs:4,-0.916578072388488)
--(axis cs:3,-0.70123949652235)
--(axis cs:2,-0.534803193468391)
--(axis cs:1,-0.377061271002886)
--(axis cs:0,-0.206886588331722)
--cycle;

\path [draw=color4, fill=color4, opacity=0.1]
(axis cs:0,-0.27)
--(axis cs:0,-0.27)
--(axis cs:1,-0.53)
--(axis cs:2,-0.75)
--(axis cs:3,-1.1)
--(axis cs:4,-1.31)
--(axis cs:5,-1.53)
--(axis cs:5,-1.41)
--(axis cs:5,-1.41)
--(axis cs:4,-1.19)
--(axis cs:3,-1.02)
--(axis cs:2,-0.71)
--(axis cs:1,-0.49)
--(axis cs:0,-0.27)
--cycle;

\addplot [semithick, color0, dashed]
table {%
0 -0.216392862495805
1 -0.519318909140708
2 -0.800831918151086
3 -1.11719683050457
4 -1.52338405846368
5 -2.20415394164974
};
\addplot [semithick, color0]
table {%
0 -0.24734669522425
1 -0.476714883255635
2 -0.686451846527256
3 -0.900498055360709
4 -1.15783428198956
5 -1.59699721432583
};
\addplot [semithick, color1, dashed]
table {%
0 -0.173692599594581
1 -0.394966599920213
2 -0.626218528224416
3 -0.925654721553479
4 -1.35900282463971
5 -2.0946938719846
};
\addplot [semithick, color1]
table {%
0 -0.250626111201343
1 -0.44369563289629
2 -0.63725422308891
3 -0.847196328663462
4 -1.1238498822998
5 -1.56180805061871
};
\addplot [semithick, color2, dashed]
table {%
0 -0.214814007142628
1 -0.576571741534859
2 -0.932433008541594
3 -1.31563165087158
4 -1.79002307158555
5 -2.60780326427961
};
\addplot [semithick, color2]
table {%
0 -0.275722924685753
1 -0.493474310222257
2 -0.693385218149105
3 -0.90138939919289
4 -1.17098407631761
5 -1.61292800213578
};
\addplot [semithick, gray, dashed]
table {%
0 -0.327780592309954
1 -0.708074800394538
2 -1.05509282552594
3 -1.46080357327624
4 -2.01881268217518
5 -2.99663121983971
};
\addplot [semithick, gray]
table {%
0 -0.265218486955043
1 -0.468494073580476
2 -0.649605298306272
3 -0.847156965932687
4 -1.10207530755182
5 -1.52949469045489
};
\addplot [semithick, color3, dashed]
table {%
0 -0.148048447412945
1 -0.346058147473841
2 -0.541744105614986
3 -0.756612686826004
4 -1.03232678237206
5 -1.51200469705379
};
\addplot [semithick, color3]
table {%
0 -0.209597245342255
1 -0.383353817433708
2 -0.547286086435276
3 -0.718790377555008
4 -0.941460334967788
5 -1.32696646686007
};
\addplot [semithick, color4, dashed]
table {%
0 -0.27
1 -0.51
2 -0.73
3 -1.06
4 -1.25
5 -1.47
};
\addplot [very thick, black]
table {%
0 -2.30258509299405
1 -2.30258509299405
2 -2.30258509299405
3 -2.30258509299405
4 -2.30258509299405
5 -2.30258509299405
};
\end{axis}

\end{tikzpicture} &
\begin{tikzpicture}

\definecolor{color0}{rgb}{0.9,0.6,0}
\definecolor{color1}{rgb}{0.35,0.7,0.9}
\definecolor{color2}{rgb}{0.95,0.9,0.25}
\definecolor{color3}{rgb}{0,0.6,0.5}
\definecolor{color4}{rgb}{0.8,0.6,0.7}

\begin{axis}[
height=3.5cm,
tick align=outside,
tick pos=left,
width=4cm,
x grid style={white!69.0196078431373!black},
xtick=\empty,
xmin=0, xmax=5,
xtick style={color=black},
y grid style={white!69.0196078431373!black},
ylabel={\footnotesize{ECE}},
ymin=0, ymax=0.4,
ytick style={color=black}
]
\path [draw=color0, fill=color0, opacity=0.1]
(axis cs:0,0.0343220004528229)
--(axis cs:0,0.0297373945177133)
--(axis cs:1,0.0724895053470303)
--(axis cs:2,0.112814383131984)
--(axis cs:3,0.154403910282164)
--(axis cs:4,0.202827122069466)
--(axis cs:5,0.286221170901763)
--(axis cs:5,0.322901939869416)
--(axis cs:5,0.322901939869416)
--(axis cs:4,0.228432352088821)
--(axis cs:3,0.167573341485948)
--(axis cs:2,0.121767602937695)
--(axis cs:1,0.079622738722546)
--(axis cs:0,0.0343220004528229)
--cycle;

\path [draw=color0, fill=color0, opacity=0.1]
(axis cs:0,0.0554853002754069)
--(axis cs:0,0.0509061557027482)
--(axis cs:1,0.0461728717224877)
--(axis cs:2,0.0448804724709837)
--(axis cs:3,0.0769709915731252)
--(axis cs:4,0.126311418409398)
--(axis cs:5,0.207869880439079)
--(axis cs:5,0.238986418603622)
--(axis cs:5,0.238986418603622)
--(axis cs:4,0.147237476711223)
--(axis cs:3,0.095587617078036)
--(axis cs:2,0.0505043264730127)
--(axis cs:1,0.0517316167218406)
--(axis cs:0,0.0554853002754069)
--cycle;

\path [draw=color1, fill=color1, opacity=0.1]
(axis cs:0,0.00930358599242805)
--(axis cs:0,0.00534132077055932)
--(axis cs:1,0.0255022462207664)
--(axis cs:2,0.0501661440700979)
--(axis cs:3,0.0806233086476648)
--(axis cs:4,0.131326702458758)
--(axis cs:5,0.211461839221543)
--(axis cs:5,0.251600307084495)
--(axis cs:5,0.251600307084495)
--(axis cs:4,0.163627689855199)
--(axis cs:3,0.105817881893126)
--(axis cs:2,0.0649921582668333)
--(axis cs:1,0.0326166920613657)
--(axis cs:0,0.00930358599242805)
--cycle;

\path [draw=color1, fill=color1, opacity=0.1]
(axis cs:0,0.0596356344077673)
--(axis cs:0,0.0538186434056673)
--(axis cs:1,0.0286365510426932)
--(axis cs:2,0.0305350096379353)
--(axis cs:3,0.0453734758965676)
--(axis cs:4,0.0738165555695727)
--(axis cs:5,0.150488385879522)
--(axis cs:5,0.179948103106494)
--(axis cs:5,0.179948103106494)
--(axis cs:4,0.0965020419855878)
--(axis cs:3,0.058267659348994)
--(axis cs:2,0.0383788003171847)
--(axis cs:1,0.0359157504982776)
--(axis cs:0,0.0596356344077673)
--cycle;

\path [draw=color2, fill=color2, opacity=0.1]
(axis cs:0,0.0335695157080696)
--(axis cs:0,0.0293465116739466)
--(axis cs:1,0.0765257084066905)
--(axis cs:2,0.11964328945765)
--(axis cs:3,0.163657307662005)
--(axis cs:4,0.21741438140408)
--(axis cs:5,0.309756152239746)
--(axis cs:5,0.334306915196472)
--(axis cs:5,0.334306915196472)
--(axis cs:4,0.237314546804271)
--(axis cs:3,0.179759585820203)
--(axis cs:2,0.132361944903045)
--(axis cs:1,0.083972295254942)
--(axis cs:0,0.0335695157080696)
--cycle;

\path [draw=color2, fill=color2, opacity=0.1]
(axis cs:0,0.0543950869268228)
--(axis cs:0,0.0483239406222056)
--(axis cs:1,0.0437392411919017)
--(axis cs:2,0.0428064215477487)
--(axis cs:3,0.073321677418278)
--(axis cs:4,0.12035081315204)
--(axis cs:5,0.199344406132491)
--(axis cs:5,0.254289906616418)
--(axis cs:5,0.254289906616418)
--(axis cs:4,0.15760497820214)
--(axis cs:3,0.0972910571379202)
--(axis cs:2,0.0491492737535456)
--(axis cs:1,0.0475175956098657)
--(axis cs:0,0.0543950869268228)
--cycle;

\path [draw=gray, fill=gray, opacity=0.1]
(axis cs:0,0.0456300830037549)
--(axis cs:0,0.0407761307864234)
--(axis cs:1,0.0844347259461712)
--(axis cs:2,0.122570621843551)
--(axis cs:3,0.16510132077887)
--(axis cs:4,0.217490692918876)
--(axis cs:5,0.306789374529895)
--(axis cs:5,0.339543443739834)
--(axis cs:5,0.339543443739834)
--(axis cs:4,0.238740635926148)
--(axis cs:3,0.176163955295539)
--(axis cs:2,0.132078535978581)
--(axis cs:1,0.0903687292277981)
--(axis cs:0,0.0456300830037549)
--cycle;

\path [draw=gray, fill=gray, opacity=0.1]
(axis cs:0,0.0550290451886)
--(axis cs:0,0.0483158765791116)
--(axis cs:1,0.0450235288472311)
--(axis cs:2,0.0421772252360639)
--(axis cs:3,0.0591857321174044)
--(axis cs:4,0.105933118080908)
--(axis cs:5,0.186886587034143)
--(axis cs:5,0.215435180772864)
--(axis cs:5,0.215435180772864)
--(axis cs:4,0.127443076455778)
--(axis cs:3,0.0751580484597784)
--(axis cs:2,0.0472956862708274)
--(axis cs:1,0.0480881806580885)
--(axis cs:0,0.0550290451886)
--cycle;

\path [draw=color3, fill=color3, opacity=0.1]
(axis cs:0,0.00764820301751617)
--(axis cs:0,0.0034292724371421)
--(axis cs:1,0.0198617261353603)
--(axis cs:2,0.0422349446268511)
--(axis cs:3,0.0706035810898349)
--(axis cs:4,0.107322952553259)
--(axis cs:5,0.174455814079189)
--(axis cs:5,0.198795817776776)
--(axis cs:5,0.198795817776776)
--(axis cs:4,0.124244406417383)
--(axis cs:3,0.0811278042007878)
--(axis cs:2,0.0492155119030046)
--(axis cs:1,0.0231239463683495)
--(axis cs:0,0.00764820301751617)
--cycle;

\path [draw=color3, fill=color3, opacity=0.1]
(axis cs:0,0.068702558187816)
--(axis cs:0,0.0657178560811545)
--(axis cs:1,0.0487115128635035)
--(axis cs:2,0.0291465684364354)
--(axis cs:3,0.0350029452435855)
--(axis cs:4,0.0524082939152871)
--(axis cs:5,0.0957724233183993)
--(axis cs:5,0.120057631560789)
--(axis cs:5,0.120057631560789)
--(axis cs:4,0.0650432211334552)
--(axis cs:3,0.0432162651128884)
--(axis cs:2,0.0324699443240846)
--(axis cs:1,0.0516200908721819)
--(axis cs:0,0.068702558187816)
--cycle;

\path [draw=color4, fill=color4, opacity=0.1]
(axis cs:0,0.01)
--(axis cs:0,0.01)
--(axis cs:1,0.03)
--(axis cs:2,0.06)
--(axis cs:3,0.09)
--(axis cs:4,0.11)
--(axis cs:5,0.14)
--(axis cs:5,0.18)
--(axis cs:5,0.18)
--(axis cs:4,0.15)
--(axis cs:3,0.13)
--(axis cs:2,0.06)
--(axis cs:1,0.03)
--(axis cs:0,0.01)
--cycle;

\addplot [semithick, color0, dashed]
table {%
0 0.0320296974852681
1 0.0760561220347881
2 0.11729099303484
3 0.160988625884056
4 0.215629737079144
5 0.30456155538559
};
\addplot [semithick, color0]
table {%
0 0.0531957279890776
1 0.0489522442221642
2 0.0476923994719982
3 0.0862793043255806
4 0.13677444756031
5 0.223428149521351
};
\addplot [semithick, color1, dashed]
table {%
0 0.00732245338149369
1 0.0290594691410661
2 0.0575791511684656
3 0.0932205952703953
4 0.147477196156979
5 0.231531073153019
};
\addplot [semithick, color1]
table {%
0 0.0567271389067173
1 0.0322761507704854
2 0.03445690497756
3 0.0518205676227808
4 0.0851592987775803
5 0.165218244493008
};
\addplot [semithick, color2, dashed]
table {%
0 0.0314580136910081
1 0.0802490018308163
2 0.126002617180347
3 0.171708446741104
4 0.227364464104176
5 0.322031533718109
};
\addplot [semithick, color2]
table {%
0 0.0513595137745142
1 0.0456284184008837
2 0.0459778476506472
3 0.0853063672780991
4 0.13897789567709
5 0.226817156374454
};
\addplot [semithick, gray, dashed]
table {%
0 0.0432031068950891
1 0.0874017275869846
2 0.127324578911066
3 0.170632638037205
4 0.228115664422512
5 0.323166409134865
};
\addplot [semithick, gray]
table {%
0 0.0516724608838558
1 0.0465558547526598
2 0.0447364557534456
3 0.0671718902885914
4 0.116688097268343
5 0.201160883903503
};
\addplot [semithick, color3, dashed]
table {%
0 0.00553873772732913
1 0.0214928362518549
2 0.0457252282649279
3 0.0758656926453114
4 0.115783679485321
5 0.186625815927982
};
\addplot [semithick, color3]
table {%
0 0.0672102071344853
1 0.0501658018678427
2 0.03080825638026
3 0.039109605178237
4 0.0587257575243711
5 0.107915027439594
};
\addplot [semithick, color4, dashed]
table {%
0 0.01
1 0.03
2 0.06
3 0.11
4 0.13
5 0.16
};
\end{axis}

\end{tikzpicture} \\
\begin{tikzpicture}

\definecolor{color0}{rgb}{0.9,0.6,0}
\definecolor{color1}{rgb}{0.35,0.7,0.9}
\definecolor{color2}{rgb}{0.95,0.9,0.25}
\definecolor{color3}{rgb}{0,0.6,0.5}

\begin{axis}[
height=3.5cm,
tick align=outside,
tick pos=left,
width=4cm,
x grid style={white!69.0196078431373!black},
xlabel={\footnotesize{Corruption}},
xmin=0, xmax=5,
xtick={0,1,2,3,4,5},
xtick style={color=black},
y grid style={white!69.0196078431373!black},
ylabel style={align=center},
ylabel={\footnotesize{CIFAR100}\\\bigskip\\\footnotesize{Accuracy}},
ymin=30, ymax=80,
ytick style={color=black}
]
\path [draw=color0, fill=color0, opacity=0.1]
(axis cs:0,76.1110243831565)
--(axis cs:0,75.2549759830544)
--(axis cs:1,63.8039588174587)
--(axis cs:2,54.9408889508453)
--(axis cs:3,49.3148194714666)
--(axis cs:4,42.6153883352843)
--(axis cs:5,32.510030465383)
--(axis cs:5,33.5521803805642)
--(axis cs:5,33.5521803805642)
--(axis cs:4,43.6132432565126)
--(axis cs:3,50.24044481076)
--(axis cs:2,55.8712173723969)
--(axis cs:1,64.5373032369847)
--(axis cs:0,76.1110243831565)
--cycle;

\path [draw=color0, fill=color0, opacity=0.1]
(axis cs:0,75.3100707098095)
--(axis cs:0,74.2319291070851)
--(axis cs:1,63.8141984811513)
--(axis cs:2,55.2342100864233)
--(axis cs:3,49.6131967052585)
--(axis cs:4,42.8292413383)
--(axis cs:5,33.1180926511404)
--(axis cs:5,34.502222137678)
--(axis cs:5,34.502222137678)
--(axis cs:4,44.2907583565242)
--(axis cs:3,51.0073286548489)
--(axis cs:2,56.3977898830591)
--(axis cs:1,64.7015905508311)
--(axis cs:0,75.3100707098095)
--cycle;

\path [draw=color1, fill=color1, opacity=0.1]
(axis cs:0,74.6145432794608)
--(axis cs:0,73.6814573919259)
--(axis cs:1,62.8285047523157)
--(axis cs:2,53.4960432097768)
--(axis cs:3,47.8020421009838)
--(axis cs:4,41.1004373868206)
--(axis cs:5,31.4500279007283)
--(axis cs:5,32.5878669204386)
--(axis cs:5,32.5878669204386)
--(axis cs:4,42.1691411654255)
--(axis cs:3,48.8663790721119)
--(axis cs:2,54.5874299957896)
--(axis cs:1,63.6346536644323)
--(axis cs:0,74.6145432794608)
--cycle;

\path [draw=color1, fill=color1, opacity=0.1]
(axis cs:0,72.2682292469565)
--(axis cs:0,70.7977695933756)
--(axis cs:1,60.4228276209551)
--(axis cs:2,51.2574805742341)
--(axis cs:3,45.8099225398342)
--(axis cs:4,39.283720911824)
--(axis cs:5,30.2559586815927)
--(axis cs:5,31.4828840918448)
--(axis cs:5,31.4828840918448)
--(axis cs:4,40.4757540332444)
--(axis cs:3,47.1106028202733)
--(axis cs:2,52.6186249250334)
--(axis cs:1,61.6329608006757)
--(axis cs:0,72.2682292469565)
--cycle;

\path [draw=color2, fill=color2, opacity=0.1]
(axis cs:0,77.5714196030215)
--(axis cs:0,76.2785773452207)
--(axis cs:1,63.6344434958997)
--(axis cs:2,54.430313955521)
--(axis cs:3,48.8489103635924)
--(axis cs:4,42.2647262526494)
--(axis cs:5,31.6781389898169)
--(axis cs:5,32.9395463281763)
--(axis cs:5,32.9395463281763)
--(axis cs:4,43.2902206467646)
--(axis cs:3,49.9399316469056)
--(axis cs:2,55.4483168733365)
--(axis cs:1,64.4506089943347)
--(axis cs:0,77.5714196030215)
--cycle;

\path [draw=color2, fill=color2, opacity=0.1]
(axis cs:0,75.9037082821386)
--(axis cs:0,74.6822933047755)
--(axis cs:1,64.1198317852711)
--(axis cs:2,55.7326745218452)
--(axis cs:3,50.2542861106176)
--(axis cs:4,43.5714643329038)
--(axis cs:5,33.8477856452016)
--(axis cs:5,35.0031611626597)
--(axis cs:5,35.0031611626597)
--(axis cs:4,44.8684305340395)
--(axis cs:3,51.6686615822535)
--(axis cs:2,57.2450094991509)
--(axis cs:1,65.5601685199047)
--(axis cs:0,75.9037082821386)
--cycle;

\path [draw=gray, fill=gray, opacity=0.1]
(axis cs:0,74.2414143937896)
--(axis cs:0,73.2225855451753)
--(axis cs:1,62.9783282425316)
--(axis cs:2,54.5195895587087)
--(axis cs:3,49.2741085636734)
--(axis cs:4,42.9544591177305)
--(axis cs:5,33.3549446414089)
--(axis cs:5,34.6317908932591)
--(axis cs:5,34.6317908932591)
--(axis cs:4,44.1698557626405)
--(axis cs:3,50.7313662898423)
--(axis cs:2,56.1491467083812)
--(axis cs:1,63.8879887435524)
--(axis cs:0,74.2414143937896)
--cycle;

\path [draw=gray, fill=gray, opacity=0.1]
(axis cs:0,76.1596681997812)
--(axis cs:0,74.9503339364493)
--(axis cs:1,64.9728238832477)
--(axis cs:2,56.8795991486019)
--(axis cs:3,51.5845491789665)
--(axis cs:4,44.9122395344191)
--(axis cs:5,35.0509119821102)
--(axis cs:5,36.4044575857609)
--(axis cs:5,36.4044575857609)
--(axis cs:4,46.3499713115281)
--(axis cs:3,52.8902928925667)
--(axis cs:2,58.2287168914372)
--(axis cs:1,66.0127550352094)
--(axis cs:0,76.1596681997812)
--cycle;

\path [draw=color3, fill=color3, opacity=0.1]
(axis cs:0,79.4503645501281)
--(axis cs:0,78.803637830243)
--(axis cs:1,67.6787826244402)
--(axis cs:2,58.7977223034845)
--(axis cs:3,52.9874135307176)
--(axis cs:4,46.0804128510067)
--(axis cs:5,35.3283479841535)
--(axis cs:5,36.0422834245379)
--(axis cs:5,36.0422834245379)
--(axis cs:4,46.8162195342472)
--(axis cs:3,53.8561655708449)
--(axis cs:2,59.5852244738593)
--(axis cs:1,68.3216400440168)
--(axis cs:0,79.4503645501281)
--cycle;

\path [draw=color3, fill=color3, opacity=0.1]
(axis cs:0,76.2982777069656)
--(axis cs:0,75.4837199736985)
--(axis cs:1,65.9226643455717)
--(axis cs:2,57.4044831637174)
--(axis cs:3,51.6845091669069)
--(axis cs:4,44.5432812077066)
--(axis cs:5,34.4891537388032)
--(axis cs:5,35.8834788600738)
--(axis cs:5,35.8834788600738)
--(axis cs:4,45.7315601008872)
--(axis cs:3,52.7834908636107)
--(axis cs:2,58.5452025052279)
--(axis cs:1,66.8349131690767)
--(axis cs:0,76.2982777069656)
--cycle;

\addplot [semithick, color0, dashed]
table {%
0 75.6830001831055
1 64.1706310272217
2 55.4060531616211
3 49.7776321411133
4 43.1143157958984
5 33.0311054229736
};
\addplot [semithick, color0]
table {%
0 74.7709999084473
1 64.2578945159912
2 55.8159999847412
3 50.3102626800537
4 43.5599998474121
5 33.8101573944092
};
\addplot [semithick, color1, dashed]
table {%
0 74.1480003356934
1 63.231579208374
2 54.0417366027832
3 48.3342105865479
4 41.634789276123
5 32.0189474105835
};
\addplot [semithick, color1]
table {%
0 71.532999420166
1 61.0278942108154
2 51.9380527496338
3 46.4602626800537
4 39.8797374725342
5 30.8694213867187
};
\addplot [semithick, color2, dashed]
table {%
0 76.9249984741211
1 64.0425262451172
2 54.9393154144287
3 49.394421005249
4 42.777473449707
5 32.3088426589966
};
\addplot [semithick, color2]
table {%
0 75.293000793457
1 64.8400001525879
2 56.488842010498
3 50.9614738464355
4 44.2199474334717
5 34.4254734039307
};
\addplot [semithick, gray, dashed]
table {%
0 73.7319999694824
1 63.433158493042
2 55.3343681335449
3 50.0027374267578
4 43.5621574401855
5 33.993367767334
};
\addplot [semithick, gray]
table {%
0 75.5550010681152
1 65.4927894592285
2 57.5541580200195
3 52.2374210357666
4 45.6311054229736
5 35.7276847839355
};
\addplot [semithick, color3, dashed]
table {%
0 79.1270011901856
1 68.0002113342285
2 59.1914733886719
3 53.4217895507812
4 46.448316192627
5 35.6853157043457
};
\addplot [semithick, color3]
table {%
0 75.890998840332
1 66.3787887573242
2 57.9748428344727
3 52.2340000152588
4 45.1374206542969
5 35.1863162994385
};
\end{axis}

\end{tikzpicture} &
\begin{tikzpicture}

\definecolor{color0}{rgb}{0.9,0.6,0}
\definecolor{color1}{rgb}{0.35,0.7,0.9}
\definecolor{color2}{rgb}{0.95,0.9,0.25}
\definecolor{color3}{rgb}{0,0.6,0.5}

\begin{axis}[
height=3.5cm,
tick align=outside,
tick pos=left,
width=4cm,
x grid style={white!69.0196078431373!black},
xlabel={\footnotesize{Corruption}},
xmin=0, xmax=5,
xtick style={color=black},
xtick={0,1,2,3,4,5},
y grid style={white!69.0196078431373!black},
ylabel={\footnotesize{Log-Likelihood}},
ymin=-6, ymax=-0.5,
ytick style={color=black}
]
\path [draw=color0, fill=color0, opacity=0.1]
(axis cs:0,-0.978680522889131)
--(axis cs:0,-1.01425473170359)
--(axis cs:1,-1.61677210239266)
--(axis cs:2,-2.12795604964373)
--(axis cs:3,-2.52640900875355)
--(axis cs:4,-3.04614799377624)
--(axis cs:5,-3.81703403786082)
--(axis cs:5,-3.63858081980871)
--(axis cs:5,-3.63858081980871)
--(axis cs:4,-2.92679340568897)
--(axis cs:3,-2.44322912125736)
--(axis cs:2,-2.05118633341619)
--(axis cs:1,-1.56076744266796)
--(axis cs:0,-0.978680522889131)
--cycle;

\path [draw=color0, fill=color0, opacity=0.1]
(axis cs:0,-1.18122219078966)
--(axis cs:0,-1.21617439337194)
--(axis cs:1,-1.7122343923705)
--(axis cs:2,-2.11203285845559)
--(axis cs:3,-2.38723278095022)
--(axis cs:4,-2.7413446747012)
--(axis cs:5,-3.23927682295913)
--(axis cs:5,-3.14496927102651)
--(axis cs:5,-3.14496927102651)
--(axis cs:4,-2.64865500429491)
--(axis cs:3,-2.31099864082778)
--(axis cs:2,-2.04590887387387)
--(axis cs:1,-1.66456328769199)
--(axis cs:0,-1.18122219078966)
--cycle;

\path [draw=color1, fill=color1, opacity=0.1]
(axis cs:0,-0.951453753744882)
--(axis cs:0,-0.985666016046563)
--(axis cs:1,-1.5390338099285)
--(axis cs:2,-2.06205863598949)
--(axis cs:3,-2.47821441785523)
--(axis cs:4,-3.03693794128766)
--(axis cs:5,-3.87378208554251)
--(axis cs:5,-3.63676395959524)
--(axis cs:5,-3.63676395959524)
--(axis cs:4,-2.87644740519967)
--(axis cs:3,-2.36601768847989)
--(axis cs:2,-1.98614982792976)
--(axis cs:1,-1.48955650003307)
--(axis cs:0,-0.951453753744882)
--cycle;

\path [draw=color1, fill=color1, opacity=0.1]
(axis cs:0,-1.14933817069797)
--(axis cs:0,-1.18472294880674)
--(axis cs:1,-1.68109129159437)
--(axis cs:2,-2.12815823469795)
--(axis cs:3,-2.43384425276949)
--(axis cs:4,-2.83899773702728)
--(axis cs:5,-3.41600738152495)
--(axis cs:5,-3.2931607244625)
--(axis cs:5,-3.2931607244625)
--(axis cs:4,-2.73276429800595)
--(axis cs:3,-2.33870107416364)
--(axis cs:2,-2.04215464771894)
--(axis cs:1,-1.6163747314329)
--(axis cs:0,-1.14933817069797)
--cycle;

\path [draw=color2, fill=color2, opacity=0.1]
(axis cs:0,-0.928936560954541)
--(axis cs:0,-0.96811236272942)
--(axis cs:1,-1.6704541275979)
--(axis cs:2,-2.2668694947629)
--(axis cs:3,-2.72995604556089)
--(axis cs:4,-3.2982966451496)
--(axis cs:5,-4.19837478782079)
--(axis cs:5,-3.83393751659963)
--(axis cs:5,-3.83393751659963)
--(axis cs:4,-3.03462818494898)
--(axis cs:3,-2.54454894146302)
--(axis cs:2,-2.14799365183959)
--(axis cs:1,-1.61145131764461)
--(axis cs:0,-0.928936560954541)
--cycle;

\path [draw=color2, fill=color2, opacity=0.1]
(axis cs:0,-1.1850148326513)
--(axis cs:0,-1.23108516722663)
--(axis cs:1,-1.72355970518792)
--(axis cs:2,-2.11065673088611)
--(axis cs:3,-2.37066143855357)
--(axis cs:4,-2.70564794050911)
--(axis cs:5,-3.18649719432343)
--(axis cs:5,-3.13294170828827)
--(axis cs:5,-3.13294170828827)
--(axis cs:4,-2.64195489943651)
--(axis cs:3,-2.30387039434169)
--(axis cs:2,-2.04031511806308)
--(axis cs:1,-1.66401800520933)
--(axis cs:0,-1.1850148326513)
--cycle;

\path [draw=gray, fill=gray, opacity=0.1]
(axis cs:0,-1.44800675901304)
--(axis cs:0,-1.51602193991352)
--(axis cs:1,-2.33940698727699)
--(axis cs:2,-3.09980806696138)
--(axis cs:3,-3.69757218650902)
--(axis cs:4,-4.47784674404268)
--(axis cs:5,-5.66843889848962)
--(axis cs:5,-5.43380846234823)
--(axis cs:5,-5.43380846234823)
--(axis cs:4,-4.24538777911175)
--(axis cs:3,-3.46501611105594)
--(axis cs:2,-2.91505265537448)
--(axis cs:1,-2.24469610349999)
--(axis cs:0,-1.44800675901304)
--cycle;

\path [draw=gray, fill=gray, opacity=0.1]
(axis cs:0,-1.15633179892758)
--(axis cs:0,-1.18648761323247)
--(axis cs:1,-1.6618458406318)
--(axis cs:2,-2.02898064637302)
--(axis cs:3,-2.2782582113477)
--(axis cs:4,-2.61200119416222)
--(axis cs:5,-3.101626833347)
--(axis cs:5,-3.02587889249313)
--(axis cs:5,-3.02587889249313)
--(axis cs:4,-2.54150704100674)
--(axis cs:3,-2.21460724499451)
--(axis cs:2,-1.96330897215232)
--(axis cs:1,-1.60870533273415)
--(axis cs:0,-1.15633179892758)
--cycle;

\path [draw=color3, fill=color3, opacity=0.1]
(axis cs:0,-0.805167361504395)
--(axis cs:0,-0.82199864442065)
--(axis cs:1,-1.31680502489541)
--(axis cs:2,-1.7180340613806)
--(axis cs:3,-2.00880869376598)
--(axis cs:4,-2.40283379587222)
--(axis cs:5,-2.97275872740162)
--(axis cs:5,-2.89429138256942)
--(axis cs:5,-2.89429138256942)
--(axis cs:4,-2.33632127388852)
--(axis cs:3,-1.9594004318409)
--(axis cs:2,-1.6778043053998)
--(axis cs:1,-1.28742170935188)
--(axis cs:0,-0.805167361504395)
--cycle;

\path [draw=color3, fill=color3, opacity=0.1]
(axis cs:0,-1.17090777420686)
--(axis cs:0,-1.18902582502715)
--(axis cs:1,-1.62492957059372)
--(axis cs:2,-2.00091075657161)
--(axis cs:3,-2.26138498236708)
--(axis cs:4,-2.60699248567148)
--(axis cs:5,-3.09249548105667)
--(axis cs:5,-3.02527081741489)
--(axis cs:5,-3.02527081741489)
--(axis cs:4,-2.55117160799252)
--(axis cs:3,-2.2143161523856)
--(axis cs:2,-1.96058430933105)
--(axis cs:1,-1.59610025547475)
--(axis cs:0,-1.17090777420686)
--cycle;

\addplot [semithick, color0, dashed]
table {%
0 -0.99646762729636
1 -1.58876977253031
2 -2.08957119152996
3 -2.48481906500545
4 -2.98647069973261
5 -3.72780742883476
};
\addplot [semithick, color0]
table {%
0 -1.1986982920808
1 -1.68839884003124
2 -2.07897086616473
3 -2.349115710889
4 -2.69499983949806
5 -3.19212304699282
};
\addplot [semithick, color1, dashed]
table {%
0 -0.968559884895723
1 -1.51429515498078
2 -2.02410423195962
3 -2.42211605316756
4 -2.95669267324367
5 -3.75527302256887
};
\addplot [semithick, color1]
table {%
0 -1.16703055975235
1 -1.64873301151364
2 -2.08515644120844
3 -2.38627266346656
4 -2.78588101751662
5 -3.35458405299372
};
\addplot [semithick, color2, dashed]
table {%
0 -0.948524461841981
1 -1.64095272262126
2 -2.20743157330124
3 -2.63725249351195
4 -3.16646241504929
5 -4.01615615221021
};
\addplot [semithick, color2]
table {%
0 -1.20804999993896
1 -1.69378885519862
2 -2.07548592447459
3 -2.33726591644763
4 -2.67380141997281
5 -3.15971945130585
};
\addplot [semithick, gray, dashed]
table {%
0 -1.48201434946328
1 -2.29205154538849
2 -3.00743036116793
3 -3.58129414878248
4 -4.36161726157722
5 -5.55112368041893
};
\addplot [semithick, gray]
table {%
0 -1.17140970608002
1 -1.63527558668297
2 -1.99614480926267
3 -2.2464327281711
4 -2.57675411758448
5 -3.06375286292007
};
\addplot [semithick, color3, dashed]
table {%
0 -0.813583002962523
1 -1.30211336712365
2 -1.6979191833902
3 -1.98410456280344
4 -2.36957753488037
5 -2.93352505498552
};
\addplot [semithick, color3]
table {%
0 -1.179966799617
1 -1.61051491303424
2 -1.98074753295133
3 -2.23785056737634
4 -2.579082046832
5 -3.05888314923578
};
\addplot [very thick, black]
table {%
0  -4.60517
1  -4.60517
2  -4.60517
3  -4.60517
4  -4.60517
5  -4.60517
};
\end{axis}

\end{tikzpicture} &
\begin{tikzpicture}

\definecolor{color0}{rgb}{0.9,0.6,0}
\definecolor{color1}{rgb}{0.35,0.7,0.9}
\definecolor{color2}{rgb}{0.95,0.9,0.25}
\definecolor{color3}{rgb}{0,0.6,0.5}

\begin{axis}[
height=3.5cm,
tick align=outside,
tick pos=left,
width=4cm,
x grid style={white!69.0196078431373!black},
xlabel={\footnotesize{Corruption}},
xmin=0, xmax=5,
xtick style={color=black},
xtick={0,1,2,3,4,5},
y grid style={white!69.0196078431373!black},
ylabel={\footnotesize{ECE}},
ymin=0, ymax=0.5,
ytick style={color=black}
]
\path [draw=color0, fill=color0, opacity=0.1]
(axis cs:0,0.083912122044592)
--(axis cs:0,0.0721508460306834)
--(axis cs:1,0.119400269320597)
--(axis cs:2,0.157348926925555)
--(axis cs:3,0.187663884908739)
--(axis cs:4,0.225915476223409)
--(axis cs:5,0.284650854800733)
--(axis cs:5,0.307489837671725)
--(axis cs:5,0.307489837671725)
--(axis cs:4,0.24402920053345)
--(axis cs:3,0.202881629317221)
--(axis cs:2,0.168789480066404)
--(axis cs:1,0.127170973488698)
--(axis cs:0,0.083912122044592)
--cycle;

\path [draw=color0, fill=color0, opacity=0.1]
(axis cs:0,0.130354879610443)
--(axis cs:0,0.113729229755497)
--(axis cs:1,0.102183539651764)
--(axis cs:2,0.0797895622283682)
--(axis cs:3,0.053134629695925)
--(axis cs:4,0.0312133014008114)
--(axis cs:5,0.0258403480975869)
--(axis cs:5,0.0463182120592353)
--(axis cs:5,0.0463182120592353)
--(axis cs:4,0.0446759619012764)
--(axis cs:3,0.0723223650760324)
--(axis cs:2,0.0910545545786158)
--(axis cs:1,0.114576487756836)
--(axis cs:0,0.130354879610443)
--cycle;

\path [draw=color1, fill=color1, opacity=0.1]
(axis cs:0,0.0293499955047057)
--(axis cs:0,0.020157475296767)
--(axis cs:1,0.0581448552462264)
--(axis cs:2,0.0963174229176995)
--(axis cs:3,0.123855519583951)
--(axis cs:4,0.162236786146513)
--(axis cs:5,0.221606640985513)
--(axis cs:5,0.254063896367049)
--(axis cs:5,0.254063896367049)
--(axis cs:4,0.184678852300295)
--(axis cs:3,0.141108181783428)
--(axis cs:2,0.10773850626626)
--(axis cs:1,0.0669877032485799)
--(axis cs:0,0.0293499955047057)
--cycle;

\path [draw=color1, fill=color1, opacity=0.1]
(axis cs:0,0.0908137159483978)
--(axis cs:0,0.078216332004731)
--(axis cs:1,0.0519536129172739)
--(axis cs:2,0.016992676459618)
--(axis cs:3,0.0087974419725146)
--(axis cs:4,0.0424143203603001)
--(axis cs:5,0.095316286287159)
--(axis cs:5,0.112410047748238)
--(axis cs:5,0.112410047748238)
--(axis cs:4,0.0510933837784557)
--(axis cs:3,0.0143476348462496)
--(axis cs:2,0.0232719188403412)
--(axis cs:1,0.0597327099207941)
--(axis cs:0,0.0908137159483978)
--cycle;

\path [draw=color2, fill=color2, opacity=0.1]
(axis cs:0,0.0923708285014165)
--(axis cs:0,0.0793996100504863)
--(axis cs:1,0.136246456885992)
--(axis cs:2,0.180965841720068)
--(axis cs:3,0.212990309336015)
--(axis cs:4,0.250880856147221)
--(axis cs:5,0.316367460379103)
--(axis cs:5,0.358180343022844)
--(axis cs:5,0.358180343022844)
--(axis cs:4,0.281689863571712)
--(axis cs:3,0.235767711661032)
--(axis cs:2,0.197970279147662)
--(axis cs:1,0.147692767595591)
--(axis cs:0,0.0923708285014165)
--cycle;

\path [draw=color2, fill=color2, opacity=0.1]
(axis cs:0,0.14041113553597)
--(axis cs:0,0.126309376973376)
--(axis cs:1,0.115979878050268)
--(axis cs:2,0.0937904500438626)
--(axis cs:3,0.0696962707392403)
--(axis cs:4,0.0422469130982351)
--(axis cs:5,0.0280842674862322)
--(axis cs:5,0.0445416240055386)
--(axis cs:5,0.0445416240055386)
--(axis cs:4,0.0573888712297488)
--(axis cs:3,0.0898181415446571)
--(axis cs:2,0.109461020581538)
--(axis cs:1,0.128525470751345)
--(axis cs:0,0.14041113553597)
--cycle;

\path [draw=gray, fill=gray, opacity=0.1]
(axis cs:0,0.162865425332033)
--(axis cs:0,0.15188771368698)
--(axis cs:1,0.220251654266381)
--(axis cs:2,0.271205068678027)
--(axis cs:3,0.310209361914152)
--(axis cs:4,0.359724691146614)
--(axis cs:5,0.432098563254164)
--(axis cs:5,0.443651805579378)
--(axis cs:5,0.443651805579378)
--(axis cs:4,0.370489159590004)
--(axis cs:3,0.322558988925463)
--(axis cs:2,0.284137331395978)
--(axis cs:1,0.226020358324981)
--(axis cs:0,0.162865425332033)
--cycle;

\path [draw=gray, fill=gray, opacity=0.1]
(axis cs:0,0.140241354964921)
--(axis cs:0,0.128286173678687)
--(axis cs:1,0.120545684630243)
--(axis cs:2,0.0997808378686514)
--(axis cs:3,0.0778050087785026)
--(axis cs:4,0.0447559317571164)
--(axis cs:5,0.0204263059236275)
--(axis cs:5,0.038455664198829)
--(axis cs:5,0.038455664198829)
--(axis cs:4,0.0671997221189498)
--(axis cs:3,0.0955159730816582)
--(axis cs:2,0.114144680226842)
--(axis cs:1,0.130361772244604)
--(axis cs:0,0.140241354964921)
--cycle;

\path [draw=color3, fill=color3, opacity=0.1]
(axis cs:0,0.0562233202294401)
--(axis cs:0,0.0447464227541395)
--(axis cs:1,0.0386091291070104)
--(axis cs:2,0.0195067145733924)
--(axis cs:3,0.00933952787366139)
--(axis cs:4,0.0343798936473306)
--(axis cs:5,0.0858631313101835)
--(axis cs:5,0.109061615143579)
--(axis cs:5,0.109061615143579)
--(axis cs:4,0.0548633455170218)
--(axis cs:3,0.0225294915080406)
--(axis cs:2,0.027715141990891)
--(axis cs:1,0.0452470273613811)
--(axis cs:0,0.0562233202294401)
--cycle;

\path [draw=color3, fill=color3, opacity=0.1]
(axis cs:0,0.190229483544795)
--(axis cs:0,0.177602470815213)
--(axis cs:1,0.17408237392129)
--(axis cs:2,0.154891931161909)
--(axis cs:3,0.131060557910209)
--(axis cs:4,0.0943522374697866)
--(axis cs:5,0.0358566237065309)
--(axis cs:5,0.0539371209767348)
--(axis cs:5,0.0539371209767348)
--(axis cs:4,0.108178997450238)
--(axis cs:3,0.143701559952493)
--(axis cs:2,0.166253636136027)
--(axis cs:1,0.184854311402425)
--(axis cs:0,0.190229483544795)
--cycle;

\addplot [semithick, color0, dashed]
table {%
0 0.0780314840376377
1 0.123285621404648
2 0.163069203495979
3 0.19527275711298
4 0.234972338378429
5 0.296070346236229
};
\addplot [semithick, color0]
table {%
0 0.12204205468297
1 0.1083800137043
2 0.085422058403492
3 0.0627284973859787
4 0.0379446316510439
5 0.0360792800784111
};
\addplot [semithick, color1, dashed]
table {%
0 0.0247537354007363
1 0.0625662792474031
2 0.10202796459198
3 0.132481850683689
4 0.173457819223404
5 0.237835268676281
};
\addplot [semithick, color1]
table {%
0 0.0845150239765644
1 0.055843161419034
2 0.0201322976499796
3 0.0115725384093821
4 0.0467538520693779
5 0.103863167017698
};
\addplot [semithick, color2, dashed]
table {%
0 0.0858852192759514
1 0.141969612240791
2 0.189468060433865
3 0.224379010498524
4 0.266285359859467
5 0.337273901700973
};
\addplot [semithick, color2]
table {%
0 0.133360256254673
1 0.122252674400806
2 0.1016257353127
3 0.0797572061419487
4 0.0498178921639919
5 0.0363129457458854
};
\addplot [semithick, gray, dashed]
table {%
0 0.157376569509506
1 0.223136006295681
2 0.277671200037003
3 0.316384175419807
4 0.365106925368309
5 0.437875184416771
};
\addplot [semithick, gray]
table {%
0 0.134263764321804
1 0.125453728437424
2 0.106962759047747
3 0.0866604909300804
4 0.0559778269380331
5 0.0294409850612283
};
\addplot [semithick, color3, dashed]
table {%
0 0.0504848714917898
1 0.0419280782341957
2 0.0236109282821417
3 0.015934509690851
4 0.0446216195821762
5 0.097462373226881
};
\addplot [semithick, color3]
table {%
0 0.183915977180004
1 0.179468342661858
2 0.160572783648968
3 0.137381058931351
4 0.101265617460012
5 0.0448968723416328
};
\end{axis}

\end{tikzpicture}
    \end{tabular}
        \raisebox{-2.25cm}{\begin{tikzpicture}

\newenvironment{customlegend}[1][]{%
    \begingroup
    \csname pgfplots@init@cleared@structures\endcsname
    \pgfplotsset{#1}%
}{%
    \csname pgfplots@createlegend\endcsname
    \endgroup
}%

\def\addlegendimage{\csname pgfplots@addlegendimage\endcsname}

\definecolor{color0}{rgb}{0.9,0.6,0}
\definecolor{color1}{rgb}{0.35,0.7,0.9}
\definecolor{color2}{rgb}{0,0.6,0.5}
\definecolor{color3}{rgb}{0.95,0.9,0.25}
\definecolor{color4}{rgb}{0.5,0.5,0.5}
\definecolor{color5}{rgb}{0.8,0.6,0.7}

\begin{customlegend}[legend entries={MAP, MAP fVI, MC Dropout, MC Dropout fVI, Ensemble, Ensemble fVI, Radial, Radial fVI, Rank-1, Rank-1 fVI, Subnetwork, Uniform prior}, legend columns={1}, legend cell align=left, legend style={draw=none, font=\footnotesize, column sep=.1cm}]
\addlegendimage{color0,semithick,dashed}
\addlegendimage{color0,semithick}
\addlegendimage{color1,semithick,dashed}
\addlegendimage{color1,semithick}
\addlegendimage{color2,semithick,dashed}
\addlegendimage{color2,semithick}
\addlegendimage{color3,semithick,dashed}
\addlegendimage{color3,semithick}
\addlegendimage{color4,semithick,dashed}
\addlegendimage{color4,semithick}
\addlegendimage{color5,semithick,dashed}
\addlegendimage{black, very thick}
\end{customlegend}
\end{tikzpicture}}}
    \caption{Image classification on corrupted CIFAR10 and CIFAR100.
    All models use a ResNet-18 architecture. 
    For CIFAR10, there is a clear benefit of fVI priors over weight-space for log-likelihood.
    For CIFAR100, the higher label dimensionality results in stronger regularization from the uniform prior. 
    This indicates prior specification requires more care for high-dimensional classification.
    Subnetwork results are taken from \cite{pmlr-v139-daxberger21a}.
    }
    \vspace{-3pt}
    \label{fig:cifar_corrupted}
\end{figure*}
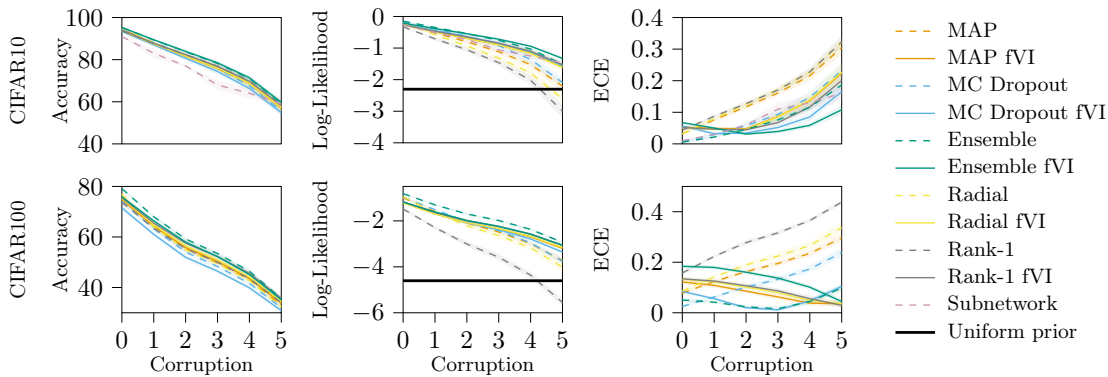

\paragraph{Toy Problem}
\label{sec:toy_problem}
To visualize the effects of function-space variational inference, we conducted a toy experiment with the Two Moons data set and MLP models.
We used MAP, MC Dropout \citep{Gal16}, and deep ensemble \citep{NIPS2017_7219} models, training each with their regular weight-space method, a uniform function prior, a GP function prior and a random forest function prior.
Figure~\ref{fig:two_moons} shows that weight-space training leads to overconfident extrapolation, while our inference approach combined with Dirichlet function priors adequately increases predictive uncertainty outside of the observed data.
While the weight-space approach learns a decision boundary that bisects the data, the function-space approaches learn richer boundaries which capture the data distribution more accurately and display properties that resemble the respective prior.
\paragraph{Rotated MNIST}
Following \cite{NIPS2019_9547}, we train on the MNIST handwritten digit classification data set \citep{lecun2010mnist} and evaluate on constructed test data with rotations of up to 180$^\circ$, which simulates a challenging OOD scenario due to the absence of data augmentation.
For this experiment, we used the same MLP models as for the toy problem.
The log-likelihood between models is shown in Figure ~\ref{fig:mnist_index_sets}.
In terms of classification error, weight-space inference and fVI yield the same performance.
In terms of log-likelihood, fVI consistently outperforms their weight-space counterparts as the data becomes more OOD.
Subnetwork linearized Laplace \citep{pmlr-v139-daxberger21a} is also reported as a competitive baseline,
however, these results were obtained using a ResNet-18.
\paragraph{Assessing Measurement Set Design}
To illustrate the importance of the measurement set, we train the fVI models for rotated MNIST using three different measurement sets: the training data, additional 90\textdegree{} augmentation, and additional 90\textdegree{} and 180\textdegree{} augmentation.
While simply using the training data without rotations already outperforms the weight-space counterparts, a direct comparison in Figure~\ref{fig:mnist_index_sets} illustrates that performance can be further increased if an appropriate measurement set, i.e. example OOD data, is available.
With the enriched measurement sets, the OOD performance move closer to that of the prior, indicating more accurate inclusion in the fELBO.
Sets for greater OOD performance could be designed through manual data augmentation, unlabeled data, or synthetic data generation.
Note, for all other image classification experiments, we use the training data as the measurement set.
\paragraph{Image Classification under Corruption}
\label{sec:corrupt}
We used the regular train splits of the CIFAR10 and CIFAR100 \citep{Krizhevsky09learningmultiple} as training data and their corrupted versions \citep{hendrycks2018benchmarking} as OOD test data.
CIFAR10 and CIFAR100 consist of natural color images of animals and vehicles.
Their corrupted versions perturb the images at five increasing levels of severity by changing the brightness, contrast or saturation, or adding noise, blur or other artifacts, such that classification becomes more difficult.
For this experiment, we used ResNet-18 CNN models \citep{7780459}.
In addition to the previous MAP, MC Dropout and deep ensemble model types, we also evaluate our fVI approach on Radial BNNs \citep{farquhar_radial_2020}, as an effective variant of MFVI, and Rank-1 BNNs \citep{dusenberry2020efficient}, which combine ensembles and VI.
In Appendix~\ref{sec:adversarial}, we investigate a similar setting where the corruptions are replaced by adversarial attacks of varying strength.

Figure~\ref{fig:cifar_corrupted} shows the results for CIFAR10 and CIFAR100 under corruption.
The function-space prior frequently provides gains in OOD uncertainty quantification with only a small decrease in (uncorrupted) test performance.
This trade-off between accuracy and robustness has been observed and discussed in the adversarial robustness setting \citep{tsipras2018robustness, accrob} and it remains an open problem if and how both qualities can be achieved in practice.
Moreover, the shared function-space prior resulted in remarkable consistency across models, compared to the variety seen in weight-space priors.
For CIFAR100, higher prior regularization due to higher dimensionality (see Appendix \ref{app:ablations}) resulted in reduced benefit over weight-space models, with improved performance only evident at stronger corruptions. 

\section{Conclusion}
\label{sec:conclusion}
We propose an approach to function-space regularization for deep Bayesian classification, which enables the use of Dirichlet predictive priors to improve uncertainty quantification.
Our approach provides a generic view of prior work on Dirichlet-based classifiers with function-space regularization, and can be applied to a general class of BNNs and stochastic models without altering their underlying architectures and mechanisms.
Experiments demonstrate that our approach generally outperforms the corresponding weight-space priors in terms of uncertainty quantification and adversarial robustness.
Different measurement sets can trade-off scalability against OOD uncertainty quantification by extending the fKL evaluation beyond the training data.
Future research should improve measurement sets for fVI, for example, by developing effective methods for constructing them to reflect the test distribution, e.g. through using data augmentation or unlabeled data.


\bibliography{references}
\newpage
\appendix
\section{Related Work}
\label{sec:related_work}
In this section, we summarize related work on Bayesian classification and function-space inference, and discuss previous research which is of particular relevance to our work.
\paragraph{Bayesian Classification}
Compared to regression, classification is non-trivial for Bayesian methods due to the nonlinear link function required to predict the class labels.
As a result, closed-form Bayesian models, such as Gaussian processes (GP), require approximate inference methods such as the Laplace approximation \citep{Rasmussen05}, variational inference \citep{883477,pmlr-v38-hensman15,pmlr-v84-salimbeni18a,pmlr-v84-izmailov18a}, and expectation propagation \citep{pmlr-v51-hernandez-lobato16}.
The P\'{o}lya-Gamma data augmentation trick \citep{polyagamma} has enabled scalable closed-form variational training of sparse Gaussian process classifiers \citep{polya}. 
Gaussian processes have also been used with a Dirichlet predictive using a log-normal approximation \citep{dirichlet_gp}.

Classification with Bayesian neural networks is possible through a wide range of approximate inference methods, including Markov chain Monte Carlo \citep{radfordthesis, Zhang2020Cyclical}, (mean-field) variational inference (MFVI) \citep{NIPS2011_4329, pmlr-v37-blundell15,zhang2018noisy}, Laplace approximations \citep{mackayBnn, NIPS1990_419, ritter2018a, NEURIPS2019_b3bbccd6, pmlr-v130-immer21a, pmlr-v139-daxberger21a}, ensembles \citep{NIPS2017_7219, randomizedPriorFunctions, barber1998ensemble, pearce2018uncertainty}, expectation propagation \citep{probBP} and Monte Carlo dropout \citep{Gal16, NIPS2015_5666}.
Radial BNNs \citep{farquhar_radial_2020} are motivated as a practical alternative to MFVI BNNs that uses Gaussian weight priors and posteriors.
By sampling weights in a radial fashion, they avoid the pathologies encountered when sampling high-dimensional Gaussian distributions.
Rank-1 BNNs \citep{dusenberry2020efficient} combine ensembles and weight priors.
Using the shared BatchEnsemble structure \citep{Wen2020BatchEnsemble} and Rank-1 covariance parameterizations, Rank-1 BNNs have a scalable memory requirement.
Alternatively, the Laplace bridge \citep{hobbhahn2021fast} approximately maps a Dirichlet predictive density backwards through the softmax into a latent Gaussian predictive. 
A Gaussian-predictive BNN can then be trained using this latent approximation. 

Alternative methods avoid propagating uncertainties by predicting Dirichlet concentrations directly with deep neural networks.
Prior networks \citep{malinin2018predictive, NEURIPS2019_7dd2ae7d} require categorical labels to be converted to Dirichlets, and resembles fVI as the objective consists of two KL divergences, for in- and outside the data distribution respectively.
They can be used to distill a trained ensemble into a single model \citep{Malinin2020Ensemble}.
Similarly, belief matching \citep{joo2020being} converts training labels to Dirichlets using Bayes rule.
This method can also be viewed as fVI where the measurement set is the training data.
Another method converts the training labels to categorical probabilities and uses a Bayes risk objective with KL regularization against a function-space prior \citep{sensoy2018evidential, Sensoy_2021_WACV}.
Compared to these methods, we introduce generic function-space regularization that allows us to use any BNN or stochastic model with the conventional categorical likelihood, avoiding the need to design networks and data representations that facilitate a model-specific training approach.
A longer discussion and comparison is provided later in this section.

\paragraph{Function-Space Variational Inference}
Function-space variational inference generalizes conventional variational inference over finite weight distributions to inference over stochastic processes, which entails difficulties because the standard KL divergence between finite-dimensional probability distributions becomes an infinite-dimensional fKL divergence between stochastic processes.
Gaussian processes \citep{Rasmussen05} are a rare exception where analytically tractable function-space inference is possible.
Sparse GPs may be viewed as variational inference over functions \citep{pmlr-v51-matthews16}, minimizing the fKL from its exact posterior via inducing points.

In functional variational BNNs (fBNNs), \cite{Sun19} derive the fKL as a supremum over an infinite set of finite, marginal KL divergences.
\cite{burt2021understanding} showed that this fKL can be infinite under certain conditions, for example when considering the divergence between two BNNs with different architectures.
\cite{Sun19} replace the intractable supremum by an expectation based on finite measurement sets.
We explain this in more detail in Section \ref{sec:fvi} because our approach is based on this approximation.
Further, \cite{Sun19} they used a trained GP as explicit function-space prior, which can be viewed as a form of empirical Bayes, and employ the spectral Stein gradient estimator (SSGE) \citep{shi2018spectral} to enable implicit function priors.
Similar approaches take a mirror descent view for batch training \citep{pmlr-v97-shi19a}.

Variational implicit processes \citep{pmlr-v97-ma19b} interpret
parametric models with stochastic parameters as stochastic processes and introduce a wake-sleep procedure for inference in the regression setting with Gaussian likelihoods.
Our generic view of Bayesian neural networks and other stochastic models can be formally understood within their stochastic process perspective of parametric models, although our inference approach is unrelated (see Appendix \ref{app:vip}).

Neural linear models have also been used with fVI, because closed-form Gaussian predictive distributions allow explicit computation of gradients \citep{watson2021bayesian,watson2021neural}.
Concurrent work \citep{rudner2021rethinking} has also adopted fVI for classification, by linearizing a neural network about a Gaussian weight distribution to estimate the fKL.
This model works with a Gaussian (latent) predictive prior and posterior which loses the intuitive aspect of function-space priors.
Moreover, the linearization requires computation of the Jacobian of the neural network function with respect to the model parameters, for which the memory requirement scales with the number of model parameters and outputs.
\cite{wang2018function} propose particle optimization methods using finite function representations to learn a particle representation of the function-space posterior through the gradient flow of the log posterior.
Function-space inference is also an attractive approach to continual learning \citep{pan2020continual}.

\paragraph{Prior Networks}
\cite{malinin2018predictive,NEURIPS2019_7dd2ae7d} use a neural network with parameters $\bm{\theta}$ to directly predict the concentration parameters $\alpha_c$ of a Dirichlet distribution $\mathrm{p}(\bm{\mu}|\bm{x}; \bm{\theta})$, given input $\bm{x}$, which is distinct from our approach of estimating a posterior Dirichlet from $M$ categorical predictions.
This model is not a Bayesian neural network in practice, as only point estimates for the weights are learned.
To ensure $\alpha_c > 0$, an element-wise exponential operation is applied as the final layer of the neural network.
Additionally, Prior Networks minimize an optimization objective consisting of two separate KL divergences, representing in- and out-of-distribution data respectively,
\begin{align}
\mathcal{L}(\bm{\theta}) =
\E_{\mathrm{p_{in}}(\bm{x})}
[\KL[\mathrm{Dir}(\bm{\mu}|\hat{\bm{\alpha}})\mid\mid\mathrm{p}(\bm{\mu}|\bm{x}; \bm{\theta})]]
+ \E_{\mathrm{p_{out}}(\bm{x})}[\KL[\mathrm{Dir}(\bm{\mu}|\tilde{\bm{\alpha}})\mid\mid\mathrm{p}(\bm{\mu}|\bm{x}; \bm{\theta})]].
\end{align}
The first expectation $\E_{\mathrm{p_{in}}}$ accounts for the actual learning, i.e. fitting the training data, whereas the second expectation $\E_{\mathrm{p_{out}}}$ is supposed to regularize the model by matching a prior distribution.
Accordingly, the first expectation is computed for the training data and can be compared to the expected log-likelihood term in our approach.
Instead of maximizing the categorical log-likelihood of $M$ observations, Prior Networks construct Dirichlet targets by smoothing categorical ground truth labels to define the Dirichlet mean and setting the precision as a hyperparameter during training.
Although we also apply `label smoothing' to the predictions, it is for numerical reasons and not for the construction of target distributions from labels.
Additionally, Prior Networks treat the precision of their constructed \emph{target} distribution as a hyperparameter, whereas we estimate the Dirichlet precision of our \emph{predicted} variational posterior distribution via maximum likelihood.
The second expectation is computed for OOD data and resembles the fKL term in our approach, where the OOD data is used as measurement set.
In contrast to Prior Networks, our more general approach also allows the training data or mixtures of training data and OOD data as measurement sets, whereas Prior Networks explicitly compute their second expectation for OOD data only.
Furthermore, both Prior Network expectations consider the KL divergence from the neural network predictive distribution (right) to the target or prior distribution (left), whereas, in our approach, and variational inference in general, the KL divergence from the prior distribution (right) to the variational posterior (left) is considered.

\paragraph{Belief Matching}
\cite{joo2020being} assume a Dirichlet prior which, together with the categorical ground truth class labels, define a target Dirichlet posterior.
A neural network is used to directly predict concentration parameters of a Dirichlet posterior $q_{\vz|\vx}$ by replacing the final softmax layer with an element-wise exponential operation.
To learn the target posterior, belief matching maximizes
\begin{align}
l_{EB}(\vy, \alpha^{\mW}(\vx))
= \E_{q_{\vz|\vx}} [\log \vz_{\vy}]
- \KL[q_{\vz|\vx}^{\mW}\mid\mid p_{\vz|\vx}],
\end{align}
where $\E_{q_{\vz|\vx}} [\log \vz_{\vy}]$ is the expected log-likelihood of the training data and $\KL{q_{\vz|\vx}^{\mW}}{p_{\vz|\vx}}$ is the KL divergence between the predicted Dirichlet posterior and the Dirichlet prior.
Therefore, their objective matches our fELBO objective (Equation (\ref{eq:felbo})) except for two differences:
Firstly, belief matching computes both the expected log-likelihood and the KL divergence with respect to their single, directly predicted Dirichlet distribution, whereas we evaluate them as arithmetic averages of $M$ stochastic categorical model outputs.
Secondly, belief matching does not recognize the function-space aspect and instead only considers evaluation of the KL divergence using the training data, which resembles the fKL in our case where the measurement set is constrained to be the training data.

\paragraph{Evidential Deep Learning}
\cite{sensoy2018evidential} directly predict the concentration parameter of a Dirichlet distribution by using a neural network with ReLU activations as final layer to assert the positive constraint.
Additionally, a loss function is derived via type-II maximum likelihood by integrating over a Dirichlet prior and the sum of squares between target labels $\vy_i$ and predicted probabilities $\vp_i$.
Furthermore, a regularizing KL divergence term is added, resulting in a total loss function,
\begin{align}
    \gL(\Theta) =
    \sum_{i=1}^N \left(
    \sum_{j=1}^K \left( (y_{ij} - \hat{p}_{ij})^2 + \frac{\hat{p}_{ij}(1 - \hat{p}_{ij})}{S_i + 1} \right)
    + \lambda_t \KL[\mathrm{Dir}(\vp_i| \tilde{\bm{\alpha}}_i)\mid\mid \mathrm{Dir}(\vp_i| \mathbf{1})]
    \right),
\end{align}
where $y_{ij}$ are individual 0-1 target labels, $\hat{p}_{ij}$ are components of the predicted Dirichlet mean, $S_i$ is the predicted Dirichlet precision, $\tilde{\bm{\alpha}}_i$ is the predicted Dirichlet concentration parameter, $\mathbf{1}$ is a vector of ones and $\lambda_t$ is a annealing coefficient for optimization.
The first part of their loss is responsible for fitting the training data and can thus be compared to the maximum likelihood objective in Section \ref{sec:classification}.
The ML objective can be derived from the categorical log-likelihood via type-I maximum likelihood, whereas their objective is derived by minimizing the sum of squares via type-II maximum likelihood.
The second part of their loss resembles the fKL in our approach.
However, they only evaluate the KL divergence for the training data and explicitly consider the uniform Dirichlet distribution with concentration $\mathbf{1}$.
Therefore, their KL divergence regularization term is a special case of our proposed regularization with the measurement set being the training data and the prior being the uniform Dirichlet distribution with precision $K$.
For both parts, a major difference between evidential deep learning and our approach is the realization of the predictive Dirichlet distribution.
Evidential deep learning directly predicts Dirichlet concentration parameters, whereas we use $M$ predictions to estimate a Dirichlet distribution via maximum likelihood.

\paragraph{Experimental Comparison}
We compare our MAP and MAP fVI models to Belief Matching and Prior Networks, which both demonstrated scalability to ResNet models.
To reproduce their results, we used the official open-source implementations
\footnote{
\url{github.com/tjoo512/belief-matching-framework}}
\footnote{
\url{github.com/KaosEngineer/PriorNetworks}}.

\begin{figure}[!htb]
    \centering
    \resizebox{\textwidth}{!}{%
\begin{tikzpicture}

\definecolor{color0}{rgb}{0.9,0.6,0}
\definecolor{color1}{rgb}{0,0.6,0.5}
\definecolor{color2}{rgb}{0.35,0.7,0.9}
\definecolor{color3}{rgb}{0.95,0.9,0.25}

\begin{axis}[
height=4cm,
tick align=outside,
tick pos=left,
width=4cm,
x grid style={white!69.0196078431373!black},
xlabel={\footnotesize{Corruption}},
xmin=0, xmax=5,
xtick={0, 1, 2, 3, 4, 5},
xtick style={color=black},
y grid style={white!69.0196078431373!black},
ylabel={\footnotesize{Accuracy}},
ymin=40, ymax=100,
ytick style={color=black}
]
\path [draw=color0, fill=color0, opacity=0.1]
(axis cs:0,94.6108142283478)
--(axis cs:0,94.0271862599334)
--(axis cs:1,87.1499125311944)
--(axis cs:2,81.1647710596095)
--(axis cs:3,75.1365252202497)
--(axis cs:4,67.6240329557773)
--(axis cs:5,55.9403220110066)
--(axis cs:5,58.6249398298138)
--(axis cs:5,58.6249398298138)
--(axis cs:4,70.1027050202969)
--(axis cs:3,76.8085278803362)
--(axis cs:2,82.362386533164)
--(axis cs:1,88.1480870415595)
--(axis cs:0,94.6108142283478)
--cycle;

\path [draw=color0, fill=color0, opacity=0.1]
(axis cs:0,94.8836985198507)
--(axis cs:0,93.912302151536)
--(axis cs:1,87.3048083212329)
--(axis cs:2,81.0045292324371)
--(axis cs:3,74.795365259401)
--(axis cs:4,67.6787951253886)
--(axis cs:5,55.9341927893019)
--(axis cs:5,58.1822287194871)
--(axis cs:5,58.1822287194871)
--(axis cs:4,69.8678357339864)
--(axis cs:3,76.8227392938217)
--(axis cs:2,82.3156810336762)
--(axis cs:1,88.0214032266186)
--(axis cs:0,94.8836985198507)
--cycle;

\path [draw=color1, fill=color1, opacity=0.1]
(axis cs:0,95.4378262637647)
--(axis cs:0,95.0881747127978)
--(axis cs:1,89.2225227040233)
--(axis cs:2,83.5129903237881)
--(axis cs:3,77.9655794179121)
--(axis cs:4,70.6899097034003)
--(axis cs:5,58.7146494840801)
--(axis cs:5,60.7366139436543)
--(axis cs:5,60.7366139436543)
--(axis cs:4,72.1529308727716)
--(axis cs:3,79.0165235483964)
--(axis cs:2,84.3586954672275)
--(axis cs:1,89.6645317393361)
--(axis cs:0,95.4378262637647)
--cycle;

\path [draw=color2, fill=color2, opacity=0.1]
(axis cs:0,94.5896)
--(axis cs:0,94.4504)
--(axis cs:1,86.7347)
--(axis cs:2,79.9586)
--(axis cs:3,73.0373)
--(axis cs:4,65.0435)
--(axis cs:5,52.9478)
--(axis cs:5,54.3942)
--(axis cs:5,54.3942)
--(axis cs:4,66.4399)
--(axis cs:3,74.2041)
--(axis cs:2,80.8142)
--(axis cs:1,87.2279)
--(axis cs:0,94.5896)
--cycle;

\path [draw=color3, fill=color3, opacity=0.1]
(axis cs:0,67.8619)
--(axis cs:0,65.4403)
--(axis cs:1,60.4201)
--(axis cs:2,57.1221)
--(axis cs:3,54.0038)
--(axis cs:4,49.5801)
--(axis cs:5,41.667)
--(axis cs:5,42.9098)
--(axis cs:5,42.9098)
--(axis cs:4,50.9085)
--(axis cs:3,55.4074)
--(axis cs:2,58.6089)
--(axis cs:1,62.1301)
--(axis cs:0,67.8619)
--cycle;

\addplot [semithick, color0, dashed]
table {%
0 94.3190002441406
1 87.6489997863769
2 81.7635787963867
3 75.972526550293
4 68.8633689880371
5 57.2826309204102
};
\addplot [semithick, color0]
table {%
0 94.3980003356934
1 87.6631057739258
2 81.6601051330566
3 75.8090522766113
4 68.7733154296875
5 57.0582107543945
};
\addplot [semithick, color1]
table {%
0 95.2630004882812
1 89.4435272216797
2 83.9358428955078
3 78.4910514831543
4 71.4214202880859
5 59.7256317138672
};
\addplot [semithick, color2]
table {%
0 94.52
1 86.9813
2 80.3864
3 73.6207
4 65.7417
5 53.671
};
\addplot [semithick, color3]
table {%
0 66.6511
1 61.2751
2 57.8655
3 54.7056
4 50.2443
5 42.2884
};
\end{axis}

\end{tikzpicture}
\begin{tikzpicture}

\definecolor{color0}{rgb}{0.9,0.6,0}
\definecolor{color1}{rgb}{0,0.6,0.5}
\definecolor{color2}{rgb}{0.35,0.7,0.9}
\definecolor{color3}{rgb}{0.95,0.9,0.25}

\begin{axis}[
height=4cm,
tick align=outside,
tick pos=left,
width=4cm,
x grid style={white!69.0196078431373!black},
xlabel={\footnotesize{Corruption}},
xmin=0, xmax=5,
xtick={0, 1, 2, 3, 4, 5},
xtick style={color=black},
y grid style={white!69.0196078431373!black},
ylabel={\footnotesize{Log-Likelihood}},
ymin=-2.5, ymax=0,
ytick style={color=black}
]
\path [draw=color0, fill=color0, opacity=0.1]
(axis cs:0,-0.207512508560266)
--(axis cs:0,-0.225273216431343)
--(axis cs:1,-0.545712976932892)
--(axis cs:2,-0.837752319252921)
--(axis cs:3,-1.1656004164273)
--(axis cs:4,-1.61821419492043)
--(axis cs:5,-2.35496679014621)
--(axis cs:5,-2.05334109315328)
--(axis cs:5,-2.05334109315328)
--(axis cs:4,-1.42855392200694)
--(axis cs:3,-1.06879324458183)
--(axis cs:2,-0.763911517049252)
--(axis cs:1,-0.492924841348524)
--(axis cs:0,-0.207512508560266)
--cycle;

\path [draw=color0, fill=color0, opacity=0.1]
(axis cs:0,-0.236197830703572)
--(axis cs:0,-0.258495559744928)
--(axis cs:1,-0.491717658395984)
--(axis cs:2,-0.714165072099431)
--(axis cs:3,-0.943246233352625)
--(axis cs:4,-1.20550087215164)
--(axis cs:5,-1.65501593998642)
--(axis cs:5,-1.53897848866524)
--(axis cs:5,-1.53897848866524)
--(axis cs:4,-1.11016769182749)
--(axis cs:3,-0.857749877368793)
--(axis cs:2,-0.65873862095508)
--(axis cs:1,-0.461712108115287)
--(axis cs:0,-0.236197830703572)
--cycle;

\path [draw=color1, fill=color1, opacity=0.1]
(axis cs:0,-0.206886588331722)
--(axis cs:0,-0.212307902352788)
--(axis cs:1,-0.38964636386453)
--(axis cs:2,-0.55976897940216)
--(axis cs:3,-0.736341258587666)
--(axis cs:4,-0.966342597547088)
--(axis cs:5,-1.36960624558302)
--(axis cs:5,-1.28432668813712)
--(axis cs:5,-1.28432668813712)
--(axis cs:4,-0.916578072388488)
--(axis cs:3,-0.70123949652235)
--(axis cs:2,-0.534803193468391)
--(axis cs:1,-0.377061271002886)
--(axis cs:0,-0.206886588331722)
--cycle;

\path [draw=color2, fill=color2, opacity=0.1]
(axis cs:0,-0.256)
--(axis cs:0,-0.2596)
--(axis cs:1,-0.5138)
--(axis cs:2,-0.746)
--(axis cs:3,-0.9941)
--(axis cs:4,-1.2857)
--(axis cs:5,-1.7391)
--(axis cs:5,-1.6691)
--(axis cs:5,-1.6691)
--(axis cs:4,-1.2265)
--(axis cs:3,-0.9485)
--(axis cs:2,-0.7148)
--(axis cs:1,-0.497)
--(axis cs:0,-0.256)
--cycle;

\path [draw=color3, fill=color3, opacity=0.1]
(axis cs:0,-1.4143)
--(axis cs:0,-1.4795)
--(axis cs:1,-1.6111)
--(axis cs:2,-1.6832)
--(axis cs:3,-1.7557)
--(axis cs:4,-1.8703)
--(axis cs:5,-2.0795)
--(axis cs:5,-2.0407)
--(axis cs:5,-2.0407)
--(axis cs:4,-1.8311)
--(axis cs:3,-1.7137)
--(axis cs:2,-1.6412)
--(axis cs:1,-1.5659)
--(axis cs:0,-1.4143)
--cycle;

\addplot [semithick, color0, dashed]
table {%
0 -0.216392862495805
1 -0.519318909140708
2 -0.800831918151086
3 -1.11719683050457
4 -1.52338405846368
5 -2.20415394164974
};
\addplot [semithick, color0]
table {%
0 -0.24734669522425
1 -0.476714883255635
2 -0.686451846527256
3 -0.900498055360709
4 -1.15783428198956
5 -1.59699721432583
};
\addplot [semithick, color1]
table {%
0 -0.209597245342255
1 -0.383353817433708
2 -0.547286086435276
3 -0.718790377555008
4 -0.941460334967788
5 -1.32696646686007
};
\addplot [semithick, color2]
table {%
0 -0.2578
1 -0.5054
2 -0.7304
3 -0.9713
4 -1.2561
5 -1.7041
};
\addplot [semithick, color3]
table {%
0 -1.4469
1 -1.5885
2 -1.6622
3 -1.7347
4 -1.8507
5 -2.0601
};
\end{axis}

\end{tikzpicture}
\begin{tikzpicture}

\definecolor{color0}{rgb}{0.9,0.6,0}
\definecolor{color1}{rgb}{0,0.6,0.5}
\definecolor{color2}{rgb}{0.35,0.7,0.9}
\definecolor{color3}{rgb}{0.95,0.9,0.25}

\begin{axis}[
height=4cm,
tick align=outside,
tick pos=left,
width=4cm,
x grid style={white!69.0196078431373!black},
xlabel={\footnotesize{Corruption}},
xmin=0, xmax=5,
xtick={0, 1, 2, 3, 4, 5},
xtick style={color=black},
y grid style={white!69.0196078431373!black},
ylabel={\footnotesize{ECE}},
ymin=0, ymax=0.4,
ytick style={color=black}
]
\path [draw=color0, fill=color0, opacity=0.1]
(axis cs:0,0.0343220004528229)
--(axis cs:0,0.0297373945177133)
--(axis cs:1,0.0724895053470303)
--(axis cs:2,0.112814383131984)
--(axis cs:3,0.154403910282164)
--(axis cs:4,0.202827122069466)
--(axis cs:5,0.286221170901763)
--(axis cs:5,0.322901939869416)
--(axis cs:5,0.322901939869416)
--(axis cs:4,0.228432352088821)
--(axis cs:3,0.167573341485948)
--(axis cs:2,0.121767602937695)
--(axis cs:1,0.079622738722546)
--(axis cs:0,0.0343220004528229)
--cycle;

\path [draw=color0, fill=color0, opacity=0.1]
(axis cs:0,0.0554853002754069)
--(axis cs:0,0.0509061557027482)
--(axis cs:1,0.0461728717224877)
--(axis cs:2,0.0448804724709837)
--(axis cs:3,0.0769709915731252)
--(axis cs:4,0.126311418409398)
--(axis cs:5,0.207869880439079)
--(axis cs:5,0.238986418603622)
--(axis cs:5,0.238986418603622)
--(axis cs:4,0.147237476711223)
--(axis cs:3,0.095587617078036)
--(axis cs:2,0.0505043264730127)
--(axis cs:1,0.0517316167218406)
--(axis cs:0,0.0554853002754069)
--cycle;

\path [draw=color1, fill=color1, opacity=0.1]
(axis cs:0,0.068702558187816)
--(axis cs:0,0.0657178560811545)
--(axis cs:1,0.0487115128635035)
--(axis cs:2,0.0291465684364354)
--(axis cs:3,0.0350029452435855)
--(axis cs:4,0.0524082939152871)
--(axis cs:5,0.0957724233183993)
--(axis cs:5,0.120057631560789)
--(axis cs:5,0.120057631560789)
--(axis cs:4,0.0650432211334552)
--(axis cs:3,0.0432162651128884)
--(axis cs:2,0.0324699443240846)
--(axis cs:1,0.0516200908721819)
--(axis cs:0,0.068702558187816)
--cycle;

\path [draw=color2, fill=color2, opacity=0.1]
(axis cs:0,0.0667)
--(axis cs:0,0.0647)
--(axis cs:1,0.0643)
--(axis cs:2,0.0663)
--(axis cs:3,0.082)
--(axis cs:4,0.1378)
--(axis cs:5,0.2271)
--(axis cs:5,0.2443)
--(axis cs:5,0.2443)
--(axis cs:4,0.1522)
--(axis cs:3,0.0928)
--(axis cs:2,0.0691)
--(axis cs:1,0.0663)
--(axis cs:0,0.0667)
--cycle;

\path [draw=color3, fill=color3, opacity=0.1]
(axis cs:0,0.2091)
--(axis cs:0,0.1955)
--(axis cs:1,0.2064)
--(axis cs:2,0.2083)
--(axis cs:3,0.2126)
--(axis cs:4,0.223)
--(axis cs:5,0.2399)
--(axis cs:5,0.2483)
--(axis cs:5,0.2483)
--(axis cs:4,0.2318)
--(axis cs:3,0.223)
--(axis cs:2,0.2195)
--(axis cs:1,0.2192)
--(axis cs:0,0.2091)
--cycle;

\addplot [semithick, color0, dashed]
table {%
0 0.0320296974852681
1 0.0760561220347881
2 0.11729099303484
3 0.160988625884056
4 0.215629737079144
5 0.30456155538559
};
\addplot [semithick, color0]
table {%
0 0.0531957279890776
1 0.0489522442221642
2 0.0476923994719982
3 0.0862793043255806
4 0.13677444756031
5 0.223428149521351
};
\addplot [semithick, color1]
table {%
0 0.0672102071344853
1 0.0501658018678427
2 0.03080825638026
3 0.039109605178237
4 0.0587257575243711
5 0.107915027439594
};
\addplot [semithick, color2]
table {%
0 0.0657
1 0.0653
2 0.0677
3 0.0874
4 0.145
5 0.2357
};
\addplot [semithick, color3]
table {%
0 0.2023
1 0.2128
2 0.2139
3 0.2178
4 0.2274
5 0.2441
};
\end{axis}

\end{tikzpicture}
        \raisebox{1.2cm}{\begin{tikzpicture}

\newenvironment{customlegend}[1][]{%
    \begingroup
    \csname pgfplots@init@cleared@structures\endcsname
    \pgfplotsset{#1}%
}{%
    \csname pgfplots@createlegend\endcsname
    \endgroup
}%

\def\addlegendimage{\csname pgfplots@addlegendimage\endcsname}

\definecolor{color0}{rgb}{.9, .6, 0.}
\definecolor{color1}{rgb}{0., .6, .5}
\definecolor{color2}{rgb}{.35, .7, .9}
\definecolor{color3}{rgb}{.95, .9, .25}

\begin{customlegend}[legend entries={MAP, MAP fVI, Ensemble fVI, Belief Matching, Prior Networks}, legend columns={1}, legend cell align=left, legend style={draw=none, font=\footnotesize, column sep=.2cm}]
\addlegendimage{color0,semithick,dashed}
\addlegendimage{color0,semithick}
\addlegendimage{color1,semithick}
\addlegendimage{color2,semithick}
\addlegendimage{color3,semithick}
\end{customlegend}
\end{tikzpicture}}}
    \caption{Comparing Prior Networks and Belief Matching against our MAP, MAP fVI and Ensemble fVI models on the corrupted CIFAR10 task.
    MAP fVI and Belief Matching achieve comparable performance, while our best model, Ensemble fVI, outperforms Belief Matching by a considerable amount.}
    \label{fig:extra_baselines}
\end{figure}
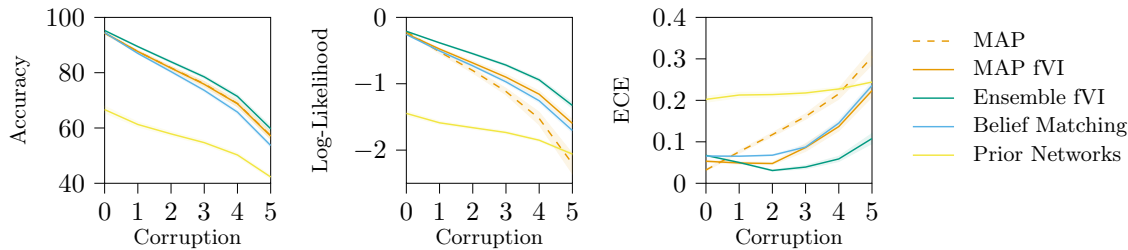

Figure~\ref{fig:extra_baselines} illustrates the test accuracy, log-likelihood, and expected calibration error for the corrupted CIFAR10 image classification task (see Section~\ref{sec:corrupt}).
We trained the models using the same procedure and hyperparameters described in Section \ref{app:implementation_details}.
The Belief Matching model corresponds closely to the MAP fVI model, as both the objectives and models are similar.
Unfortunately, we were not able to reproduce the Prior Networks performance described in the paper \citep{NEURIPS2019_7dd2ae7d}, neither with their listed hyperparameters (Table 4, \citep{NEURIPS2019_7dd2ae7d}) or the hyperparameters used in Appendix \ref{app:implementation_details}.
It is uncertain whether this is due to the model, implementation bugs or unrecorded hyperparameters
\footnote{The authors did not respond to personal correspondence regarding this matter.}.
\newpage

\section{Deep Stochastic Classifiers as Implicit Stochastic Processes}
\label{app:vip}
An implicit stochastic process \citep{pmlr-v97-ma19b} is an infinite set of random variables $\vf$, such that any finite subset
$\vf_{\vx_{1:L}} = \{\vf_{\vx_1}, \vf_{\vx_2}, ..., \vf_{\vx_L}\}$ with $L \in \mathbb{N}$ has a joint distribution which is implicitly defined as
\begin{align*}
    \vw \sim p(\vw), & & \vf_{\vx_l} = \vg(\vx_l, \vw), & & \forall\,\vx_l\in \gX, & & 1 \leq l \leq L,
\end{align*}
where the classifiers which we consider, such as BNNs and other stochastic neural networks, are instantiated as a feedforward or convolutional neural networks with stochastic weights, such that
$\vg(\vx_l, \vw) = \bm{\sigma}\left(\bm{\phi}\left(\vx; \vw\right)\right)$ in our case.
In practice, the implicit stochastic process interpretation of BNNs and stochastic models entails that we consider the weight distribution $p(\vw)$ in a parameterized form $q_\vtheta(\vw)$ with parameters $\vtheta$ which we wish to optimize.
The actual form and meaning of $\vtheta$ depends on the specific neural network architecture, encoded through $\bm{\phi}$.
Table~\ref{tab:vip_param} lists mathematical expressions to describe $\vtheta$ for various models.
Different model-specific parameterizations $q_\vtheta(\vw)$ induce the same generic variational posterior over functions $q(\vf |\vtheta)$ which allows us to implement function-space regularization independent of the specific model.
\begin{table}[b!]
      \centering
       \caption{Summary of stochastic (and deterministic) representations of weights $\vw$ which correspond to popular BNN approaches, applying either to the whole network or per layer.
       Note $\mSigma_\theta$ is typically factorized in practice.
       While we use Radial BNNs rather than MFVI, we include it for completeness.
       Rank-1 references specifically the Gaussian realization.}
       \label{tab:vip_param}
       {%
             \begin{tabular}{lll}
             \toprule
              Model & Parameterization for $\vw \sim q_\vtheta(\cdot)$ & Scope\\ \midrule
		      MAP        &     $\delta(\vw-\vw_\vtheta)$ & Network             \\
		      MFVI   &     $\gN(\vmu_\vtheta, \mSigma_\vtheta)$ & Layer   \\
		      Radial   &       $\vw \sim \vmu_\theta + \sqrt{\mSigma_\theta}
		      \odot \hat{\bm{\epsilon}} \cdot |r|,\;
		      \hat{\bm{\epsilon}} = \bm{\epsilon} / |\bm{\epsilon}|, \;
		      \bm{\epsilon} \sim \gN(\bm{0}, \mI), \; r \sim \gN(0, 1) $ & Layer  \\
		      MC Dropout &  $\vw \sim \bar{\vw}_\vtheta \odot \vb,\;\; b_i \sim \text{Bernoulli}(p_\text{dropout}$) & Layer      \\
		      Ensemble   &     $\frac{1}{M}\sum_m\delta(\vw{-}\vw^m_\vtheta)$ & Network \\
		      Rank-1  &       $\frac{1}{M}\sum_m\delta(\vw{-}\vw^m_\vtheta),\;\vw^m_\vtheta \sim \bar{\vw}_\vtheta \odot \delta\vw^m_\vtheta,\;\delta\vw^m_\vtheta \sim \gN(\vmu^m_\theta, {\vv^m_\theta}{\vu^m_\theta}^\top)$  & Layer  \\
		      \bottomrule     
		      \end{tabular}
		}
\end{table}

Formally, the stochastic process $\vf$ is defined on the sample space
$\Omega$
with an index set $\gX$ defined by the data type, such that
$\vf : \gX \times \Omega \rightarrow \Delta^{K-1}$, where
$\Delta^{K-1}$ is the state space, which is the $K{-}1$ simplex.
A random variable $\vf(\vx): \Omega \rightarrow \Delta^{K-1}$ can be defined for each $\vx\in\gX$ and we write $\vf(\vx) = \vf_{\vx}$.
Kolmogorov's extension theorem \citep{tao2013introduction} guarantees the existence of a stochastic process $\vf$ if for each $L \in \mathbb{N}$ the finite marginal joint distributions $p_{1:L}(\vf_{\vx_{1:L}})$, where $\vf_{\vx_{1:L}} = \{\vf_{\vx_1}, \dots, \vf_{\vx_L}\}$, satisfy exchangeability and consistency.
\paragraph{Exchangeability}
For any permutation $\pi$ of $1,\dots,L$, $p_{\pi(\vx_{1:L})}(\vf_{\pi(x_{1:L})}) = p_{\vx_{1:L}}(\vf_{\vx_{1:L}})$.
This requires that the process behavior is invariant to the order of inputs. For a feedforward neural network, this is satisfied because the respective predictions do not change if the order of inputs changes.
\paragraph{Consistency}
For any $1\leq L' < L$, $p_{{1:L'}}(\vf_{\vx_{1:L'}}) =\int p_{{1:L}}(\vf_{\vx_{1:L}}) \; \mathrm{d}\vf_{\vx_{L'+1:L}}$.
This requires that future evaluations are independent of past evaluations. For a feedforward neural network, this is satisfied because predictions do not depend on previous predictions.

\section{Optimization and Prior Specification}
\label{sec:opt_and_prior_spec}
Here, we provide additional methodological details which were omitted in Section~\ref{sec:dirichlet_function_priors}.

\paragraph{Optimization}
We optimize $\bm{\theta}$ using backpropagation on the fELBO objective.
For some models, such as deep ensembles, this is standard gradient descent, while for others, such as MFVI, the reparameterization trick is required.
In cases where only a single sample is available ($M{\,=\,}1$), such as MAP models or MC Dropout with single forward pass, we set the precision $z_\vx$ to the size of the training data.
When computing gradients, we assume that $\bm{\alpha}_\vx$ does not depend on $\bm{\theta}$.
This serves the practical purpose of pruning the Dirichlet MLE from the computation graph, speeding up computation and evoking expectation maximization-style inference.
In Appendix \ref{sec:gradient_estimation}, we connect this approximation to the pathwise gradient estimator \citep{roeder2017sticking}, which can in fact be lower variance than the total gradient and lead to faster convergence in terms of computation time.
In terms of mini-batching, like \cite{Sun19}, we divide both the batched expected log-likelihood and the fKL by the mini-batch size for numerical stability.
Consequently, the fKL weight in the total ELBO depends on the mini-batch size, which is theoretically undesirable, but, in practice, KL divergence scaling in (weight-space) variational inference is a topic of active debate \citep{wenzel2020good,aitchison2021a} and is frequently scaled or annealed for numerical reasons \citep{dusenberry2020efficient}.

\paragraph{Prior Specification}
We require the Dirichlet function prior $p(\vf)$ to be defined as a regular $K$-dimensional Dirichlet distribution $p(\vf_\vx)$ at each input location $\vx$.
For most experiments in this paper, we choose $p(\vf_\vx) = \text{Dir}(\cdot\mid\bm{\beta})$ with $\bm{\beta} = \mathbf{1}$, which is a constant uniform Dirichlet distribution with precision $K$.
One might criticize that this constant uniform prior is factorized and does not encode any correlations between input locations.
However, the posterior will still be correlated among input locations due to the neural network.
As we learn a variational posterior over functions by adapting the implicit neural network weights, the neural network function induces smoothness in the variational posterior despite the factorized prior.
A similar scenario arises in conventional weight-space variational inference with factorized Gaussian priors (MFVI):
The weights are also not correlated by the prior yet the neural network function enables learning.
In practice, it is often difficult to define correlated priors in domains with high-dimensional inputs, such as images.
In a toy problem, we show that it is also possible to use more sophisticated priors based on, for example, GPs or random forests.
Nonetheless, the constant uniform prior is simple, scalable, intuitive to understand, yet effective (see Section~\ref{sec:experiments}).

\section{Image Classification under Adversarial Attacks}
\label{sec:adversarial}
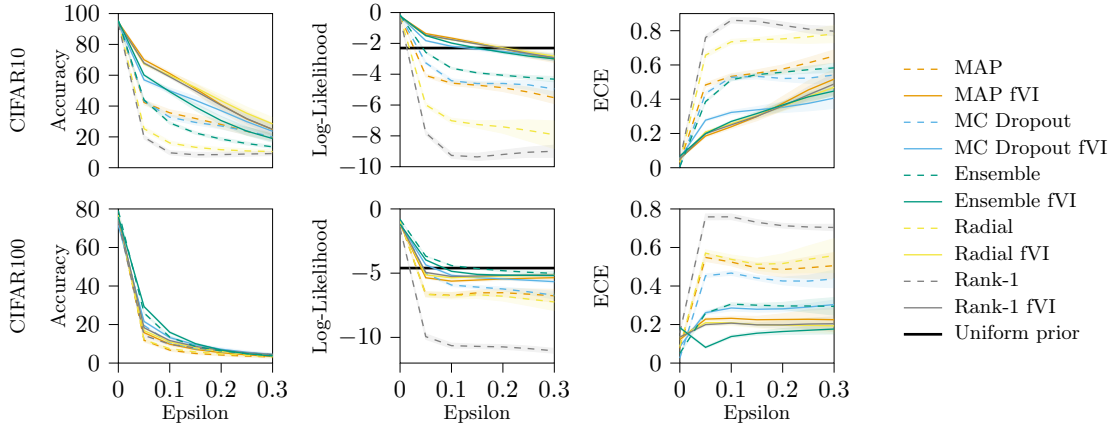
\begin{figure*}[hbt!]
    \centering
    \resizebox{\textwidth}{!}{
    \begin{tabular}{l@{\hspace{0cm}}l@{\hspace{0cm}}l}
\begin{tikzpicture}

\definecolor{color0}{rgb}{0.9,0.6,0}
\definecolor{color1}{rgb}{0.35,0.7,0.9}
\definecolor{color2}{rgb}{0.95,0.9,0.25}
\definecolor{color3}{rgb}{0,0.6,0.5}

\begin{axis}[
height=4cm,
tick align=outside,
tick pos=left,
width=4cm,
x grid style={white!69.0196078431373!black},
xtick=\empty,
xmin=0, xmax=0.3,
xtick style={color=black},
y grid style={white!69.0196078431373!black},
ylabel style={align=center},
ylabel={\footnotesize{CIFAR10}\\\bigskip\\\footnotesize{Accuracy}},
ymin=0, ymax=100,
ytick style={color=black}
]
\path [draw=color0, fill=color0, opacity=0.1]
(axis cs:0,94.6108142283478)
--(axis cs:0,94.0271862599334)
--(axis cs:0.05,40.9051583577401)
--(axis cs:0.1,33.632191786963)
--(axis cs:0.15,29.501391347833)
--(axis cs:0.2,25.4327904832884)
--(axis cs:0.25,21.0899453482987)
--(axis cs:0.3,17.4895282358629)
--(axis cs:0.3,21.8084720998304)
--(axis cs:0.3,21.8084720998304)
--(axis cs:0.25,26.1160551705001)
--(axis cs:0.2,30.6552092420534)
--(axis cs:0.15,34.3606089268252)
--(axis cs:0.1,37.5938084266601)
--(axis cs:0.05,43.7928412150138)
--(axis cs:0,94.6108142283478)
--cycle;

\path [draw=color0, fill=color0, opacity=0.1]
(axis cs:0,94.8836985198507)
--(axis cs:0,93.912302151536)
--(axis cs:0.05,68.9659925380813)
--(axis cs:0.1,59.8452164853179)
--(axis cs:0.15,48.3507295043377)
--(axis cs:0.2,36.5756097257126)
--(axis cs:0.25,26.7210990925316)
--(axis cs:0.3,20.799840849897)
--(axis cs:0.3,28.7901593026908)
--(axis cs:0.3,28.7901593026908)
--(axis cs:0.25,36.472901151609)
--(axis cs:0.2,45.144390732051)
--(axis cs:0.15,54.2792708008381)
--(axis cs:0.1,62.3587828432953)
--(axis cs:0.05,71.1160066684617)
--(axis cs:0,94.8836985198507)
--cycle;

\path [draw=color1, fill=color1, opacity=0.1]
(axis cs:0,94.4731262915458)
--(axis cs:0,94.1628722436105)
--(axis cs:0.05,42.3263296723931)
--(axis cs:0.1,31.2166033361677)
--(axis cs:0.15,27.8232600028486)
--(axis cs:0.2,25.1127218785078)
--(axis cs:0.25,21.4101912351945)
--(axis cs:0.3,17.2120546250459)
--(axis cs:0.3,22.7619455428008)
--(axis cs:0.3,22.7619455428008)
--(axis cs:0.25,26.2438088563582)
--(axis cs:0.2,29.0552781520098)
--(axis cs:0.15,30.9607403938799)
--(axis cs:0.1,34.2993970300432)
--(axis cs:0.05,44.2696694731148)
--(axis cs:0,94.4731262915458)
--cycle;

\path [draw=color1, fill=color1, opacity=0.1]
(axis cs:0,93.6896987627851)
--(axis cs:0,93.162301664461)
--(axis cs:0.05,55.9743309827544)
--(axis cs:0.1,48.0847760632935)
--(axis cs:0.15,41.2806297811766)
--(axis cs:0.2,33.370443925426)
--(axis cs:0.25,25.5411811596229)
--(axis cs:0.3,19.3457779978794)
--(axis cs:0.3,27.91022175798)
--(axis cs:0.3,27.91022175798)
--(axis cs:0.25,34.1768192523644)
--(axis cs:0.2,40.5635549454236)
--(axis cs:0.15,46.1733687844972)
--(axis cs:0.1,51.7312235399779)
--(axis cs:0.05,58.0116683153413)
--(axis cs:0,93.6896987627851)
--cycle;

\path [draw=color2, fill=color2, opacity=0.1]
(axis cs:0,95.303299582967)
--(axis cs:0,94.796700417033)
--(axis cs:0.05,24.0065870726729)
--(axis cs:0.1,14.8186965824624)
--(axis cs:0.15,12.2402371994809)
--(axis cs:0.2,10.2881440241081)
--(axis cs:0.25,9.62540225734679)
--(axis cs:0.3,9.27399906119351)
--(axis cs:0.3,11.6920009235477)
--(axis cs:0.3,11.6920009235477)
--(axis cs:0.25,12.1605976129535)
--(axis cs:0.2,13.091855899598)
--(axis cs:0.15,14.2637630828067)
--(axis cs:0.1,17.3213033793907)
--(axis cs:0.05,26.5754128968095)
--(axis cs:0,95.303299582967)
--cycle;

\path [draw=color2, fill=color2, opacity=0.1]
(axis cs:0,93.9336630742158)
--(axis cs:0,93.5163353999053)
--(axis cs:0.05,66.5336861912175)
--(axis cs:0.1,58.3294352490801)
--(axis cs:0.15,49.5380812371648)
--(axis cs:0.2,39.6815190477712)
--(axis cs:0.25,31.2844744484796)
--(axis cs:0.3,24.2774176626697)
--(axis cs:0.3,32.6065819711194)
--(axis cs:0.3,32.6065819711194)
--(axis cs:0.25,39.3535252768622)
--(axis cs:0.2,47.2444811658519)
--(axis cs:0.15,54.5999200140559)
--(axis cs:0.1,61.5125645678144)
--(axis cs:0.05,68.1163138087825)
--(axis cs:0,93.9336630742158)
--cycle;

\path [draw=gray, fill=gray, opacity=0.1]
(axis cs:0,94.0263689884237)
--(axis cs:0,93.3236310115763)
--(axis cs:0.05,18.6586484040118)
--(axis cs:0.1,8.42220640069692)
--(axis cs:0.15,6.81169721824008)
--(axis cs:0.2,7.0097769669465)
--(axis cs:0.25,7.42841188996775)
--(axis cs:0.3,7.87822723612207)
--(axis cs:0.3,10.4717727638779)
--(axis cs:0.3,10.4717727638779)
--(axis cs:0.25,10.3395880451824)
--(axis cs:0.2,10.0362227507659)
--(axis cs:0.15,9.99430272835416)
--(axis cs:0.1,10.9217936527088)
--(axis cs:0.05,20.9093520079755)
--(axis cs:0,94.0263689884237)
--cycle;

\path [draw=gray, fill=gray, opacity=0.1]
(axis cs:0,94.113341949524)
--(axis cs:0,93.5966601867065)
--(axis cs:0.05,67.0142845424174)
--(axis cs:0.1,58.1980824139985)
--(axis cs:0.15,48.3618230405845)
--(axis cs:0.2,38.1953561988214)
--(axis cs:0.25,28.6095260827909)
--(axis cs:0.3,21.2893108234427)
--(axis cs:0.3,28.4906894817331)
--(axis cs:0.3,28.4906894817331)
--(axis cs:0.25,35.0264732153048)
--(axis cs:0.2,42.760643557038)
--(axis cs:0.15,51.3921770509682)
--(axis cs:0.1,59.7739177080718)
--(axis cs:0.05,68.7697150913717)
--(axis cs:0,94.113341949524)
--cycle;

\path [draw=color3, fill=color3, opacity=0.1]
(axis cs:0,95.5419428413185)
--(axis cs:0,95.0480580742089)
--(axis cs:0.05,43.369743467331)
--(axis cs:0.1,28.2404645696941)
--(axis cs:0.15,21.7082554345806)
--(axis cs:0.2,17.6333804981246)
--(axis cs:0.25,14.7544503850405)
--(axis cs:0.3,12.5974929920741)
--(axis cs:0.3,14.4825073131017)
--(axis cs:0.3,14.4825073131017)
--(axis cs:0.25,16.6015499430235)
--(axis cs:0.2,19.5246193035111)
--(axis cs:0.15,23.3957438940327)
--(axis cs:0.1,29.8855348809895)
--(axis cs:0.05,44.6182562580108)
--(axis cs:0,95.5419428413185)
--cycle;

\path [draw=color3, fill=color3, opacity=0.1]
(axis cs:0,95.4378262637647)
--(axis cs:0,95.0881747127978)
--(axis cs:0.05,59.4977928128091)
--(axis cs:0.1,48.0315915546069)
--(axis cs:0.15,37.0941487884389)
--(axis cs:0.2,27.9347776499223)
--(axis cs:0.25,21.0121742494106)
--(axis cs:0.3,16.4491445815544)
--(axis cs:0.3,21.4008550369759)
--(axis cs:0.3,21.4008550369759)
--(axis cs:0.25,26.4558257811069)
--(axis cs:0.2,33.2372228536177)
--(axis cs:0.15,41.6658510589732)
--(axis cs:0.1,50.6564089336743)
--(axis cs:0.05,60.7482063326988)
--(axis cs:0,95.4378262637647)
--cycle;

\addplot [semithick, color0, dashed]
table {%
0 94.3190002441406
0.05 42.348999786377
0.1 35.6130001068115
0.15 31.9310001373291
0.2 28.0439998626709
0.25 23.6030002593994
0.3 19.6490001678467
};
\addplot [semithick, color0]
table {%
0 94.3980003356934
0.05 70.0409996032715
0.1 61.1019996643066
0.15 51.3150001525879
0.2 40.8600002288818
0.25 31.5970001220703
0.3 24.7950000762939
};
\addplot [semithick, color1, dashed]
table {%
0 94.3179992675781
0.05 43.2979995727539
0.1 32.7580001831055
0.15 29.3920001983643
0.2 27.0840000152588
0.25 23.8270000457764
0.3 19.9870000839233
};
\addplot [semithick, color1]
table {%
0 93.426000213623
0.05 56.9929996490479
0.1 49.9079998016357
0.15 43.7269992828369
0.2 36.9669994354248
0.25 29.8590002059937
0.3 23.6279998779297
};
\addplot [semithick, color2, dashed]
table {%
0 95.05
0.05 25.2909999847412
0.1 16.0699999809265
0.15 13.2520001411438
0.2 11.689999961853
0.25 10.8929999351501
0.3 10.4829999923706
};
\addplot [semithick, color2]
table {%
0 93.7249992370605
0.05 67.325
0.1 59.9209999084473
0.15 52.0690006256104
0.2 43.4630001068115
0.25 35.3189998626709
0.3 28.4419998168945
};
\addplot [semithick, gray, dashed]
table {%
0 93.675
0.05 19.7840002059937
0.1 9.67200002670288
0.15 8.40299997329712
0.2 8.5229998588562
0.25 8.88399996757507
0.3 9.175
};
\addplot [semithick, gray]
table {%
0 93.8550010681152
0.05 67.8919998168945
0.1 58.9860000610352
0.15 49.8770000457764
0.2 40.4779998779297
0.25 31.8179996490479
0.3 24.8900001525879
};
\addplot [semithick, color3, dashed]
table {%
0 95.2950004577637
0.05 43.9939998626709
0.1 29.0629997253418
0.15 22.5519996643066
0.2 18.5789999008179
0.25 15.678000164032
0.3 13.5400001525879
};
\addplot [semithick, color3]
table {%
0 95.2630004882812
0.05 60.1229995727539
0.1 49.3440002441406
0.15 39.3799999237061
0.2 30.58600025177
0.25 23.7340000152588
0.3 18.9249998092651
};
\end{axis}

\end{tikzpicture} &
\begin{tikzpicture}

\definecolor{color0}{rgb}{0.9,0.6,0}
\definecolor{color1}{rgb}{0.35,0.7,0.9}
\definecolor{color2}{rgb}{0.95,0.9,0.25}
\definecolor{color3}{rgb}{0,0.6,0.5}

\begin{axis}[
height=4cm,
tick align=outside,
tick pos=left,
width=4cm,
x grid style={white!69.0196078431373!black},
xtick=\empty,
xmin=0, xmax=0.3,
xtick style={color=black},
y grid style={white!69.0196078431373!black},
ylabel={\footnotesize{Log-Likelihood}},
ymin=-10, ymax=0,
ytick style={color=black}
]
\addplot [very thick, black]
table {%
0 -2.30258509299405
0.05 -2.30258509299405
0.1 -2.30258509299405
0.15 -2.30258509299405
0.2 -2.30258509299405
0.25 -2.30258509299405
0.3 -2.30258509299405
};
\path [draw=color0, fill=color0, opacity=0.1]
(axis cs:0,-0.207512508560266)
--(axis cs:0,-0.225273216431343)
--(axis cs:0.05,-4.17280242339622)
--(axis cs:0.1,-4.72131754029052)
--(axis cs:0.15,-4.89286293206323)
--(axis cs:0.2,-5.12132305323769)
--(axis cs:0.25,-5.47859009649374)
--(axis cs:0.3,-5.90704190571119)
--(axis cs:0.3,-5.14635745317656)
--(axis cs:0.3,-5.14635745317656)
--(axis cs:0.25,-4.84406814022099)
--(axis cs:0.2,-4.62631267900388)
--(axis cs:0.15,-4.53040317929397)
--(axis cs:0.1,-4.43763529022228)
--(axis cs:0.05,-3.99883903498889)
--(axis cs:0,-0.207512508560266)
--cycle;

\path [draw=color0, fill=color0, opacity=0.1]
(axis cs:0,-0.236197830703572)
--(axis cs:0,-0.258495559744928)
--(axis cs:0.05,-1.40787307311407)
--(axis cs:0.1,-1.70945569272697)
--(axis cs:0.15,-2.08130519711412)
--(axis cs:0.2,-2.52722122409255)
--(axis cs:0.25,-2.93764345288274)
--(axis cs:0.3,-3.24333010138393)
--(axis cs:0.3,-2.78117975664286)
--(axis cs:0.3,-2.78117975664286)
--(axis cs:0.25,-2.48520732657323)
--(axis cs:0.2,-2.16081914287404)
--(axis cs:0.15,-1.86768558659293)
--(axis cs:0.1,-1.60238741376573)
--(axis cs:0.05,-1.32171806272176)
--(axis cs:0,-0.236197830703572)
--cycle;

\path [draw=color1, fill=color1, opacity=0.1]
(axis cs:0,-0.166141660199167)
--(axis cs:0,-0.179173976275434)
--(axis cs:0.05,-3.34361517605955)
--(axis cs:0.1,-4.54631486102083)
--(axis cs:0.15,-4.73366766889482)
--(axis cs:0.2,-4.75555845161387)
--(axis cs:0.25,-4.98504552231609)
--(axis cs:0.3,-5.4132861965144)
--(axis cs:0.3,-4.5065149473605)
--(axis cs:0.3,-4.5065149473605)
--(axis cs:0.25,-4.39694911567655)
--(axis cs:0.2,-4.44283031287506)
--(axis cs:0.15,-4.47688434765885)
--(axis cs:0.1,-4.29738081862133)
--(axis cs:0.05,-3.18620332800994)
--(axis cs:0,-0.166141660199167)
--cycle;

\path [draw=color1, fill=color1, opacity=0.1]
(axis cs:0,-0.24524401764516)
--(axis cs:0,-0.25544842150254)
--(axis cs:0.05,-1.85861248089591)
--(axis cs:0.1,-2.26261441397275)
--(axis cs:0.15,-2.46403668935565)
--(axis cs:0.2,-2.64075701048647)
--(axis cs:0.25,-2.87110414407406)
--(axis cs:0.3,-3.12711925244282)
--(axis cs:0.3,-2.67197072745468)
--(axis cs:0.3,-2.67197072745468)
--(axis cs:0.25,-2.53840176526829)
--(axis cs:0.2,-2.40482850927462)
--(axis cs:0.15,-2.2701493085105)
--(axis cs:0.1,-2.11786746736053)
--(axis cs:0.05,-1.78869071436454)
--(axis cs:0,-0.24524401764516)
--cycle;

\path [draw=color2, fill=color2, opacity=0.1]
(axis cs:0,-0.205289010262891)
--(axis cs:0,-0.224209782980809)
--(axis cs:0.05,-6.11763347063336)
--(axis cs:0.1,-7.17945687582199)
--(axis cs:0.15,-7.49308290386622)
--(axis cs:0.2,-7.82882540329849)
--(axis cs:0.25,-8.33720532972593)
--(axis cs:0.3,-8.82011234712889)
--(axis cs:0.3,-7.031325053131)
--(axis cs:0.3,-7.031325053131)
--(axis cs:0.25,-6.96585749024107)
--(axis cs:0.2,-6.99883641238184)
--(axis cs:0.15,-7.02808263508935)
--(axis cs:0.1,-6.85778800231174)
--(axis cs:0.05,-5.8569868203999)
--(axis cs:0,-0.205289010262891)
--cycle;

\path [draw=color2, fill=color2, opacity=0.1]
(axis cs:0,-0.270104089830272)
--(axis cs:0,-0.281332139539057)
--(axis cs:0.05,-1.53463913780793)
--(axis cs:0.1,-1.81110597200379)
--(axis cs:0.15,-2.08995584638272)
--(axis cs:0.2,-2.41164170786956)
--(axis cs:0.25,-2.73794023823861)
--(axis cs:0.3,-3.0311527997853)
--(axis cs:0.3,-2.63164328138076)
--(axis cs:0.3,-2.63164328138076)
--(axis cs:0.25,-2.41011286815129)
--(axis cs:0.2,-2.16336495885998)
--(axis cs:0.15,-1.93036215658503)
--(axis cs:0.1,-1.7078482972002)
--(axis cs:0.05,-1.4652968712518)
--(axis cs:0,-0.270104089830272)
--cycle;

\path [draw=gray, fill=gray, opacity=0.1]
(axis cs:0,-0.31700722432758)
--(axis cs:0,-0.340314159559194)
--(axis cs:0.05,-7.93862960203612)
--(axis cs:0.1,-9.45339952141893)
--(axis cs:0.15,-9.59835583383632)
--(axis cs:0.2,-9.48539567091486)
--(axis cs:0.25,-9.38935491525003)
--(axis cs:0.3,-9.39431133258872)
--(axis cs:0.3,-8.60274150631344)
--(axis cs:0.3,-8.60274150631344)
--(axis cs:0.25,-8.73359109725091)
--(axis cs:0.2,-8.92778663662128)
--(axis cs:0.15,-9.12986605322066)
--(axis cs:0.1,-9.08486302489239)
--(axis cs:0.05,-7.67458692410058)
--(axis cs:0,-0.31700722432758)
--cycle;

\path [draw=gray, fill=gray, opacity=0.1]
(axis cs:0,-0.25447999022056)
--(axis cs:0,-0.27665711549816)
--(axis cs:0.05,-1.49208914729732)
--(axis cs:0.1,-1.78475659009744)
--(axis cs:0.15,-2.08345987457295)
--(axis cs:0.2,-2.44031261757386)
--(axis cs:0.25,-2.82627503610651)
--(axis cs:0.3,-3.15974832689612)
--(axis cs:0.3,-2.70651923206562)
--(axis cs:0.3,-2.70651923206562)
--(axis cs:0.25,-2.48482004151515)
--(axis cs:0.2,-2.24619874755029)
--(axis cs:0.15,-1.99231680639798)
--(axis cs:0.1,-1.72258154631441)
--(axis cs:0.05,-1.4202568946304)
--(axis cs:0,-0.25447999022056)
--cycle;

\path [draw=color3, fill=color3, opacity=0.1]
(axis cs:0,-0.143297958578635)
--(axis cs:0,-0.152798936238788)
--(axis cs:0.05,-2.62333376013297)
--(axis cs:0.1,-3.66375851682255)
--(axis cs:0.15,-3.97038252890382)
--(axis cs:0.2,-4.14769888356455)
--(axis cs:0.25,-4.3294788695326)
--(axis cs:0.3,-4.4931288227411)
--(axis cs:0.3,-4.15027248545463)
--(axis cs:0.3,-4.15027248545463)
--(axis cs:0.25,-4.08619789439954)
--(axis cs:0.2,-3.98643522419655)
--(axis cs:0.15,-3.85180929892082)
--(axis cs:0.1,-3.55758077120115)
--(axis cs:0.05,-2.53750558182255)
--(axis cs:0,-0.143297958578635)
--cycle;

\path [draw=color3, fill=color3, opacity=0.1]
(axis cs:0,-0.206886588365407)
--(axis cs:0,-0.212307902323363)
--(axis cs:0.05,-1.51641301620409)
--(axis cs:0.1,-2.02194863612416)
--(axis cs:0.15,-2.36624866237577)
--(axis cs:0.2,-2.67054427841055)
--(axis cs:0.25,-2.92776372290799)
--(axis cs:0.3,-3.12007537512172)
--(axis cs:0.3,-2.88573399447438)
--(axis cs:0.3,-2.88573399447438)
--(axis cs:0.25,-2.72192331432628)
--(axis cs:0.2,-2.50134687135921)
--(axis cs:0.15,-2.23648243355189)
--(axis cs:0.1,-1.93733398054145)
--(axis cs:0.05,-1.46966955435232)
--(axis cs:0,-0.206886588365407)
--cycle;

\addplot [semithick, color0, dashed]
table {%
0 -0.216392862495805
0.05 -4.08582072919255
0.1 -4.5794764152564
0.15 -4.7116330556786
0.2 -4.87381786612079
0.25 -5.16132911835736
0.3 -5.52669967944387
};
\addplot [semithick, color0]
table {%
0 -0.24734669522425
0.05 -1.36479556791792
0.1 -1.65592155324635
0.15 -1.97449539185352
0.2 -2.34402018348329
0.25 -2.71142538972798
0.3 -3.01225492901339
};
\addplot [semithick, color1, dashed]
table {%
0 -0.1726578182373
0.05 -3.26490925203474
0.1 -4.42184783982108
0.15 -4.60527600827684
0.2 -4.59919438224446
0.25 -4.69099731899632
0.3 -4.95990057193745
};
\addplot [semithick, color1]
table {%
0 -0.25034621957385
0.05 -1.82365159763022
0.1 -2.19024094066664
0.15 -2.36709299893308
0.2 -2.52279275988055
0.25 -2.70475295467117
0.3 -2.89954498994875
};
\addplot [semithick, color2, dashed]
table {%
0 -0.21474939662185
0.05 -5.98731014551663
0.1 -7.01862243906686
0.15 -7.26058276947778
0.2 -7.41383090784016
0.25 -7.6515314099835
0.3 -7.92571870012994
};
\addplot [semithick, color2]
table {%
0 -0.275718114684665
0.05 -1.49996800452987
0.1 -1.759477134602
0.15 -2.01015900148388
0.2 -2.28750333336477
0.25 -2.57402655319495
0.3 -2.83139804058303
};
\addplot [semithick, gray, dashed]
table {%
0 -0.328660691943387
0.05 -7.80660826306835
0.1 -9.26913127315566
0.15 -9.36411094352849
0.2 -9.20659115376807
0.25 -9.06147300625047
0.3 -8.99852641945108
};
\addplot [semithick, gray]
table {%
0 -0.26556855285936
0.05 -1.45617302096386
0.1 -1.75366906820593
0.15 -2.03788834048546
0.2 -2.34325568256208
0.25 -2.65554753881083
0.3 -2.93313377948087
};
\addplot [semithick, color3, dashed]
table {%
0 -0.148048447408711
0.05 -2.58041967097776
0.1 -3.61066964401185
0.15 -3.91109591391232
0.2 -4.06706705388055
0.25 -4.20783838196607
0.3 -4.32170065409787
};
\addplot [semithick, color3]
table {%
0 -0.209597245344385
0.05 -1.4930412852782
0.1 -1.9796413083328
0.15 -2.30136554796383
0.2 -2.58594557488488
0.25 -2.82484351861714
0.3 -3.00290468479805
};
\end{axis}

\end{tikzpicture} &
\begin{tikzpicture}

\definecolor{color0}{rgb}{0.9,0.6,0}
\definecolor{color1}{rgb}{0.35,0.7,0.9}
\definecolor{color2}{rgb}{0.95,0.9,0.25}
\definecolor{color3}{rgb}{0,0.6,0.5}

\begin{axis}[
height=4cm,
tick align=outside,
tick pos=left,
width=4cm,
x grid style={white!69.0196078431373!black},
xtick=\empty,
xmin=0, xmax=0.3,
xtick style={color=black},
y grid style={white!69.0196078431373!black},
ylabel={\footnotesize{ECE}},
ymin=0, ymax=0.9,
ytick style={color=black}
]
\path [draw=color0, fill=color0, opacity=0.1]
(axis cs:0,0.0343220004528229)
--(axis cs:0,0.0297373945177133)
--(axis cs:0.05,0.471783801150859)
--(axis cs:0.1,0.517434273923202)
--(axis cs:0.15,0.532905856661249)
--(axis cs:0.2,0.55182141882232)
--(axis cs:0.25,0.582674172316266)
--(axis cs:0.3,0.616196685075138)
--(axis cs:0.3,0.687954408884671)
--(axis cs:0.3,0.687954408884671)
--(axis cs:0.25,0.642551204766559)
--(axis cs:0.2,0.599928347050635)
--(axis cs:0.15,0.571710260816168)
--(axis cs:0.1,0.548979736601548)
--(axis cs:0.05,0.49493664877361)
--(axis cs:0,0.0343220004528229)
--cycle;

\path [draw=color0, fill=color0, opacity=0.1]
(axis cs:0,0.0554853002754069)
--(axis cs:0,0.0509061557027482)
--(axis cs:0.05,0.177243134364403)
--(axis cs:0.1,0.227600381631398)
--(axis cs:0.15,0.274529120972664)
--(axis cs:0.2,0.334653139030411)
--(axis cs:0.25,0.400905341735876)
--(axis cs:0.3,0.462129047916748)
--(axis cs:0.3,0.572486915780686)
--(axis cs:0.3,0.572486915780686)
--(axis cs:0.25,0.50767288125416)
--(axis cs:0.2,0.41615056999974)
--(axis cs:0.15,0.323560329625099)
--(axis cs:0.1,0.250528102260089)
--(axis cs:0.05,0.193299475327217)
--(axis cs:0,0.0554853002754069)
--cycle;

\path [draw=color1, fill=color1, opacity=0.1]
(axis cs:0,0.00988195434127554)
--(axis cs:0,0.00464262591760055)
--(axis cs:0.05,0.425583598447992)
--(axis cs:0.1,0.511196108839199)
--(axis cs:0.15,0.51851063158835)
--(axis cs:0.2,0.504011475091875)
--(axis cs:0.25,0.492561320852974)
--(axis cs:0.3,0.488957385133468)
--(axis cs:0.3,0.594270368505753)
--(axis cs:0.3,0.594270368505753)
--(axis cs:0.25,0.55322970907714)
--(axis cs:0.2,0.538464749807417)
--(axis cs:0.15,0.548522027465453)
--(axis cs:0.1,0.544925550439671)
--(axis cs:0.05,0.442937185930059)
--(axis cs:0,0.00988195434127554)
--cycle;

\path [draw=color1, fill=color1, opacity=0.1]
(axis cs:0,0.0599976294577991)
--(axis cs:0,0.0526324234067524)
--(axis cs:0.05,0.271132909505099)
--(axis cs:0.1,0.31274919615708)
--(axis cs:0.15,0.31909216511981)
--(axis cs:0.2,0.321890577806772)
--(axis cs:0.25,0.332243088244065)
--(axis cs:0.3,0.346259614136969)
--(axis cs:0.3,0.46683649191543)
--(axis cs:0.3,0.46683649191543)
--(axis cs:0.25,0.419228931905166)
--(axis cs:0.2,0.383896180854499)
--(axis cs:0.15,0.356719310395556)
--(axis cs:0.1,0.333360945120233)
--(axis cs:0.05,0.283871383221418)
--(axis cs:0,0.0599976294577991)
--cycle;

\path [draw=color2, fill=color2, opacity=0.1]
(axis cs:0,0.0340907307123518)
--(axis cs:0,0.0287865931386376)
--(axis cs:0.05,0.643448023108635)
--(axis cs:0.1,0.721646224322341)
--(axis cs:0.15,0.73452772744767)
--(axis cs:0.2,0.730147471433954)
--(axis cs:0.25,0.726245846143939)
--(axis cs:0.3,0.730572516109966)
--(axis cs:0.3,0.83230091891048)
--(axis cs:0.3,0.83230091891048)
--(axis cs:0.25,0.802442656167767)
--(axis cs:0.2,0.774646804326697)
--(axis cs:0.15,0.758793619788963)
--(axis cs:0.1,0.748807693657853)
--(axis cs:0.05,0.670880563946571)
--(axis cs:0,0.0340907307123518)
--cycle;

\path [draw=color2, fill=color2, opacity=0.1]
(axis cs:0,0.0542262308625221)
--(axis cs:0,0.0484405889781476)
--(axis cs:0.05,0.201534934581718)
--(axis cs:0.1,0.248497372354952)
--(axis cs:0.15,0.288311946046776)
--(axis cs:0.2,0.330083836513053)
--(axis cs:0.25,0.378177996242434)
--(axis cs:0.3,0.422179423955774)
--(axis cs:0.3,0.51702301582923)
--(axis cs:0.3,0.51702301582923)
--(axis cs:0.25,0.4534603578206)
--(axis cs:0.2,0.386886678738107)
--(axis cs:0.15,0.319379389631324)
--(axis cs:0.1,0.266017872605833)
--(axis cs:0.05,0.215390591083565)
--(axis cs:0,0.0542262308625221)
--cycle;

\path [draw=gray, fill=gray, opacity=0.1]
(axis cs:0,0.0463174025084555)
--(axis cs:0,0.0398997036252439)
--(axis cs:0.05,0.747756067736501)
--(axis cs:0.1,0.845643461014655)
--(axis cs:0.15,0.836479789435533)
--(axis cs:0.2,0.812086272987796)
--(axis cs:0.25,0.788082839356035)
--(axis cs:0.3,0.769739151452345)
--(axis cs:0.3,0.824313771273332)
--(axis cs:0.3,0.824313771273332)
--(axis cs:0.25,0.831166874064833)
--(axis cs:0.2,0.850663267818021)
--(axis cs:0.15,0.874291032135817)
--(axis cs:0.1,0.876059603427026)
--(axis cs:0.05,0.772823282258158)
--(axis cs:0,0.0463174025084555)
--cycle;

\path [draw=gray, fill=gray, opacity=0.1]
(axis cs:0,0.0555848281928598)
--(axis cs:0,0.0474501650682391)
--(axis cs:0.05,0.189872265747525)
--(axis cs:0.1,0.241602764856383)
--(axis cs:0.15,0.287800076538005)
--(axis cs:0.2,0.337211950164297)
--(axis cs:0.25,0.379188922015089)
--(axis cs:0.3,0.427739518559058)
--(axis cs:0.3,0.549710983359734)
--(axis cs:0.3,0.549710983359734)
--(axis cs:0.25,0.472489252719027)
--(axis cs:0.2,0.383463554043314)
--(axis cs:0.15,0.313257154888234)
--(axis cs:0.1,0.259221230614618)
--(axis cs:0.05,0.208270897575877)
--(axis cs:0,0.0555848281928598)
--cycle;

\path [draw=color3, fill=color3, opacity=0.1]
(axis cs:0,0.00764820301751617)
--(axis cs:0,0.0034292724371421)
--(axis cs:0.05,0.37942987495362)
--(axis cs:0.1,0.506402387017436)
--(axis cs:0.15,0.535190238040007)
--(axis cs:0.2,0.54504470036556)
--(axis cs:0.25,0.552022083806313)
--(axis cs:0.3,0.55579659979972)
--(axis cs:0.3,0.610008430069359)
--(axis cs:0.3,0.610008430069359)
--(axis cs:0.25,0.592664686632834)
--(axis cs:0.2,0.57014586522053)
--(axis cs:0.15,0.552203212696993)
--(axis cs:0.1,0.520504735117726)
--(axis cs:0.05,0.392029919090875)
--(axis cs:0,0.00764820301751617)
--cycle;

\path [draw=color3, fill=color3, opacity=0.1]
(axis cs:0,0.068702558187816)
--(axis cs:0,0.0657178560811545)
--(axis cs:0.05,0.193723684370083)
--(axis cs:0.1,0.265210516264021)
--(axis cs:0.15,0.305276080620499)
--(axis cs:0.2,0.349465724398086)
--(axis cs:0.25,0.390503197626829)
--(axis cs:0.3,0.419759226361156)
--(axis cs:0.3,0.475631647786259)
--(axis cs:0.3,0.475631647786259)
--(axis cs:0.25,0.437535620017291)
--(axis cs:0.2,0.38366340739589)
--(axis cs:0.15,0.327817669856338)
--(axis cs:0.1,0.275332485626162)
--(axis cs:0.05,0.205384746373134)
--(axis cs:0,0.068702558187816)
--cycle;

\addplot [semithick, color0, dashed]
table {%
0 0.0320296974852681
0.05 0.483360224962235
0.1 0.533207005262375
0.15 0.552308058738708
0.2 0.575874882936478
0.25 0.612612688541412
0.3 0.652075546979904
};
\addplot [semithick, color0]
table {%
0 0.0531957279890776
0.05 0.18527130484581
0.1 0.239064241945744
0.15 0.299044725298882
0.2 0.375401854515076
0.25 0.454289111495018
0.3 0.517307981848717
};
\addplot [semithick, color1, dashed]
table {%
0 0.00726229012943804
0.05 0.434260392189026
0.1 0.528060829639435
0.15 0.533516329526901
0.2 0.521238112449646
0.25 0.522895514965057
0.3 0.541613876819611
};
\addplot [semithick, color1]
table {%
0 0.0563150264322758
0.05 0.277502146363258
0.1 0.323055070638657
0.15 0.337905737757683
0.2 0.352893379330635
0.25 0.375736010074615
0.3 0.406548053026199
};
\addplot [semithick, color2, dashed]
table {%
0 0.0314386619254947
0.05 0.657164293527603
0.1 0.735226958990097
0.15 0.746660673618317
0.2 0.752397137880325
0.25 0.764344251155853
0.3 0.781436717510223
};
\addplot [semithick, color2]
table {%
0 0.0513334099203348
0.05 0.208462762832642
0.1 0.257257622480392
0.15 0.30384566783905
0.2 0.35848525762558
0.25 0.415819177031517
0.3 0.469601219892502
};
\addplot [semithick, gray, dashed]
table {%
0 0.0431085530668497
0.05 0.76028967499733
0.1 0.86085153222084
0.15 0.855385410785675
0.2 0.831374770402908
0.25 0.809624856710434
0.3 0.797026461362839
};
\addplot [semithick, gray]
table {%
0 0.0515174966305494
0.05 0.199071581661701
0.1 0.2504119977355
0.15 0.30052861571312
0.2 0.360337752103806
0.25 0.425839087367058
0.3 0.488725250959396
};
\addplot [semithick, color3, dashed]
table {%
0 0.00553873772732913
0.05 0.385729897022247
0.1 0.513453561067581
0.15 0.5436967253685
0.2 0.557595282793045
0.25 0.572343385219574
0.3 0.58290251493454
};
\addplot [semithick, color3]
table {%
0 0.0672102071344853
0.05 0.199554215371609
0.1 0.270271500945091
0.15 0.316546875238419
0.2 0.366564565896988
0.25 0.41401940882206
0.3 0.447695437073708
};
\end{axis}

\end{tikzpicture} \\
\begin{tikzpicture}

\definecolor{color0}{rgb}{0.9,0.6,0}
\definecolor{color1}{rgb}{0.35,0.7,0.9}
\definecolor{color2}{rgb}{0.95,0.9,0.25}
\definecolor{color3}{rgb}{0,0.6,0.5}

\begin{axis}[
height=4cm,
tick align=outside,
tick pos=left,
width=4cm,
x grid style={white!69.0196078431373!black},
xlabel={\footnotesize{Epsilon}},
xmin=0, xmax=0.3,
xtick style={color=black},
y grid style={white!69.0196078431373!black},
ylabel style={align=center},
ylabel={\footnotesize{CIFAR100}\\\bigskip\\\footnotesize{Accuracy}},
ymin=0, ymax=80,
ytick style={color=black}
]
\path [draw=color0, fill=color0, opacity=0.1]
(axis cs:0,76.1110243831565)
--(axis cs:0,75.2549759830544)
--(axis cs:0.05,10.9961683339608)
--(axis cs:0.1,6.18405840588568)
--(axis cs:0.15,4.48329074633582)
--(axis cs:0.2,3.72202200541471)
--(axis cs:0.25,3.02221934255125)
--(axis cs:0.3,2.60736505458691)
--(axis cs:0.3,3.74663498928211)
--(axis cs:0.3,3.74663498928211)
--(axis cs:0.25,4.21178067270754)
--(axis cs:0.2,4.62997808804537)
--(axis cs:0.15,5.47670929181115)
--(axis cs:0.1,7.22594153689386)
--(axis cs:0.05,12.6898315363395)
--(axis cs:0,76.1110243831565)
--cycle;

\path [draw=color0, fill=color0, opacity=0.1]
(axis cs:0,75.3100707098095)
--(axis cs:0,74.2319291070851)
--(axis cs:0.05,15.1469539258)
--(axis cs:0.1,9.02261461591177)
--(axis cs:0.15,6.54191507285736)
--(axis cs:0.2,5.04593509041014)
--(axis cs:0.25,3.80265652600484)
--(axis cs:0.3,3.02027657268313)
--(axis cs:0.3,3.90572340252134)
--(axis cs:0.3,3.90572340252134)
--(axis cs:0.25,4.80134361323161)
--(axis cs:0.2,5.70806490577516)
--(axis cs:0.15,7.62608486229278)
--(axis cs:0.1,10.5573853077943)
--(axis cs:0.05,16.5090460207943)
--(axis cs:0,75.3100707098095)
--cycle;

\path [draw=color1, fill=color1, opacity=0.1]
(axis cs:0,74.5259853677831)
--(axis cs:0,73.5740146322169)
--(axis cs:0.05,18.938063156909)
--(axis cs:0.1,10.0573972595465)
--(axis cs:0.15,6.88274306596201)
--(axis cs:0.2,5.35352016983878)
--(axis cs:0.25,4.53214898866058)
--(axis cs:0.3,3.57814951403015)
--(axis cs:0.3,4.43385037915833)
--(axis cs:0.3,4.43385037915833)
--(axis cs:0.25,5.28385099608063)
--(axis cs:0.2,6.44647983016122)
--(axis cs:0.15,8.01925688444693)
--(axis cs:0.1,11.0666027175653)
--(axis cs:0.05,20.1559370872317)
--(axis cs:0,74.5259853677831)
--cycle;

\path [draw=color1, fill=color1, opacity=0.1]
(axis cs:0,72.2673819338342)
--(axis cs:0,70.9486169064978)
--(axis cs:0.05,21.1278775503745)
--(axis cs:0.1,12.6765607358374)
--(axis cs:0.15,8.9502292634808)
--(axis cs:0.2,6.40870482212687)
--(axis cs:0.25,4.91042906785072)
--(axis cs:0.3,3.8118961587148)
--(axis cs:0.3,5.16010386798808)
--(axis cs:0.3,5.16010386798808)
--(axis cs:0.25,6.31957095122276)
--(axis cs:0.2,7.98129513972616)
--(axis cs:0.15,10.0817705152668)
--(axis cs:0.1,13.7674393175684)
--(axis cs:0.05,22.0701220223794)
--(axis cs:0,72.2673819338342)
--cycle;

\path [draw=color2, fill=color2, opacity=0.1]
(axis cs:0,77.5301111735643)
--(axis cs:0,76.2858876667678)
--(axis cs:0.05,11.3684827651264)
--(axis cs:0.1,6.59336841522858)
--(axis cs:0.15,4.97601265595039)
--(axis cs:0.2,3.97840371115728)
--(axis cs:0.25,3.281011154682)
--(axis cs:0.3,2.65352196274487)
--(axis cs:0.3,4.12647805632862)
--(axis cs:0.3,4.12647805632862)
--(axis cs:0.25,4.70898875948731)
--(axis cs:0.2,5.53359637276606)
--(axis cs:0.15,6.63398738219658)
--(axis cs:0.1,8.37263156951263)
--(axis cs:0.05,12.7375173722027)
--(axis cs:0,77.5301111735643)
--cycle;

\path [draw=color2, fill=color2, opacity=0.1]
(axis cs:0,75.9198983622807)
--(axis cs:0,74.6700995014888)
--(axis cs:0.05,16.0699612708987)
--(axis cs:0.1,10.390779249624)
--(axis cs:0.15,7.32979311053532)
--(axis cs:0.2,5.41722435437607)
--(axis cs:0.25,4.05349498563996)
--(axis cs:0.3,3.36026808329297)
--(axis cs:0.3,4.62773192815112)
--(axis cs:0.3,4.62773192815112)
--(axis cs:0.25,5.59650501436004)
--(axis cs:0.2,6.70077567614151)
--(axis cs:0.15,8.49020715649349)
--(axis cs:0.1,11.5452208114112)
--(axis cs:0.05,17.6260386375486)
--(axis cs:0,75.9198983622807)
--cycle;

\path [draw=gray, fill=gray, opacity=0.1]
(axis cs:0,74.2169857028171)
--(axis cs:0,73.2090137478665)
--(axis cs:0.05,13.3334329556555)
--(axis cs:0.1,9.19823505565819)
--(axis cs:0.15,7.17081048490634)
--(axis cs:0.2,5.89379646322343)
--(axis cs:0.25,4.65544726415045)
--(axis cs:0.3,3.94966688968784)
--(axis cs:0.3,4.85033315799588)
--(axis cs:0.3,4.85033315799588)
--(axis cs:0.25,5.86255276636713)
--(axis cs:0.2,6.89220359781173)
--(axis cs:0.15,8.45918962953458)
--(axis cs:0.1,10.3757649214536)
--(axis cs:0.05,15.3865669299036)
--(axis cs:0,74.2169857028171)
--cycle;

\path [draw=gray, fill=gray, opacity=0.1]
(axis cs:0,76.1387465672957)
--(axis cs:0,75.0192520899308)
--(axis cs:0.05,17.8484946941568)
--(axis cs:0.1,11.2233879188096)
--(axis cs:0.15,8.02406387193721)
--(axis cs:0.2,6.12343966957255)
--(axis cs:0.25,4.8107745776451)
--(axis cs:0.3,3.90607271976931)
--(axis cs:0.3,4.85592736415403)
--(axis cs:0.3,4.85592736415403)
--(axis cs:0.25,5.76522534987565)
--(axis cs:0.2,7.21456024650411)
--(axis cs:0.15,9.2799358381458)
--(axis cs:0.1,12.1306121727431)
--(axis cs:0.05,18.8295059467123)
--(axis cs:0,76.1387465672957)
--cycle;

\path [draw=color3, fill=color3, opacity=0.1]
(axis cs:0,79.4503645501281)
--(axis cs:0,78.803637830243)
--(axis cs:0.05,25.6985123285993)
--(axis cs:0.1,12.9535573400717)
--(axis cs:0.15,8.16489954811608)
--(axis cs:0.2,5.94422137071468)
--(axis cs:0.25,4.4505577520648)
--(axis cs:0.3,3.50515087331069)
--(axis cs:0.3,4.43284918581711)
--(axis cs:0.3,4.43284918581711)
--(axis cs:0.25,5.39144225556459)
--(axis cs:0.2,6.6897786445441)
--(axis cs:0.15,9.11110047477211)
--(axis cs:0.1,14.0744427285928)
--(axis cs:0.05,26.3854876866595)
--(axis cs:0,79.4503645501281)
--cycle;

\path [draw=color3, fill=color3, opacity=0.1]
(axis cs:0,76.2982777069656)
--(axis cs:0,75.4837199736985)
--(axis cs:0.05,28.9380053167234)
--(axis cs:0.1,15.4789243279928)
--(axis cs:0.15,9.65991199801955)
--(axis cs:0.2,6.43334724552382)
--(axis cs:0.25,4.48188755756145)
--(axis cs:0.3,3.31045471874327)
--(axis cs:0.3,4.01354530605227)
--(axis cs:0.3,4.01354530605227)
--(axis cs:0.25,5.15811240429158)
--(axis cs:0.2,7.11265275829087)
--(axis cs:0.15,10.5140875976225)
--(axis cs:0.1,16.5210756720072)
--(axis cs:0.05,29.7019948358645)
--(axis cs:0,76.2982777069656)
--cycle;

\addplot [semithick, color0, dashed]
table {%
0 75.6830001831055
0.05 11.8429999351501
0.1 6.70499997138977
0.15 4.98000001907349
0.2 4.17600004673004
0.25 3.61700000762939
0.3 3.17700002193451
};
\addplot [semithick, color0]
table {%
0 74.7709999084473
0.05 15.8279999732971
0.1 9.78999996185303
0.15 7.08399996757507
0.2 5.37699999809265
0.25 4.30200006961823
0.3 3.46299998760223
};
\addplot [semithick, color1, dashed]
table {%
0 74.05
0.05 19.5470001220703
0.1 10.5619999885559
0.15 7.45099997520447
0.2 5.9
0.25 4.90799999237061
0.3 4.00599994659424
};
\addplot [semithick, color1]
table {%
0 71.607999420166
0.05 21.598999786377
0.1 13.2220000267029
0.15 9.51599988937378
0.2 7.19499998092651
0.25 5.61500000953674
0.3 4.48600001335144
};
\addplot [semithick, color2, dashed]
table {%
0 76.907999420166
0.05 12.0530000686646
0.1 7.48299999237061
0.15 5.80500001907349
0.2 4.75600004196167
0.25 3.99499995708466
0.3 3.39000000953674
};
\addplot [semithick, color2]
table {%
0 75.2949989318848
0.05 16.8479999542236
0.1 10.9680000305176
0.15 7.9100001335144
0.2 6.05900001525879
0.25 4.825
0.3 3.99400000572205
};
\addplot [semithick, gray, dashed]
table {%
0 73.7129997253418
0.05 14.3599999427795
0.1 9.78699998855591
0.15 7.81500005722046
0.2 6.39300003051758
0.25 5.25900001525879
0.3 4.40000002384186
};
\addplot [semithick, gray]
table {%
0 75.5789993286133
0.05 18.3390003204346
0.1 11.6770000457764
0.15 8.6519998550415
0.2 6.66899995803833
0.25 5.28799996376038
0.3 4.38100004196167
};
\addplot [semithick, color3, dashed]
table {%
0 79.1270011901856
0.05 26.0420000076294
0.1 13.5140000343323
0.15 8.63800001144409
0.2 6.31700000762939
0.25 4.9210000038147
0.3 3.9690000295639
};
\addplot [semithick, color3]
table {%
0 75.890998840332
0.05 29.3200000762939
0.1 16
0.15 10.086999797821
0.2 6.77300000190735
0.25 4.81999998092651
0.3 3.66200001239777
};
\end{axis}

\end{tikzpicture} &
\begin{tikzpicture}

\definecolor{color0}{rgb}{0.9,0.6,0}
\definecolor{color1}{rgb}{0.35,0.7,0.9}
\definecolor{color2}{rgb}{0.95,0.9,0.25}
\definecolor{color3}{rgb}{0,0.6,0.5}

\begin{axis}[
height=4cm,
tick align=outside,
tick pos=left,
width=4cm,
x grid style={white!69.0196078431373!black},
xlabel={\footnotesize{Epsilon}},
xmin=0, xmax=0.3,
xtick style={color=black},
y grid style={white!69.0196078431373!black},
ylabel={\footnotesize{Log-Likelihood}},
ymin=-12, ymax=0,
ytick style={color=black}
]
\addplot [very thick, black]
table {%
0  -4.60517
0.05  -4.60517
0.1  -4.60517
0.15  -4.60517
0.2  -4.60517
0.25  -4.60517
0.3  -4.60517
};
\path [draw=color0, fill=color0, opacity=0.1]
(axis cs:0,-0.978680519498818)
--(axis cs:0,-1.01425473204278)
--(axis cs:0.05,-6.83801206382157)
--(axis cs:0.1,-6.92337401248642)
--(axis cs:0.15,-6.7511134020674)
--(axis cs:0.2,-6.82937012411189)
--(axis cs:0.25,-7.04455932019431)
--(axis cs:0.3,-7.26082656079686)
--(axis cs:0.3,-6.26084249074832)
--(axis cs:0.3,-6.26084249074832)
--(axis cs:0.25,-6.22716535760809)
--(axis cs:0.2,-6.21604930352708)
--(axis cs:0.15,-6.31417303228461)
--(axis cs:0.1,-6.52140707499163)
--(axis cs:0.05,-6.44391928032203)
--(axis cs:0,-0.978680519498818)
--cycle;

\path [draw=color0, fill=color0, opacity=0.1]
(axis cs:0,-1.18122219181438)
--(axis cs:0,-1.21617439267238)
--(axis cs:0.05,-5.43655824940458)
--(axis cs:0.1,-5.72327015488132)
--(axis cs:0.15,-5.63981675733553)
--(axis cs:0.2,-5.55892104539759)
--(axis cs:0.25,-5.51559852712097)
--(axis cs:0.3,-5.49692968870619)
--(axis cs:0.3,-5.21399930703789)
--(axis cs:0.3,-5.21399930703789)
--(axis cs:0.25,-5.27482141469166)
--(axis cs:0.2,-5.34205219867977)
--(axis cs:0.15,-5.42947435891676)
--(axis cs:0.1,-5.5198253391716)
--(axis cs:0.05,-5.27934566928922)
--(axis cs:0,-1.18122219181438)
--cycle;

\path [draw=color1, fill=color1, opacity=0.1]
(axis cs:0,-0.954646138991573)
--(axis cs:0,-0.981094654218129)
--(axis cs:0.05,-5.10200106327222)
--(axis cs:0.1,-6.03349118206191)
--(axis cs:0.15,-6.19797889818731)
--(axis cs:0.2,-6.36987549879769)
--(axis cs:0.25,-6.65725292036197)
--(axis cs:0.3,-7.01290609617803)
--(axis cs:0.3,-6.41502681027274)
--(axis cs:0.3,-6.41502681027274)
--(axis cs:0.25,-6.25339714464788)
--(axis cs:0.2,-6.08944955275371)
--(axis cs:0.15,-5.96780551774608)
--(axis cs:0.1,-5.81810364022996)
--(axis cs:0.05,-4.93774528581741)
--(axis cs:0,-0.954646138991573)
--cycle;

\path [draw=color1, fill=color1, opacity=0.1]
(axis cs:0,-1.14957064043493)
--(axis cs:0,-1.18499318881824)
--(axis cs:0.05,-4.41856020882803)
--(axis cs:0.1,-5.26214330143989)
--(axis cs:0.15,-5.4398420323041)
--(axis cs:0.2,-5.53759144988863)
--(axis cs:0.25,-5.64212590508398)
--(axis cs:0.3,-5.78002947143666)
--(axis cs:0.3,-5.54262871019893)
--(axis cs:0.3,-5.54262871019893)
--(axis cs:0.25,-5.46446842982742)
--(axis cs:0.2,-5.35709576549722)
--(axis cs:0.15,-5.25763670107184)
--(axis cs:0.1,-5.0708180278817)
--(axis cs:0.05,-4.29688975069527)
--(axis cs:0,-1.14957064043493)
--cycle;

\path [draw=color2, fill=color2, opacity=0.1]
(axis cs:0,-0.928846631922824)
--(axis cs:0,-0.968164288868685)
--(axis cs:0.05,-6.84361706184874)
--(axis cs:0.1,-7.01315785769952)
--(axis cs:0.15,-6.96976475636822)
--(axis cs:0.2,-7.16953651812835)
--(axis cs:0.25,-7.5317602090272)
--(axis cs:0.3,-7.90974726220342)
--(axis cs:0.3,-6.56796087568399)
--(axis cs:0.3,-6.56796087568399)
--(axis cs:0.25,-6.51184927104678)
--(axis cs:0.2,-6.41370074219471)
--(axis cs:0.15,-6.39166306233167)
--(axis cs:0.1,-6.50327961507523)
--(axis cs:0.05,-6.46491477367469)
--(axis cs:0,-0.928846631922824)
--cycle;

\path [draw=color2, fill=color2, opacity=0.1]
(axis cs:0,-1.18509348729643)
--(axis cs:0,-1.23098388710506)
--(axis cs:0.05,-5.21545010491674)
--(axis cs:0.1,-5.47525385204102)
--(axis cs:0.15,-5.4040762025079)
--(axis cs:0.2,-5.34226750833671)
--(axis cs:0.25,-5.30421856120328)
--(axis cs:0.3,-5.27654289734117)
--(axis cs:0.3,-5.04453378447947)
--(axis cs:0.3,-5.04453378447947)
--(axis cs:0.25,-5.05710823503491)
--(axis cs:0.2,-5.0943995437558)
--(axis cs:0.15,-5.16752650167643)
--(axis cs:0.1,-5.25066516744223)
--(axis cs:0.05,-5.05499238424765)
--(axis cs:0,-1.18509348729643)
--cycle;

\path [draw=gray, fill=gray, opacity=0.1]
(axis cs:0,-1.44670054232071)
--(axis cs:0,-1.52116839464175)
--(axis cs:0.05,-10.154496742774)
--(axis cs:0.1,-10.8398419011041)
--(axis cs:0.15,-10.864514199141)
--(axis cs:0.2,-10.9187831207274)
--(axis cs:0.25,-11.0640992867567)
--(axis cs:0.3,-11.237996706207)
--(axis cs:0.3,-10.858570809575)
--(axis cs:0.3,-10.858570809575)
--(axis cs:0.25,-10.6752974532464)
--(axis cs:0.2,-10.5602479691206)
--(axis cs:0.15,-10.5368494587923)
--(axis cs:0.1,-10.4475309588059)
--(axis cs:0.05,-9.73083181368618)
--(axis cs:0,-1.44670054232071)
--cycle;

\path [draw=gray, fill=gray, opacity=0.1]
(axis cs:0,-1.15608579791962)
--(axis cs:0,-1.18491410795403)
--(axis cs:0.05,-5.00141343664589)
--(axis cs:0.1,-5.32898880999214)
--(axis cs:0.15,-5.27522447975533)
--(axis cs:0.2,-5.22124707734591)
--(axis cs:0.25,-5.23120569995453)
--(axis cs:0.3,-5.26846341556439)
--(axis cs:0.3,-5.05691226676566)
--(axis cs:0.3,-5.05691226676566)
--(axis cs:0.25,-5.08856621922727)
--(axis cs:0.2,-5.12323602306486)
--(axis cs:0.15,-5.16046568958474)
--(axis cs:0.1,-5.18979353134723)
--(axis cs:0.05,-4.88954763427204)
--(axis cs:0,-1.15608579791962)
--cycle;

\path [draw=color3, fill=color3, opacity=0.1]
(axis cs:0,-0.80516736163722)
--(axis cs:0,-0.821998643698554)
--(axis cs:0.05,-3.70510804953334)
--(axis cs:0.1,-4.43117782944411)
--(axis cs:0.15,-4.69639403906428)
--(axis cs:0.2,-4.88983910539978)
--(axis cs:0.25,-5.04131337167301)
--(axis cs:0.3,-5.14501055456182)
--(axis cs:0.3,-4.8841117183854)
--(axis cs:0.3,-4.8841117183854)
--(axis cs:0.25,-4.82355433086941)
--(axis cs:0.2,-4.72865656083299)
--(axis cs:0.15,-4.59895987053076)
--(axis cs:0.1,-4.38177207906714)
--(axis cs:0.05,-3.6502513134411)
--(axis cs:0,-0.80516736163722)
--cycle;

\path [draw=color3, fill=color3, opacity=0.1]
(axis cs:0,-1.17090777459014)
--(axis cs:0,-1.18902582428883)
--(axis cs:0.05,-3.99462910693233)
--(axis cs:0.1,-4.90546088991591)
--(axis cs:0.15,-5.15754128902123)
--(axis cs:0.2,-5.21724896114018)
--(axis cs:0.25,-5.23850690766054)
--(axis cs:0.3,-5.26794057786399)
--(axis cs:0.3,-5.0930335133618)
--(axis cs:0.3,-5.0930335133618)
--(axis cs:0.25,-5.0972892805111)
--(axis cs:0.2,-5.0914020593644)
--(axis cs:0.15,-5.0358216835941)
--(axis cs:0.1,-4.80662456614608)
--(axis cs:0.05,-3.94567236209626)
--(axis cs:0,-1.17090777459014)
--cycle;

\addplot [semithick, color0, dashed]
table {%
0 -0.996467625770798
0.05 -6.6409656720718
0.1 -6.72239054373903
0.15 -6.53264321717601
0.2 -6.52270971381949
0.25 -6.6358623389012
0.3 -6.76083452577259
};
\addplot [semithick, color0]
table {%
0 -1.19869829224338
0.05 -5.3579519593469
0.1 -5.62154774702646
0.15 -5.53464555812615
0.2 -5.45048662203868
0.25 -5.39520997090631
0.3 -5.35546449787204
};
\addplot [semithick, color1, dashed]
table {%
0 -0.967870396604851
0.05 -5.01987317454481
0.1 -5.92579741114594
0.15 -6.08289220796669
0.2 -6.2296625257757
0.25 -6.45532503250493
0.3 -6.71396645322539
};
\addplot [semithick, color1]
table {%
0 -1.16728191462659
0.05 -4.35772497976165
0.1 -5.1664806646608
0.15 -5.34873936668797
0.2 -5.44734360769292
0.25 -5.5532971674557
0.3 -5.66132909081779
};
\addplot [semithick, color2, dashed]
table {%
0 -0.948505460395755
0.05 -6.65426591776171
0.1 -6.75821873638738
0.15 -6.68071390934994
0.2 -6.79161863016153
0.25 -7.02180474003699
0.3 -7.2388540689437
};
\addplot [semithick, color2]
table {%
0 -1.20803868720075
0.05 -5.1352212445822
0.1 -5.36295950974163
0.15 -5.28580135209217
0.2 -5.21833352604625
0.25 -5.1806633981191
0.3 -5.16053834091032
};
\addplot [semithick, gray, dashed]
table {%
0 -1.48393446848123
0.05 -9.94266427823007
0.1 -10.643686429955
0.15 -10.7006818289667
0.2 -10.739515544924
0.25 -10.8696983700016
0.3 -11.048283757891
};
\addplot [semithick, gray]
table {%
0 -1.17049995293682
0.05 -4.94548053545896
0.1 -5.25939117066969
0.15 -5.21784508467003
0.2 -5.17224155020538
0.25 -5.1598859595909
0.3 -5.16268784116503
};
\addplot [semithick, color3, dashed]
table {%
0 -0.813583002667887
0.05 -3.67767968148722
0.1 -4.40647495425563
0.15 -4.64767695479752
0.2 -4.80924783311639
0.25 -4.93243385127121
0.3 -5.01456113647361
};
\addplot [semithick, color3]
table {%
0 -1.17996679943948
0.05 -3.9701507345143
0.1 -4.856042728031
0.15 -5.09668148630767
0.2 -5.15432551025229
0.25 -5.16789809408582
0.3 -5.1804870456129
};
\end{axis}

\end{tikzpicture} &
\begin{tikzpicture}

\definecolor{color0}{rgb}{0.9,0.6,0}
\definecolor{color1}{rgb}{0.35,0.7,0.9}
\definecolor{color2}{rgb}{0.95,0.9,0.25}
\definecolor{color3}{rgb}{0,0.6,0.5}

\begin{axis}[
height=4cm,
tick align=outside,
tick pos=left,
width=4cm,
x grid style={white!69.0196078431373!black},
xlabel={\footnotesize{Epsilon}},
xmin=0, xmax=0.3,
xtick style={color=black},
y grid style={white!69.0196078431373!black},
ylabel={\footnotesize{ECE}},
ymin=0, ymax=0.8,
ytick style={color=black}
]
\path [draw=color0, fill=color0, opacity=0.1]
(axis cs:0,0.083912112890415)
--(axis cs:0,0.0721508581650926)
--(axis cs:0.05,0.52766281887893)
--(axis cs:0.1,0.505078419860414)
--(axis cs:0.15,0.46725236713975)
--(axis cs:0.2,0.445997044607601)
--(axis cs:0.25,0.439223217917857)
--(axis cs:0.3,0.437450310877524)
--(axis cs:0.3,0.573907819100656)
--(axis cs:0.3,0.573907819100656)
--(axis cs:0.25,0.550144958542409)
--(axis cs:0.2,0.525629398063221)
--(axis cs:0.15,0.520627862810201)
--(axis cs:0.1,0.543763032738157)
--(axis cs:0.05,0.568924433737952)
--(axis cs:0,0.083912112890415)
--cycle;

\path [draw=color0, fill=color0, opacity=0.1]
(axis cs:0,0.130354881313163)
--(axis cs:0,0.113729223582429)
--(axis cs:0.05,0.220337520142397)
--(axis cs:0.1,0.221525765733194)
--(axis cs:0.15,0.207378620096188)
--(axis cs:0.2,0.196846133544489)
--(axis cs:0.25,0.178596760227625)
--(axis cs:0.3,0.162688720638954)
--(axis cs:0.3,0.287428286258972)
--(axis cs:0.3,0.287428286258972)
--(axis cs:0.25,0.27619267563587)
--(axis cs:0.2,0.255669656321959)
--(axis cs:0.15,0.242322885803241)
--(axis cs:0.1,0.243272640033293)
--(axis cs:0.05,0.236094718078772)
--(axis cs:0,0.130354881313163)
--cycle;

\path [draw=color1, fill=color1, opacity=0.1]
(axis cs:0,0.0303815386838012)
--(axis cs:0,0.0210942659878439)
--(axis cs:0.05,0.444641330793319)
--(axis cs:0.1,0.455479621013925)
--(axis cs:0.15,0.429226816462101)
--(axis cs:0.2,0.406287651203184)
--(axis cs:0.25,0.395875664866913)
--(axis cs:0.3,0.391343558648258)
--(axis cs:0.3,0.482169680973858)
--(axis cs:0.3,0.482169680973858)
--(axis cs:0.25,0.45535851486827)
--(axis cs:0.2,0.445934690095873)
--(axis cs:0.15,0.453466319276272)
--(axis cs:0.1,0.478974420705512)
--(axis cs:0.05,0.461884573146882)
--(axis cs:0,0.0303815386838012)
--cycle;

\path [draw=color1, fill=color1, opacity=0.1]
(axis cs:0,0.0908832392290212)
--(axis cs:0,0.0802128174707316)
--(axis cs:0.05,0.25251361639839)
--(axis cs:0.1,0.271014839040616)
--(axis cs:0.15,0.264469625300678)
--(axis cs:0.2,0.265851027074487)
--(axis cs:0.25,0.274572577581962)
--(axis cs:0.3,0.280817108151487)
--(axis cs:0.3,0.326829005482623)
--(axis cs:0.3,0.326829005482623)
--(axis cs:0.25,0.307357815398614)
--(axis cs:0.2,0.297322457727759)
--(axis cs:0.15,0.294505749397961)
--(axis cs:0.1,0.300854683292529)
--(axis cs:0.05,0.270455122872774)
--(axis cs:0,0.0908832392290212)
--cycle;

\path [draw=color2, fill=color2, opacity=0.1]
(axis cs:0,0.0920262612515982)
--(axis cs:0,0.0801573164647524)
--(axis cs:0.05,0.553494336815618)
--(axis cs:0.1,0.51571017419214)
--(axis cs:0.15,0.482242117118335)
--(axis cs:0.2,0.466809325417906)
--(axis cs:0.25,0.469245201171187)
--(axis cs:0.3,0.46953725186449)
--(axis cs:0.3,0.645901507462444)
--(axis cs:0.3,0.645901507462444)
--(axis cs:0.25,0.607973355232927)
--(axis cs:0.2,0.566299874582857)
--(axis cs:0.15,0.543449076223874)
--(axis cs:0.1,0.560895193468383)
--(axis cs:0.05,0.58805910089419)
--(axis cs:0,0.0920262612515982)
--cycle;

\path [draw=color2, fill=color2, opacity=0.1]
(axis cs:0,0.140544596130458)
--(axis cs:0,0.126208329384717)
--(axis cs:0.05,0.197880338053438)
--(axis cs:0.1,0.195570440638368)
--(axis cs:0.15,0.183376967122266)
--(axis cs:0.2,0.170268583622864)
--(axis cs:0.25,0.159533479884115)
--(axis cs:0.3,0.15368123663655)
--(axis cs:0.3,0.232587554919808)
--(axis cs:0.3,0.232587554919808)
--(axis cs:0.25,0.232643440291437)
--(axis cs:0.2,0.223459836515495)
--(axis cs:0.15,0.21540120620518)
--(axis cs:0.1,0.222131000411207)
--(axis cs:0.05,0.22074893273857)
--(axis cs:0,0.140544596130458)
--cycle;

\path [draw=gray, fill=gray, opacity=0.1]
(axis cs:0,0.163241413830855)
--(axis cs:0,0.152019282580278)
--(axis cs:0.05,0.743362356270708)
--(axis cs:0.1,0.74465439666115)
--(axis cs:0.15,0.716882310551301)
--(axis cs:0.2,0.701106934758803)
--(axis cs:0.25,0.693200837492226)
--(axis cs:0.3,0.687217986426514)
--(axis cs:0.3,0.720537376561004)
--(axis cs:0.3,0.720537376561004)
--(axis cs:0.25,0.719718147397758)
--(axis cs:0.2,0.725100714948991)
--(axis cs:0.15,0.744517363864288)
--(axis cs:0.1,0.773848542086256)
--(axis cs:0.05,0.774130009566389)
--(axis cs:0,0.163241413830855)
--cycle;

\path [draw=gray, fill=gray, opacity=0.1]
(axis cs:0,0.140477701530677)
--(axis cs:0,0.128782976164597)
--(axis cs:0.05,0.1925150826351)
--(axis cs:0.1,0.200916121706143)
--(axis cs:0.15,0.187674488935553)
--(axis cs:0.2,0.179375298746428)
--(axis cs:0.25,0.173669783571827)
--(axis cs:0.3,0.167666017987635)
--(axis cs:0.3,0.240957349560354)
--(axis cs:0.3,0.240957349560354)
--(axis cs:0.25,0.231791545649898)
--(axis cs:0.2,0.217193041316667)
--(axis cs:0.15,0.20906915434829)
--(axis cs:0.1,0.213603957191333)
--(axis cs:0.05,0.206655661783998)
--(axis cs:0,0.140477701530677)
--cycle;

\path [draw=color3, fill=color3, opacity=0.1]
(axis cs:0,0.0562233198076524)
--(axis cs:0,0.0447464343517982)
--(axis cs:0.05,0.254157546864244)
--(axis cs:0.1,0.299294135234376)
--(axis cs:0.15,0.291068164574502)
--(axis cs:0.2,0.278057104462662)
--(axis cs:0.25,0.263569529892685)
--(axis cs:0.3,0.248834691422687)
--(axis cs:0.3,0.339975770336881)
--(axis cs:0.3,0.339975770336881)
--(axis cs:0.25,0.329084525464294)
--(axis cs:0.2,0.314778083449325)
--(axis cs:0.15,0.30984323812783)
--(axis cs:0.1,0.313064530231933)
--(axis cs:0.05,0.263611366643218)
--(axis cs:0,0.0562233198076524)
--cycle;

\path [draw=color3, fill=color3, opacity=0.1]
(axis cs:0,0.19022948737172)
--(axis cs:0,0.177602469968521)
--(axis cs:0.05,0.0759885073978117)
--(axis cs:0.1,0.130505764548903)
--(axis cs:0.15,0.145348284892772)
--(axis cs:0.2,0.148515028051375)
--(axis cs:0.25,0.149830758337549)
--(axis cs:0.3,0.150780513629182)
--(axis cs:0.3,0.202647739902274)
--(axis cs:0.3,0.202647739902274)
--(axis cs:0.25,0.190626448626944)
--(axis cs:0.2,0.178960874386789)
--(axis cs:0.15,0.164543594665791)
--(axis cs:0.1,0.145082968408937)
--(axis cs:0.05,0.0874162866395304)
--(axis cs:0,0.19022948737172)
--cycle;

\addplot [semithick, color0, dashed]
table {%
0 0.0780314855277538
0.05 0.548293626308441
0.1 0.524420726299286
0.15 0.493940114974976
0.2 0.485813221335411
0.25 0.494684088230133
0.3 0.50567906498909
};
\addplot [semithick, color0]
table {%
0 0.122042052447796
0.05 0.228216119110584
0.1 0.232399202883244
0.15 0.224850752949715
0.2 0.226257894933224
0.25 0.227394717931747
0.3 0.225058503448963
};
\addplot [semithick, color1, dashed]
table {%
0 0.0257379023358226
0.05 0.4532629519701
0.1 0.467227020859718
0.15 0.441346567869186
0.2 0.426111170649529
0.25 0.425617089867592
0.3 0.436756619811058
};
\addplot [semithick, color1]
table {%
0 0.0855480283498764
0.05 0.261484369635582
0.1 0.285934761166573
0.15 0.279487687349319
0.2 0.281586742401123
0.25 0.290965196490288
0.3 0.303823056817055
};
\addplot [semithick, color2, dashed]
table {%
0 0.0860917888581753
0.05 0.570776718854904
0.1 0.538302683830261
0.15 0.512845596671104
0.2 0.516554600000381
0.25 0.538609278202057
0.3 0.557719379663467
};
\addplot [semithick, color2]
table {%
0 0.133376462757587
0.05 0.209314635396004
0.1 0.208850720524788
0.15 0.199389086663723
0.2 0.19686421006918
0.25 0.196088460087776
0.3 0.193134395778179
};
\addplot [semithick, gray, dashed]
table {%
0 0.157630348205566
0.05 0.758746182918549
0.1 0.759251469373703
0.15 0.730699837207794
0.2 0.713103824853897
0.25 0.706459492444992
0.3 0.703877681493759
};
\addplot [semithick, gray]
table {%
0 0.134630338847637
0.05 0.199585372209549
0.1 0.207260039448738
0.15 0.198371821641922
0.2 0.198284170031548
0.25 0.202730664610863
0.3 0.204311683773994
};
\addplot [semithick, color3, dashed]
table {%
0 0.0504848770797253
0.05 0.258884456753731
0.1 0.306179332733154
0.15 0.300455701351166
0.2 0.296417593955994
0.25 0.29632702767849
0.3 0.294405230879784
};
\addplot [semithick, color3]
table {%
0 0.18391597867012
0.05 0.081702397018671
0.1 0.13779436647892
0.15 0.154945939779282
0.2 0.163737951219082
0.25 0.170228603482246
0.3 0.176714126765728
};
\end{axis}

\end{tikzpicture}
    \end{tabular}
        \raisebox{-2cm}{
        \begin{tikzpicture}

\newenvironment{customlegend}[1][]{%
    \begingroup
    \csname pgfplots@init@cleared@structures\endcsname
    \pgfplotsset{#1}%
}{%
    \csname pgfplots@createlegend\endcsname
    \endgroup
}%

\def\addlegendimage{\csname pgfplots@addlegendimage\endcsname}

\definecolor{color0}{rgb}{0.9,0.6,0}
\definecolor{color1}{rgb}{0.35,0.7,0.9}
\definecolor{color2}{rgb}{0,0.6,0.5}
\definecolor{color3}{rgb}{0.95,0.9,0.25}
\definecolor{color4}{rgb}{0.5,0.5,0.5}

\begin{customlegend}[legend entries={MAP, MAP fVI, MC Dropout, MC Dropout fVI, Ensemble, Ensemble fVI, Radial, Radial fVI, Rank-1, Rank-1 fVI, Uniform prior}, legend columns={1}, legend cell align=left, legend style={draw=none, font=\footnotesize, column sep=.1cm}]
\addlegendimage{color0,semithick,dashed}
\addlegendimage{color0,semithick}
\addlegendimage{color1,semithick,dashed}
\addlegendimage{color1,semithick}
\addlegendimage{color2,semithick,dashed}
\addlegendimage{color2,semithick}
\addlegendimage{color3,semithick,dashed}
\addlegendimage{color3,semithick}
\addlegendimage{color4,semithick,dashed}
\addlegendimage{color4,semithick}
\addlegendimage{black, very thick}
\end{customlegend}
\end{tikzpicture}}}
    \caption{
    Metrics for adversarial examples on CIFAR10 (top) and CIFAR100 (bottom).
    All models use a ResNet-18 architecture. 
    For CIFAR10, there are significant benefits of fVI over weight-space approaches across all metrics. 
    For CIFAR100, the fVI benefits are still evident, but the higher label dimensionality results in stronger regularization from the uniform prior. 
    As a result, the weight-space ensembles achieve slightly better performance over all epsilons.
    }
    \vspace{-.3cm}
    \label{fig:cifar_adversarial}
\end{figure*}

Despite the success of CNNs in computer vision, adversarial attacks are one of the biggest risks when it comes to practical applications \citep{Akhtar2018ThreatOA}.
We evaluate the robustness of fVI compared to standard weight-space prior approaches on the CIFAR10 and CIFAR100 data using the fast gradient sign method (FGSM) \citep{FGSM}.
Figure~\ref{fig:cifar_adversarial} compares the accuracy and the log-likelihood of the test data with increasing amounts of perturbation, ranging from $\epsilon{\,=\,}0$ (no attack) to $\epsilon{\,=\,}0.3$.
Although both weight-space and function-space models lose their classification accuracy when the FGSM attack is introduced, the fVI models only suffer small decreases in log-likelihood, whereas the weight-space LLH performance drops significantly.
We also observe the accuracy vs robustness trade-off in the fVI models.
We attribute this behavior to the quality of the uncertainty quantification at the decision boundary.
While both approaches have brittle boundaries due to the nature of CNNs, the predictive uncertainty at these decision boundaries is richer for fVI.

\section{Implementation Details and Computational Complexity}
\label{app:implementation_details}
We implemented all models using the \texttt{PyTorch} library \citep{NEURIPS2019_9015} and all experiments were conducted using a i5-6600K CPU, a GTX1070 GPU and a GTX2080 GPU with less than 300 hours of total runtime.

The Two Moons data was generated by the \texttt{make\_moons} function from the \texttt{scikit-learn} library using 100 samples, 0.2 noise and random state 456, the \texttt{PyTorch} manual seed was set to 123.
For this toy problem, all models were MLPs with two hidden layers consisting of 25 hidden units each, bias terms enabled and ReLU activation.
For the Dropout models, the dropout rate was set to 0.2 and for the Ensemble models, we used 10 members per ensemble.
All models were trained for 1000 epochs at a learning rate of 0.005 using the Adam optimizer \citep{AdamOpt} with default parameters.
The measurement set for the KL divergence was the visible 2D input plane, discretized at steps of 0.05.
Since there was no mini-batch training, the KL term was not scaled according to Section~\ref{sec:fvi}.
This toy experiment is the only exception in this regard.
For the constant uniform prior, $\bm{\beta}$ was set to $(1, 1)$.
The GP prior and the random forest prior were implemented using their respective \texttt{scikit-learn} implementations by taking their categorical predictions as a Dirichlet mean and using a Dirichlet precision $z = K = 2$ to match the precision of the uniform prior.
The GP used the RBF kernel with optimized hyperparameters and the random forest used 20 trees, the 'entropy' criterion and a maximum depth of 10.
The random seeds were set to 123 for both GP and random forest.

For Rotated MNIST, we used all 60000 images of shape 28x28x1 reshaped to 784 from the train split with pixel values normalized to $[-1, 1]$ and no other pre-processing or data augmentation.
All 10000 images from the test set were used during evaluation, rotated by a fixed degree, ranging from 0\textdegree{} to 180\textdegree{} in 10\textdegree{} steps, resulting in a total of 190000 test images.
The MNIST \citep{lecun2010mnist} data is available under the terms of the Creative Commons Attribution-Share Alike 3.0 license.
All models were MLPs with two hidden layers consisting of 50 units each, bias terms enabled and ReLU activation.
For the Dropout models, the dropout rate was set to 0.2 and for the Ensemble models, we used 10 members per ensemble.
All models were trained for 30 epochs at a learning rate of 0.001 using a mini-batch size of 256.
The measurement set for the KL divergence was the training data itself, except for the measurement set comparison section, where the different measurement sets are stated explicitly.
Results were obtained using 10 random seeds.

For corrupted CIFAR10 and CIFAR100, we used all 50000 images of shape 32x32x3 from the regular train splits.
Following \citep{7780459}, we normalized pixel values using the empirical mean and standard deviation, and employed data augmentation during training by first selecting random crops of size 32x32x3 after adding 4 pixels of zero padding to each side and then randomly flipping 50\% of the images horizontally.
All 10000 images from the regular test set were used during evaluation plus their corrupted versions \citep{hendrycks2018benchmarking} with 19 different corruptions and 5 levels of severity, resulting in a total of 960000 test images.
The CIFAR10 and CIFAR100 \citep{Krizhevsky09learningmultiple} data is available under the terms of the MIT License and the corrupted CIFAR10 and corrupted CIFAR100 \citep{hendrycks2018benchmarking} data is available under the terms of the Apache License 2.0.
All models were CNNs following the ResNet-18 architecture \citep{7780459},
designed for CIFAR images, rather than ImageNet.

Adopting \citep{7780459}, we trained with a batch size of 128 and used the SGD optimizer with momentum (0.9) for 200 epochs and scaled the learning rate by 0.1 at epochs 100 and 150.
For the MAP, MC Dropout and Ensemble models without fVI, we used 0.0005 weight decay.
For MC Dropout models \citep{Gal16}, the dropout rate was set to 0.2.
For Ensemble \citep{NIPS2017_7219} and Ensemble fVI, we used 5 members per ensemble.

For Radial BNNs \citep{farquhar_radial_2020} and Radial fVI, we implemented weight priors for all convolutional weights but not for the final linear layer.
The standard deviation $\sigma$ was parameterized using $\sigma = \log(1 + \exp(\rho))$ and $\rho$ was initialized to -5 while the means were initialized using the \texttt{PyTorch} default initialization scheme for CNNs.
For Radial BNNs without fVI, we used a closed-form Gaussian weight KL divergence with a Gaussian prior with a mean of 0 and a standard deviation of 0.1.
For Radial fVI, we used our fKL instead of the weight-space KL.

For Rank-1 BNNs \citep{dusenberry2020efficient} and Rank-1 fVI, we used 4 ensemble members and 250 training epochs instead of 200 due to slow convergence and scaled the learning rate by 0.1 at epochs 150 and 200.
During training, we used implicit batch ensembling \citep{Wen2020BatchEnsemble}, whereas during prediction, we created explicit ensemble predictions by replicating the input.
We placed Rank-1 Gaussian distributions over all convolutional weights but not over the final linear layer.
The standard deviation $\sigma$ was parameterized using $\sigma = \log(1 + \exp(\rho))$ and $\rho$ was initialized to -3 while the means were initialized to 1.
For Rank-1 BNNs without fVI, following \citep{dusenberry2020efficient}, the Rank-1 priors were Gaussian with a mean of 1 and a standard deviation of 0.1, and weight decay of 0.0001 was used.
We did not use KL annealing epochs.
For Rank-1 fVI, we used our fKL instead of the weight-space KL.
For all fVI models, the measurement set for the KL divergence during fVI training was always the training data itself.
Results were obtained using 10 random seeds.

When scaling to higher dimensional classification tasks, specifically $K \geq 100$, we observed numerical issues with the fELBO objective when using the uniform Dirichlet predictive prior.
In higher dimensions, this prior would provide greater regularization.
This is because the magnitude of the categorical likelihood does not change with dimensionality, as it is the log probability of the label class. 
Conversely, the KL between two Dirichlet densities requires summing over the parameters, so the magnitude naturally increases with $K$.
To alleviate this over-regularization, we adopt the strategy of \citep{joo2020being} and apply additional scaling to the KL term in the fELBO.
This scaling can be shown to be numerically equivalent to a certain prior, i.e. $\bm{\beta}$ (Section 3.4, \citep{joo2020being}).
Therefore, optimizing this scaling is a form of model selection.
For our CIFAR100 experiments, we simply chose a scaling such that that the fKL magnitude was close to the CIFAR10 values. 
We found this to be about $0.1$, which matches a $10$x scaling suggested by the Dirichlet KL due to the summation terms.

To ensure numerical stability, we defined a minimum and maximum precision for the posterior Dirichlet estimation: ${z}_{\mathrm{min}} = K$ and ${z}_{\mathrm{max}} = N$, where $K$ is the number of classes and $N$ is the number of training examples.
For the MAP models, we skipped the Dirichlet MLE and set $z = {z}_{\mathrm{max}}$ because $M = 1$.
Similarly, we used $M = 1$ for the MC Dropout and Radial BNN models during training and also set $z = {z}_{\mathrm{max}}$, although we set $M = 10$ during evaluation.
For the Ensemble models, $M$ was always the number of members in the ensemble and for the Rank-1 models, we replicated the input M times during evaluation, which results in M distinct predictions.
Furthermore, we applied a small amount of label smoothing ${f_{\vx}}_k^{(m)} \approx (1 - \gamma) {f_{\vx}}_k^{(m)} + \gamma \frac{1}{K}$ throughout all steps of the KL divergence estimation, where $\gamma$ was set to $10^{-4}$ for our experiments.

Minka's quasi-Newton maximum likelihood Dirichlet precision estimator \citep{minka2000estimating}, which we used for our implementation, translated to our notation, is given by
\begin{align}
    \frac{1}{z} &= \frac{1}{z} + \frac{1}{{z}^2} \frac{\Delta_1}{\Delta_2},
    \qquad \bar{\alpha}_k = \frac{1}{M}\sum_{m=1}^M {f_{\vx}}_k^{(m)},
    \qquad \breve{\alpha}_k = \frac{1}{M}\sum_{m=1}^M \log {f_{\vx}}_k^{(m)},\\
    \Delta_1 &=
    M \left( \psi_0(z)
    - \sum_{k=1}^K \bar{\alpha}_k \psi_0(z \bar{\alpha}_k)
    + \sum_{k=1}^K \bar{\alpha}_k \breve{\alpha}_k
    \right), \\
    \Delta_2 &=
    M \left( \psi_1(z)
    - \sum_{k=1}^K {\bar{\alpha}_k}^2 \psi_1(z \bar{\alpha}_k)
    \right),
\end{align}
where $\psi_0$ is the digamma function and $\psi_1$ is the trigamma function.

We initialized the algorithm with an approximate maximum likelihood solution using Stirling's approximation to the gamma function $\Gamma$ \citep{minka2000estimating},
\begin{align}
    z^{(0)} = \frac{K - 1}{-2 \sum_{k=1}^K \bar{\alpha}_k (\log \breve{\alpha}_k - \log \bar{\alpha}_k)}.
\end{align}

We stopped the algorithm once the change per step is less than $10^{-5}$. 
Counting the number of iterations until convergence for a trained Ensemble fVI model with $M = 10$ ensemble members and 10000 MNIST test examples, the mean was 3.0796, the $95^{\text{th}}$ quantile was 3, the $99^{\text{th}}$ quantile was 15 and the maximum was 1172.
Note that the number of iterations until convergence in vectorized mini-batch computation is equal to the maximum number of iterations until convergence of the items in the mini-batch.

Although the computational complexity of the underlying deep learning model depends on the model architecture, data input size, number of parameters, etc., for the following comparison, we assume that a single forward pass through the model takes $\mathcal{O}(1)$, i.e. a constant amount of time, because the weight-space and function-space objectives share the same model.
With a mini-batch size of $B$, computation of the standard ML objective takes $\mathcal{O}(BMK)$ time per mini-batch iteration.
Assuming a constant number of quasi-Newton steps, the Dirichlet precision estimation takes $\mathcal{O}(SK + M)$ time for an measurement set of size $S$. Computing the fKL for an measurement set of size $S$ takes $\mathcal{O}(SMK)$ time.
If the training data is used as measurement set the forward pass through the model can be shared between the log-likelihood and fKL calculation, resulting in an overall asymptotic time complexity of $\mathcal{O}(BMK)$ per mini-batch iteration.
In case of a different measurement set, the asymptotic time complexity becomes $\mathcal{O}((B{+}S)MK)$.

\begin{algorithm}[hb!]
\caption{Function-Space Regularization for Deep Bayesian Classification}
\label{alg:VIP}
{\bfseries Require:} training data $\gD = \{(\vx_n, \vy_n)\}_{n=1}^N$,
implicit stochastic process $\vg(\vx, \vw)$,\\
variational posterior $q_{{\bm{\theta}}}(\vw)$,
function-space prior $p(\vf_\vx) = \text{Dir}(\beta_\vx)$
\begin{algorithmic}[1]
 \WHILE{not converged}
 \STATE Sample mini-batch $\{(\vx_l, \vy_l)\}_{l=1}^L \subset
 \gD$
 \STATE Sample $M$ predictions per mini-batch item:
 $\vw^{(m)} \sim q_{\bm{\theta}}(\vw),\quad \vf_l^{(m)}(\vx_l^{(m)}) = g_{{\bm{\theta}}}(\vx_l^{(m)}, \vw^{(m)}) $
 \STATE Compute the expected log-likelihood $\gL_\phi(\gD)\approx\frac{1}{LM}\sum_{l,m=1}^{L,M}\text{Cat}(\vy_l, \vf_l^{(m)})$
 \STATE Sample measurement set $\{\vx_s\}_{s=1}^S \subset
 \gX$
 \STATE Sample $M$ predictions per measurement item:
 $\vw^{(m)} \sim q_{\bm{\theta}}(\vw),\quad f_s^{(m)}(\vx_s^{(m)}) = g_{{\bm{\theta}}}(\vx_s^{(m)}, \vw^{(m)}) $
 \STATE Estimate $\alpha_{\vx_s}$ using Newton method from samples $\{f^{(s)}_{\vx_s}\}_{s=1}^S$ 
 \STATE Estimate factorized fKL: $\KL[q_{\bm{\theta}} \mid\mid p] = \frac{1}{SM}\sum_{s,m=1}^{S,M}(\log q_{\bm{\theta}}(\vf^{(m)}_s) - \log p(\vf^{(m)}_s))$
 \STATE Gradient decent of fELBO approximation
 $\gL({\bm{\theta}}) \approx \gL_{\bm{\theta}}(\gD) + \frac{1}{L} \KL[q_{\bm{\theta}} \mid\mid p]$,\\
 using reparameterization gradients or otherwise, depending on $q_{\bm{\theta}}(\vw)$ (Table \ref{tab:vip_param}).
 \ENDWHILE
\end{algorithmic}
\end{algorithm}

\section{Ablation Studies}
\label{app:ablations}
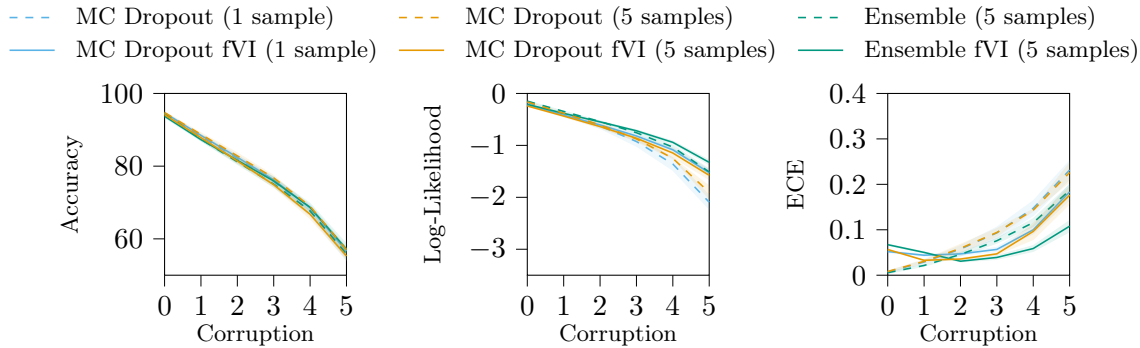
\begin{figure}[t]
    \centering
        \begin{tikzpicture}

\newenvironment{customlegend}[1][]{%
    \begingroup
    \csname pgfplots@init@cleared@structures\endcsname
    \pgfplotsset{#1}%
}{%
    \csname pgfplots@createlegend\endcsname
    \endgroup
}%

\def\addlegendimage{\csname pgfplots@addlegendimage\endcsname}

\definecolor{color0}{rgb}{0.9,0.6,0}
\definecolor{color1}{rgb}{0.35,0.7,0.9}
\definecolor{color2}{rgb}{0,0.6,0.5}

\begin{customlegend}[legend entries={
MC Dropout (1 sample),
MC Dropout (5 samples),
Ensemble (5 samples),
MC Dropout fVI (1 sample),
MC Dropout fVI (5 samples),
Ensemble fVI (5 samples)
},
legend columns={3}, legend cell align=left, legend style={draw=none, font=\footnotesize, column sep=.2cm}]
\addlegendimage{color1,semithick,dashed}
\addlegendimage{color0,semithick,dashed}
\addlegendimage{color2,semithick,dashed}
\addlegendimage{color1,semithick}
\addlegendimage{color0,semithick}
\addlegendimage{color2,semithick}

\end{customlegend}
\end{tikzpicture}
\begin{tikzpicture}

\definecolor{color0}{rgb}{0.9,0.6,0}
\definecolor{color1}{rgb}{0.35,0.7,0.9}
\definecolor{color2}{rgb}{0,0.6,0.5}

\begin{axis}[
height=4cm,
tick align=outside,
tick pos=left,
width=4cm,
x grid style={white!69.0196078431373!black},
xlabel={\footnotesize{Corruption}},
xmin=0, xmax=5,
xtick style={color=black},
xtick={0, 1, 2, 3, 4, 5},
y grid style={white!69.0196078431373!black},
ylabel={\footnotesize{Accuracy}},
ymin=50, ymax=100,
ytick style={color=black}
]

\path [draw=color1, fill=color1, opacity=0.1]
(axis cs:0,94.5882495714158)
--(axis cs:0,94.0617489027053)
--(axis cs:1,87.7747586812721)
--(axis cs:2,81.2880830887978)
--(axis cs:3,74.6034900570646)
--(axis cs:4,66.2739794743415)
--(axis cs:5,54.1050725425429)
--(axis cs:5,57.3355585609728)
--(axis cs:5,57.3355585609728)
--(axis cs:4,68.9145459162834)
--(axis cs:3,77.0137737368807)
--(axis cs:2,82.9738111372764)
--(axis cs:1,88.6416643533959)
--(axis cs:0,94.5882495714158)
--cycle;

\path [draw=color1, fill=color1, opacity=0.1]
(axis cs:0,94.920296942939)
--(axis cs:0,94.4017047050102)
--(axis cs:1,88.1247784934809)
--(axis cs:2,81.8841601434479)
--(axis cs:3,75.6373265427062)
--(axis cs:4,67.7460591451042)
--(axis cs:5,55.5905152848589)
--(axis cs:5,57.6502218672407)
--(axis cs:5,57.6502218672407)
--(axis cs:4,69.3493082865365)
--(axis cs:3,76.8387797195008)
--(axis cs:2,83.0139462408295)
--(axis cs:1,88.7908007301519)
--(axis cs:0,94.920296942939)
--cycle;

\path [draw=color2, fill=color2, opacity=0.1]
(axis cs:0,94.6808288457506)
--(axis cs:0,94.3911682245619)
--(axis cs:1,87.0086823811369)
--(axis cs:2,80.5252307748765)
--(axis cs:3,74.2368478143568)
--(axis cs:4,66.7540322430087)
--(axis cs:5,54.7344830925678)
--(axis cs:5,57.0223589484478)
--(axis cs:5,57.0223589484478)
--(axis cs:4,68.6653372638272)
--(axis cs:3,75.8766226446276)
--(axis cs:2,81.8490856313735)
--(axis cs:1,87.8487913737459)
--(axis cs:0,94.6808288457506)
--cycle;

\path [draw=color2, fill=color2, opacity=0.1]
(axis cs:0,94.0730959550861)
--(axis cs:0,93.3609052045818)
--(axis cs:1,86.9533027517641)
--(axis cs:2,80.9233354015213)
--(axis cs:3,74.93903321621)
--(axis cs:4,67.5911954208134)
--(axis cs:5,55.7635020583534)
--(axis cs:5,58.591235032711)
--(axis cs:5,58.591235032711)
--(axis cs:4,69.7473330215694)
--(axis cs:3,76.9173892081064)
--(axis cs:2,82.3071898059982)
--(axis cs:1,87.6435402047789)
--(axis cs:0,94.0730959550861)
--cycle;

\addplot [semithick, color1, dashed]
table {%
0 94.3249992370606
1 88.208211517334
2 82.1309471130371
3 75.8086318969727
4 67.5942626953125
5 55.7203155517578
};
\addplot [semithick, color1]
table {%
0 94.6610008239746
1 88.4577896118164
2 82.4490531921387
3 76.2380531311035
4 68.5476837158203
5 56.6203685760498
};
\addplot [semithick, color2, dashed]
table {%
0 94.5359985351563
1 87.4287368774414
2 81.187158203125
3 75.0567352294922
4 67.709684753418
5 55.8784210205078
};
\addplot [semithick, color2]
table {%
0 93.717000579834
1 87.2984214782715
2 81.6152626037598
3 75.9282112121582
4 68.6692642211914
5 57.1773685455322
};

\path [draw=color0, fill=color0, opacity=0.1]
(axis cs:0,94.935001373291)
--(axis cs:0,94.5150032043457)
--(axis cs:1,88.3365783691406)
--(axis cs:2,82.0202598571777)
--(axis cs:3,76.1044731140137)
--(axis cs:4,68.879997253418)
--(axis cs:5,56.3589458465576)
--(axis cs:5,57.5484218597412)
--(axis cs:5,57.5484218597412)
--(axis cs:4,69.0315780639648)
--(axis cs:3,77.4107933044434)
--(axis cs:2,83.6297416687012)
--(axis cs:1,89.1481628417969)
--(axis cs:0,94.935001373291)
--cycle;

\path [draw=color0, fill=color0, opacity=0.1]
(axis cs:0,94.2350044250488)
--(axis cs:0,94.0549964904785)
--(axis cs:1,87.4931526184082)
--(axis cs:2,80.8147392272949)
--(axis cs:3,74.2631607055664)
--(axis cs:4,65.5734176635742)
--(axis cs:5,54.2913131713867)
--(axis cs:5,56.2376327514648)
--(axis cs:5,56.2376327514648)
--(axis cs:4,67.9449996948242)
--(axis cs:3,75.5199966430664)
--(axis cs:2,82.0336875915527)
--(axis cs:1,88.1773719787598)
--(axis cs:0,94.2350044250488)
--cycle;

\addplot [semithick, color0, dashed]
table {%
0 94.7250022888184
1 88.7423706054688
2 82.8250007629395
3 76.7576332092285
4 68.9557876586914
5 56.9536838531494
};
\addplot [semithick, color0]
table {%
0 94.1450004577637
1 87.835262298584
2 81.4242134094238
3 74.8915786743164
4 66.7592086791992
5 55.2644729614258
};

\end{axis}

\end{tikzpicture} \hspace{.3cm}
\begin{tikzpicture}

\definecolor{color0}{rgb}{0.9,0.6,0}
\definecolor{color1}{rgb}{0.35,0.7,0.9}
\definecolor{color2}{rgb}{0,0.6,0.5}

\begin{axis}[
height=4cm,
tick align=outside,
tick pos=left,
width=4cm,
x grid style={white!69.0196078431373!black},
xlabel={\footnotesize{Corruption}},
xmin=0, xmax=5,
xtick style={color=black},
xtick={0, 1, 2, 3, 4, 5},
y grid style={white!69.0196078431373!black},
ylabel={\footnotesize{Log-Likelihood}},
ymin=-3.5, ymax=0,
ytick style={color=black}
]

\path [draw=color1, fill=color1, opacity=0.1]
(axis cs:0,-0.167124000800952)
--(axis cs:0,-0.18026119838821)
--(axis cs:1,-0.415498332687885)
--(axis cs:2,-0.673819808896183)
--(axis cs:3,-1.01470769396123)
--(axis cs:4,-1.4780014925916)
--(axis cs:5,-2.22769618612646)
--(axis cs:5,-1.96169155784273)
--(axis cs:5,-1.96169155784273)
--(axis cs:4,-1.24000415668782)
--(axis cs:3,-0.836601749145725)
--(axis cs:2,-0.578617247552648)
--(axis cs:1,-0.374434867152542)
--(axis cs:0,-0.167124000800952)
--cycle;

\path [draw=color1, fill=color1, opacity=0.1]
(axis cs:0,-0.215957181508538)
--(axis cs:0,-0.228001987376165)
--(axis cs:1,-0.427080451405458)
--(axis cs:2,-0.62805855573085)
--(axis cs:3,-0.843857512895297)
--(axis cs:4,-1.11858484880371)
--(axis cs:5,-1.58242242050449)
--(axis cs:5,-1.46868787586959)
--(axis cs:5,-1.46868787586959)
--(axis cs:4,-1.05241429896049)
--(axis cs:3,-0.792858327910666)
--(axis cs:2,-0.585110133677346)
--(axis cs:1,-0.404106760171514)
--(axis cs:0,-0.215957181508538)
--cycle;

\path [draw=color2, fill=color2, opacity=0.1]
(axis cs:0,-0.143297958571428)
--(axis cs:0,-0.152798936254461)
--(axis cs:1,-0.353251536140419)
--(axis cs:2,-0.555707132442149)
--(axis cs:3,-0.780645275580325)
--(axis cs:4,-1.06655176863937)
--(axis cs:5,-1.56623520588887)
--(axis cs:5,-1.45777418821871)
--(axis cs:5,-1.45777418821871)
--(axis cs:4,-0.998101796104743)
--(axis cs:3,-0.732580098071682)
--(axis cs:2,-0.527781078787823)
--(axis cs:1,-0.338864758807262)
--(axis cs:0,-0.143297958571428)
--cycle;

\path [draw=color2, fill=color2, opacity=0.1]
(axis cs:0,-0.206886588331722)
--(axis cs:0,-0.212307902352788)
--(axis cs:1,-0.38964636386453)
--(axis cs:2,-0.55976897940216)
--(axis cs:3,-0.736341258587666)
--(axis cs:4,-0.966342597547088)
--(axis cs:5,-1.36960624558302)
--(axis cs:5,-1.28432668813712)
--(axis cs:5,-1.28432668813712)
--(axis cs:4,-0.916578072388488)
--(axis cs:3,-0.70123949652235)
--(axis cs:2,-0.534803193468391)
--(axis cs:1,-0.377061271002886)
--(axis cs:0,-0.206886588331722)
--cycle;

\path [draw=color0, fill=color0, opacity=0.1]
(axis cs:0,-0.163426374181273)
--(axis cs:0,-0.164908764373085)
--(axis cs:1,-0.393769281611239)
--(axis cs:2,-0.643928764844473)
--(axis cs:3,-0.923434069305706)
--(axis cs:4,-1.2523759856622)
--(axis cs:5,-2.02657207452419)
--(axis cs:5,-1.78418515405313)
--(axis cs:5,-1.78418515405313)
--(axis cs:4,-1.24699031714555)
--(axis cs:3,-0.82785538476422)
--(axis cs:2,-0.569878901120287)
--(axis cs:1,-0.369649756765966)
--(axis cs:0,-0.163426374181273)
--cycle;

\path [draw=color0, fill=color0, opacity=0.1]
(axis cs:0,-0.236143968768138)
--(axis cs:0,-0.245774890701152)
--(axis cs:1,-0.450322997410862)
--(axis cs:2,-0.667149575907316)
--(axis cs:3,-0.904355041967489)
--(axis cs:4,-1.22575414261057)
--(axis cs:5,-1.65710382067268)
--(axis cs:5,-1.49678991530157)
--(axis cs:5,-1.49678991530157)
--(axis cs:4,-1.06883653725098)
--(axis cs:3,-0.820638257696397)
--(axis cs:2,-0.608326474999698)
--(axis cs:1,-0.420490969734458)
--(axis cs:0,-0.236143968768138)
--cycle;

\addplot [semithick, color1, dashed]
table {%
0 -0.173692599594581
1 -0.394966599920213
2 -0.626218528224416
3 -0.925654721553479
4 -1.35900282463971
5 -2.0946938719846
};
\addplot [semithick, color1]
table {%
0 -0.221979584442351
1 -0.415593605788486
2 -0.606584344704098
3 -0.818357920402981
4 -1.0854995738821
5 -1.52555514818704
};
\addplot [semithick, color2, dashed]
table {%
0 -0.148048447412945
1 -0.346058147473841
2 -0.541744105614986
3 -0.756612686826004
4 -1.03232678237206
5 -1.51200469705379
};
\addplot [semithick, color2]
table {%
0 -0.209597245342255
1 -0.383353817433708
2 -0.547286086435276
3 -0.718790377555008
4 -0.941460334967788
5 -1.32696646686007
};


\addplot [semithick, color0, dashed]
table {%
0 -0.164167569277179
1 -0.381709519188603
2 -0.60690383298238
3 -0.875644727034963
4 -1.24968315140387
5 -1.90537861428866
};
\addplot [semithick, color0]
table {%
0 -0.240959429734645
1 -0.43540698357266
2 -0.637738025453507
3 -0.862496649831943
4 -1.14729533993078
5 -1.57694686798712
};

\end{axis}

\end{tikzpicture} \hspace{.3cm}
\begin{tikzpicture}

\definecolor{color0}{rgb}{0.9,0.6,0}
\definecolor{color1}{rgb}{0.35,0.7,0.9}
\definecolor{color2}{rgb}{0,0.6,0.5}

\begin{axis}[
height=4cm,
tick align=outside,
tick pos=left,
width=4cm,
x grid style={white!69.0196078431373!black},
xlabel={\footnotesize{Corruption}},
xmin=0, xmax=5,
xtick style={color=black},
xtick={0, 1, 2, 3, 4, 5},
y grid style={white!69.0196078431373!black},
ylabel={\footnotesize{ECE}},
ymin=0, ymax=0.4,
ytick style={color=black}
]

\path [draw=color1, fill=color1, opacity=0.1]
(axis cs:0,0.00930358599242805)
--(axis cs:0,0.00534132077055932)
--(axis cs:1,0.0255022462207664)
--(axis cs:2,0.0501661440700979)
--(axis cs:3,0.0806233086476648)
--(axis cs:4,0.131326702458758)
--(axis cs:5,0.211461839221543)
--(axis cs:5,0.251600307084495)
--(axis cs:5,0.251600307084495)
--(axis cs:4,0.163627689855199)
--(axis cs:3,0.105817881893126)
--(axis cs:2,0.0649921582668333)
--(axis cs:1,0.0326166920613657)
--(axis cs:0,0.00930358599242805)
--cycle;

\path [draw=color1, fill=color1, opacity=0.1]
(axis cs:0,0.0541858990513501)
--(axis cs:0,0.0497939515090767)
--(axis cs:1,0.0422048705050397)
--(axis cs:2,0.0438093988072562)
--(axis cs:3,0.0522413664759478)
--(axis cs:4,0.0910869449408548)
--(axis cs:5,0.165983835567195)
--(axis cs:5,0.200830685387891)
--(axis cs:5,0.200830685387891)
--(axis cs:4,0.108947189149092)
--(axis cs:3,0.0613205014465967)
--(axis cs:2,0.0500209020603013)
--(axis cs:1,0.0457895075550627)
--(axis cs:0,0.0541858990513501)
--cycle;

\path [draw=color2, fill=color2, opacity=0.1]
(axis cs:0,0.00764820301751617)
--(axis cs:0,0.0034292724371421)
--(axis cs:1,0.0198617261353603)
--(axis cs:2,0.0422349446268511)
--(axis cs:3,0.0706035810898349)
--(axis cs:4,0.107322952553259)
--(axis cs:5,0.174455814079189)
--(axis cs:5,0.198795817776776)
--(axis cs:5,0.198795817776776)
--(axis cs:4,0.124244406417383)
--(axis cs:3,0.0811278042007878)
--(axis cs:2,0.0492155119030046)
--(axis cs:1,0.0231239463683495)
--(axis cs:0,0.00764820301751617)
--cycle;

\path [draw=color2, fill=color2, opacity=0.1]
(axis cs:0,0.068702558187816)
--(axis cs:0,0.0657178560811545)
--(axis cs:1,0.0487115128635035)
--(axis cs:2,0.0291465684364354)
--(axis cs:3,0.0350029452435855)
--(axis cs:4,0.0524082939152871)
--(axis cs:5,0.0957724233183993)
--(axis cs:5,0.120057631560789)
--(axis cs:5,0.120057631560789)
--(axis cs:4,0.0650432211334552)
--(axis cs:3,0.0432162651128884)
--(axis cs:2,0.0324699443240846)
--(axis cs:1,0.0516200908721819)
--(axis cs:0,0.068702558187816)
--cycle;

\addplot [semithick, color1, dashed]
table {%
0 0.00732245338149369
1 0.0290594691410661
2 0.0575791511684656
3 0.0932205952703953
4 0.147477196156979
5 0.231531073153019
};
\addplot [semithick, color1]
table {%
0 0.0519899252802134
1 0.0439971890300512
2 0.0469151504337788
3 0.0567809339612722
4 0.100017067044973
5 0.183407260477543
};
\addplot [semithick, color2, dashed]
table {%
0 0.00553873772732913
1 0.0214928362518549
2 0.0457252282649279
3 0.0758656926453114
4 0.115783679485321
5 0.186625815927982
};
\addplot [semithick, color2]
table {%
0 0.0672102071344853
1 0.0501658018678427
2 0.03080825638026
3 0.039109605178237
4 0.0587257575243711
5 0.107915027439594
};

\path [draw=color0, fill=color0, opacity=0.1]
(axis cs:0,0.00883672619238496)
--(axis cs:0,0.00750950397923589)
--(axis cs:1,0.0270273722708225)
--(axis cs:2,0.0504160840064287)
--(axis cs:3,0.0854721218347549)
--(axis cs:4,0.141239799559116)
--(axis cs:5,0.205038197338581)
--(axis cs:5,0.247041262686253)
--(axis cs:5,0.247041262686253)
--(axis cs:4,0.145154722034931)
--(axis cs:3,0.102654859423637)
--(axis cs:2,0.0678575728088617)
--(axis cs:1,0.0340386815369129)
--(axis cs:0,0.00883672619238496)
--cycle;

\path [draw=color0, fill=color0, opacity=0.1]
(axis cs:0,0.0574053470045328)
--(axis cs:0,0.0557487327605486)
--(axis cs:1,0.0321354269981384)
--(axis cs:2,0.0337143838405609)
--(axis cs:3,0.0452508293092251)
--(axis cs:4,0.0785649530589581)
--(axis cs:5,0.157082334160805)
--(axis cs:5,0.195527508854866)
--(axis cs:5,0.195527508854866)
--(axis cs:4,0.114244934171438)
--(axis cs:3,0.0487187914550304)
--(axis cs:2,0.0373888462781906)
--(axis cs:1,0.0335085690021515)
--(axis cs:0,0.0574053470045328)
--cycle;

\addplot [semithick, color0, dashed]
table {%
0 0.00817311508581042
1 0.0305330269038677
2 0.0591368284076452
3 0.0940634906291962
4 0.143197260797024
5 0.226039730012417
};
\addplot [semithick, color0]
table {%
0 0.0565770398825407
1 0.032821998000145
2 0.0355516150593758
3 0.0469848103821278
4 0.0964049436151981
5 0.176304921507835
};

\end{axis}

\end{tikzpicture}
    \caption{
    Reproduction of CIFAR10 corruption results in Figure \ref{fig:cifar_corrupted},
    including MC Dropout results with 5 predictive samples during training.
    For 5 sample MC Dropout, 3 random seeds were used rather than 10 due to the additional training time.
    }
    \label{fig:cifar_sample_ablation}
\end{figure}

\paragraph{Samples During Training}
In Section \ref{sec:estimation}, a Dirichlet estimation procedure was proposed using $M$ samples.
In the single sample case $M = 1$, motivated by MAP models, a crude approximation was proposed to approximate the precision with the number of training data samples. During training, the $M = 1$ approximation was also used for MC Dropout, Radial and Rank-1 BNNs, akin to their respective weight-space variational inference procedures.
To assess the consequence of this approximation, we repeated the CIFAR10 corruption experiment for MC dropout with $M = 5$, matching the Ensemble models. 
Figure \ref{fig:cifar_sample_ablation} shows that the 5 sample MC Dropout performance is closer to the 1 sample MC Dropout performance than the Ensemble.
This result indicates that the model, rather than $M$ during training, has greater impact.
The similarity in performance between 1 and 5 sample MC Dropout suggests that the 1 sample approximation is reasonable. 
\paragraph{Scaling Issues with High Label Dimensionality}
The CIFAR100 experiments revealed an issue with the fELBO objective that caused underfitting for the larger label dimension.
To examine why, recall that the categorical likelihood is $\log {f_{\vx}}_k$ when $y_k{\,=\,}1$. 
Therefore, the dimensionality of $\vy$ does not directly influence the value.
Conversely, the KL divergence between two Dirichlet densities $\KL(p_1||p_2)$ does incorporate the label dimensionality $K$ \citep{rauber2008probabilistic},
\begin{align*}
    \KL(p_1||p_2) &= 
    \log\Gamma(z^{(1)})
    - \sum_{k=1}^K
    \log \Gamma(\alpha^{(1)}_k)
    - \log\Gamma(z^{(2)})
    + \sum_{k=1}^K
    \log\Gamma(\alpha_k^{(2)}) \\
    &\quad + \sum_{k=1}^K
    (\alpha_k^{(1)}{-}\alpha_k^{(2)})
    (\Psi(\alpha_k^{(1)}){-}\Psi(\alpha_k^{(2)})).
\end{align*}

To counteract this linear increase due to the summation terms, we can assess a heuristic annealing scale factor on the fKL during training of $1/K = \lambda$,
\begin{align*}
\mathcal{L}(\bm{\theta}) = \E_{\,\vf \sim q(\cdot |\bm{\theta})}
\left[ \log p(\mathcal{D} | \vf) \right]
- \lambda\KL[{q(\vf |\bm{\theta}) \mid\mid p(\vf)}].
\end{align*}

To investigate this relationship between the Dirichlet KL divergence and the number of classes of the classification, we conducted a toy experiment in a hypercube $[-1, 1]^D$ with fixed number of input dimensions $D$ and increasing number of classes $K$.
The classes were created by using each dimension as the decision boundary, i.e. $\vx_d{\,=\,}0$, and leveraging all permutations to create up to $K{\,=\,}2^D$ classes, where $D$ was set to 8.
The training data, measurement set for fKL, and test data, consisting of 1000 data points each, were all sampled uniformly at random from the hypercube.
We used a MAP model with 2-layer MLP architecture with 25 hidden units each, bias terms enabled, and ReLU activation functions.
We trained for 3000 epochs using Adam optimizer with a learning rate of 0.005 and default parameters otherwise.

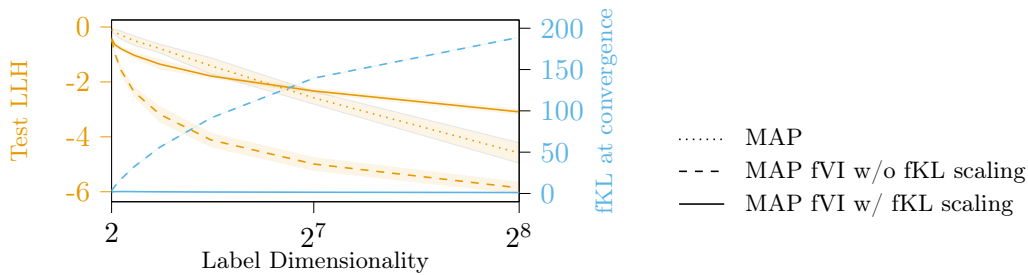
\begin{figure}[tb]
    \centering
        \raisebox{-.75cm}{
\begin{tikzpicture}

\definecolor{color0}{rgb}{0.9,0.6,0}
\definecolor{color1}{rgb}{0.35,0.7,0.9}

\begin{axis}[
height=4cm,
width=7cm,
tick align=outside,
tick pos=left,
x grid style={white!69.0196078431373!black},
xlabel={\footnotesize{Label Dimensionality}},
xtick={2, 128, 256},
xticklabels={
$2$,
$2^7$,
$2^8$},
xmin=2, xmax=256,
xtick style={color=black},
y grid style={white!69.0196078431373!black},
ylabel={\color{color0}\footnotesize{Test LLH}},
ymin=-6.3638164953769, ymax=0.256788927969079,
ytick style={color=color0},
ytick={0, -2, -4, -6},
yticklabels={
\color{color0}0,
\color{color0}-2,
\color{color0}-4,
\color{color0}-6}
]
\path [draw=black, fill=color0, opacity=0.1]
(axis cs:2,-0.0441476821830113)
--(axis cs:2,-0.14067357481219)
--(axis cs:4,-0.290263897496851)
--(axis cs:8,-0.443911132471814)
--(axis cs:16,-0.661424272896653)
--(axis cs:32,-0.936316823370671)
--(axis cs:64,-1.70869575281564)
--(axis cs:128,-2.80818961000075)
--(axis cs:256,-4.95786873177793)
--(axis cs:256,-4.21566461249087)
--(axis cs:256,-4.21566461249087)
--(axis cs:128,-2.34986502790819)
--(axis cs:64,-1.12087532739218)
--(axis cs:32,-0.62806738673904)
--(axis cs:16,-0.327148402093047)
--(axis cs:8,-0.148934590561137)
--(axis cs:4,-0.0934629288133064)
--(axis cs:2,-0.0441476821830113)
--cycle;

\path [draw=color0, fill=color0, opacity=0.1]
(axis cs:2,-0.397056691878598)
--(axis cs:2,-0.525914127594669)
--(axis cs:4,-1.02229719299495)
--(axis cs:8,-1.71561165373951)
--(axis cs:16,-2.52455142836458)
--(axis cs:32,-3.45040411243395)
--(axis cs:64,-4.36266946678249)
--(axis cs:128,-5.22098761740905)
--(axis cs:256,-6.06287988522481)
--(axis cs:256,-5.65810571810771)
--(axis cs:256,-5.65810571810771)
--(axis cs:128,-4.75885943230408)
--(axis cs:64,-3.86605906600865)
--(axis cs:32,-2.90104566326185)
--(axis cs:16,-2.19511309762113)
--(axis cs:8,-1.45515968281597)
--(axis cs:4,-0.895773004110458)
--(axis cs:2,-0.397056691878598)
--cycle;

\path [draw=color0, fill=color0, opacity=0.1]
(axis cs:2,-0.340885840708734)
--(axis cs:2,-0.485628385489463)
--(axis cs:4,-0.693599198542022)
--(axis cs:8,-0.861899040415307)
--(axis cs:16,-1.13296070427057)
--(axis cs:32,-1.52595924438632)
--(axis cs:64,-1.85864285725444)
--(axis cs:128,-2.41667455528998)
--(axis cs:256,-3.17375063558933)
--(axis cs:256,-3.0029301200259)
--(axis cs:256,-3.0029301200259)
--(axis cs:128,-2.24192186499811)
--(axis cs:64,-1.71524968844564)
--(axis cs:32,-1.17725842891538)
--(axis cs:16,-0.92068999677542)
--(axis cs:8,-0.731323613450507)
--(axis cs:4,-0.592904247560121)
--(axis cs:2,-0.340885840708734)
--cycle;

\addplot [dotted, semithick, color0]
table {%
2 -0.0924106284976006
4 -0.191863413155079
8 -0.296422861516476
16 -0.49428633749485
32 -0.782192105054855
64 -1.41478554010391
128 -2.57902731895447
256 -4.5867666721344
};
\addplot [semithick, color0, dashed]
table {%
2 -0.461485409736633
4 -0.959035098552704
8 -1.58538566827774
16 -2.35983226299286
32 -3.1757248878479
64 -4.11436426639557
128 -4.98992352485657
256 -5.86049280166626
};
\addplot [semithick, color0]
table {%
2 -0.413257113099098
4 -0.643251723051071
8 -0.796611326932907
16 -1.02682535052299
32 -1.35160883665085
64 -1.78694627285004
128 -2.32929821014404
256 -3.08834037780762
};
\end{axis}

\begin{axis}[
height=4cm,
width=7cm,
axis y line=right,
y axis line style=-,
tick align=outside,
x grid style={white!69.0196078431373!black},
xmin=2, xmax=256,
xticklabels={,,},
xtick=\empty,
y grid style={white!69.0196078431373!black},
ylabel={\color{color1}\footnotesize{fKL at convergence}},
ymin=-10, ymax=210,
ytick pos=right,
ytick style={color=black},
ytick={0, 50, 100, 150, 200},
yticklabels={
\color{color1}0,
\color{color1}50,
\color{color1}100,
\color{color1}150,
\color{color1}200}
]
\path [draw=color1, fill=color1, opacity=0.1]
(axis cs:2,3.27958701356345)
--(axis cs:2,3.27197639242715)
--(axis cs:4,8.64931854056797)
--(axis cs:8,17.5677889875069)
--(axis cs:16,32.2917922454107)
--(axis cs:32,55.8547736718749)
--(axis cs:64,91.4275335685086)
--(axis cs:128,139.549354235174)
--(axis cs:256,189.0325209648)
--(axis cs:256,189.038270661176)
--(axis cs:256,189.038270661176)
--(axis cs:128,139.561067517756)
--(axis cs:64,91.4595864876438)
--(axis cs:32,55.9036529943849)
--(axis cs:16,32.3543861908686)
--(axis cs:8,17.6484074541435)
--(axis cs:4,8.6952933902792)
--(axis cs:2,3.27958701356345)
--cycle;

\path [draw=color1, fill=color1, opacity=0.1]
(axis cs:2,1.66789877504964)
--(axis cs:2,1.65222458749156)
--(axis cs:4,2.25187220381691)
--(axis cs:8,2.35500407113775)
--(axis cs:16,2.23591469520838)
--(axis cs:32,2.02395913410822)
--(axis cs:64,1.76451056485765)
--(axis cs:128,1.47431526810231)
--(axis cs:256,1.13788977497729)
--(axis cs:256,1.15430931216566)
--(axis cs:256,1.15430931216566)
--(axis cs:128,1.48881296485362)
--(axis cs:64,1.77984446997054)
--(axis cs:32,2.04203398417791)
--(axis cs:16,2.25371824508397)
--(axis cs:8,2.37501530752436)
--(axis cs:4,2.26057272149131)
--(axis cs:2,1.66789877504964)
--cycle;

\addplot [semithick, color1, dashed]
table {%
2 3.2757817029953
4 8.67230596542358
8 17.6080982208252
16 32.3230892181396
32 55.8792133331299
64 91.4435600280762
128 139.555210876465
256 189.035395812988
};
\addplot [semithick, color1]
table {%
2 1.6600616812706
4 2.25622246265411
8 2.36500968933105
16 2.24481647014618
32 2.03299655914307
64 1.77217751741409
128 1.48156411647797
256 1.14609954357147
};
\end{axis}

\end{tikzpicture}}
        \hspace{.2cm}
        \begin{tikzpicture}

\newenvironment{customlegend}[1][]{%
    \begingroup
    \csname pgfplots@init@cleared@structures\endcsname
    \pgfplotsset{#1}%
}{%
    \csname pgfplots@createlegend\endcsname
    \endgroup
}%

\def\addlegendimage{\csname pgfplots@addlegendimage\endcsname}

\definecolor{color0}{rgb}{0.9,0.6,0}
\definecolor{color1}{rgb}{0.35,0.7,0.9}

\begin{customlegend}[legend entries={
MAP,
MAP fVI w/o fKL scaling,
MAP fVI w/ fKL scaling,
},
legend columns={1},
legend cell align=left,
legend style={draw=none, font=\footnotesize, column sep=.2cm}]
\addlegendimage{dotted, semithick}
\addlegendimage{dashed, semithick}
\addlegendimage{semithick}
\end{customlegend}
\end{tikzpicture}
    \caption{
    A toy classification example on a hypercube to illustrate the scaling issues associated with the Dirichlet density.
    The vanilla MAP performance acts as a baseline, and demonstrates the typical range of log-likelihood values for this task across increasing label dimensionality.
    For fVI, the Dirichlet fKL reports a significantly larger range that is x100 the log-likelihood range. 
    This value imbalance affects the fELBO objective, resulting in significant underfitting.
    Applying a heuristic to scale the fKL term, keeping the fKL invariant across label dimensionality avoids the underfitting phemonema. 
    Plot reports mean and 2 standard deviations over 10 seeds.}
    \label{fig:hypercube}
\end{figure}

Figure~\ref{fig:hypercube} illustrates the model's test log-likelihood and fKL after training.
The regular MAP model represents a decent baseline with linear decrease in performance as the label dimensionality increases.
In contrast, the test log-likelihood of the MAP fVI model without KL scaling decreases exponentially while the fKL increases approximately linearly.
Applying the above proposed scaling to the fKL during training aids the optimization, keeps the final fKL at convergence consistently low and significantly improves the model's test log-likelihood.

\paragraph{Changing Prior Dirichlet Parameters}
The uniform Dirichlet distribution with concentration parameters $\beta_k = 1$ is a natural choice for an uninformed prior over the simplex.
However, potentially interesting cases to consider are priors where all $\beta_k$ are set to another value which is greater or smaller than 1.
While the Dirichlet mean remains the same, $\beta_k > 1$ corresponds to greater confidence that the class probabilities are uniformly distributed and $\beta_k < 1$ prefers dominance of any particular class.
It was also hypothesized that scaling $\beta_k$ could yield results comparable to scaling the fKL as discussed in the previous subsection.
To test this hypothesis, we repeated the hypercube experiment from the previous subsection with the MAP fVI model without fKL scaling while using different $\beta_k$ as prior parameters.

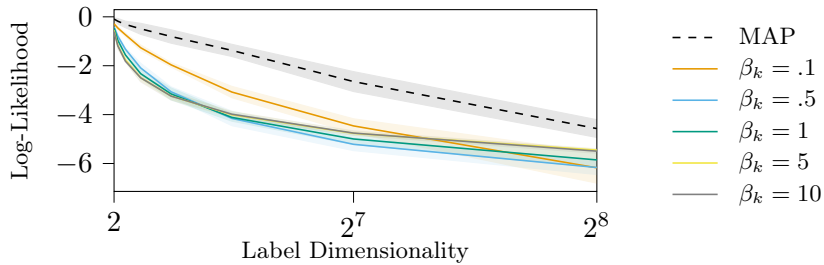
\begin{figure}[tb]
    \centering
        \raisebox{-.75cm}{
\begin{tikzpicture}

\definecolor{color0}{rgb}{0.9,0.6,0}
\definecolor{color1}{rgb}{0.35,0.7,0.9}
\definecolor{color2}{rgb}{0,0.6,0.5}
\definecolor{color3}{rgb}{0.95,0.9,0.25}

\begin{axis}[
height=4cm,
width=8cm,
tick align=outside,
tick pos=left,
x grid style={white!69.0196078431373!black},
xlabel={\footnotesize{Label Dimensionality}},
xtick={2, 128, 256},
xticklabels={
$2$,
$2^7$,
$2^8$},
xmin=2, xmax=256,
xtick style={color=black},
y grid style={white!69.0196078431373!black},
ylabel={\footnotesize{Log-Likelihood}},
ymin=-7.14216475812512, ymax=0.293853130957089,
ytick style={color=black}
]
\path [draw=black, fill=black, opacity=0.1]
(axis cs:2,-0.0441476821830113)
--(axis cs:2,-0.14067357481219)
--(axis cs:4,-0.292841432403753)
--(axis cs:8,-0.446463312902103)
--(axis cs:16,-0.717020344370989)
--(axis cs:32,-1.07986829335761)
--(axis cs:64,-1.62349168996547)
--(axis cs:128,-3.05739327111033)
--(axis cs:256,-4.93610342975314)
--(axis cs:256,-4.20681839947049)
--(axis cs:256,-4.20681839947049)
--(axis cs:128,-2.23980173430654)
--(axis cs:64,-1.12326257486607)
--(axis cs:32,-0.528547214912652)
--(axis cs:16,-0.264103722935529)
--(axis cs:8,-0.176634264788975)
--(axis cs:4,-0.0673141485341093)
--(axis cs:2,-0.0441476821830113)
--cycle;

\path [draw=color0, fill=color0, opacity=0.1]
(axis cs:2,-0.203860381613201)
--(axis cs:2,-0.363337963451915)
--(axis cs:4,-0.516284413171811)
--(axis cs:8,-0.803729523675463)
--(axis cs:16,-1.37890920189158)
--(axis cs:32,-2.05431686268656)
--(axis cs:64,-3.31102478376547)
--(axis cs:128,-4.75552404186749)
--(axis cs:256,-6.80416394498502)
--(axis cs:256,-5.54845188829745)
--(axis cs:256,-5.54845188829745)
--(axis cs:128,-4.1700079176662)
--(axis cs:64,-2.84478380807718)
--(axis cs:32,-1.87394107951315)
--(axis cs:16,-1.15035021754964)
--(axis cs:8,-0.665671823961713)
--(axis cs:4,-0.403341590331988)
--(axis cs:2,-0.203860381613201)
--cycle;

\path [draw=color1, fill=color1, opacity=0.1]
(axis cs:2,-0.35769824241529)
--(axis cs:2,-0.474182672749658)
--(axis cs:4,-0.878522877532072)
--(axis cs:8,-1.37583656000734)
--(axis cs:16,-2.28241649705629)
--(axis cs:32,-3.26989037557)
--(axis cs:64,-4.45455616987882)
--(axis cs:128,-5.4732390411027)
--(axis cs:256,-6.45125310458432)
--(axis cs:256,-5.87527353726138)
--(axis cs:256,-5.87527353726138)
--(axis cs:128,-4.9567281715743)
--(axis cs:64,-3.85857044183078)
--(axis cs:32,-2.86337759928352)
--(axis cs:16,-1.89796444338103)
--(axis cs:8,-1.2782952816522)
--(axis cs:4,-0.713081391018801)
--(axis cs:2,-0.35769824241529)
--cycle;

\path [draw=color2, fill=color2, opacity=0.1]
(axis cs:2,-0.392910723210331)
--(axis cs:2,-0.501855822085385)
--(axis cs:4,-1.03645033633983)
--(axis cs:8,-1.69274940177835)
--(axis cs:16,-2.512075857516)
--(axis cs:32,-3.39719408297284)
--(axis cs:64,-4.27004286457109)
--(axis cs:128,-5.22160025628793)
--(axis cs:256,-6.15914745181481)
--(axis cs:256,-5.5474157348307)
--(axis cs:256,-5.5474157348307)
--(axis cs:128,-4.77221049276603)
--(axis cs:64,-3.95940230678511)
--(axis cs:32,-2.91056091046588)
--(axis cs:16,-2.13771476328783)
--(axis cs:8,-1.42908881023489)
--(axis cs:4,-0.888470675584917)
--(axis cs:2,-0.392910723210331)
--cycle;

\path [draw=color3, fill=color3, opacity=0.1]
(axis cs:2,-0.51566013250234)
--(axis cs:2,-0.591944013501383)
--(axis cs:4,-1.29880367509981)
--(axis cs:8,-1.8517521354168)
--(axis cs:16,-2.63676699309711)
--(axis cs:32,-3.38319542922395)
--(axis cs:64,-4.14107035461575)
--(axis cs:128,-4.84156637109073)
--(axis cs:256,-5.50798881585509)
--(axis cs:256,-5.39481021826357)
--(axis cs:256,-5.39481021826357)
--(axis cs:128,-4.69068756186215)
--(axis cs:64,-3.87298318084568)
--(axis cs:32,-3.09607741318328)
--(axis cs:16,-2.33658820480939)
--(axis cs:8,-1.67395551476783)
--(axis cs:4,-1.05319617994169)
--(axis cs:2,-0.51566013250234)
--cycle;

\path [draw=gray, fill=gray, opacity=0.1]
(axis cs:2,-0.543316316015273)
--(axis cs:2,-0.656594992273301)
--(axis cs:4,-1.31057882536347)
--(axis cs:8,-1.84984118659873)
--(axis cs:16,-2.65997286583978)
--(axis cs:32,-3.40569031605995)
--(axis cs:64,-4.08305677678781)
--(axis cs:128,-4.84847271769379)
--(axis cs:256,-5.58114406843556)
--(axis cs:256,-5.40859353761302)
--(axis cs:256,-5.40859353761302)
--(axis cs:128,-4.6619622913089)
--(axis cs:64,-3.88581343385977)
--(axis cs:32,-3.08068511118614)
--(axis cs:16,-2.32966956351202)
--(axis cs:8,-1.78303804676156)
--(axis cs:4,-1.07205021153992)
--(axis cs:2,-0.543316316015273)
--cycle;

\addplot [semithick, black, dashed]
table {%
2 -0.0924106284976006
4 -0.180077790468931
8 -0.311548788845539
16 -0.490562033653259
32 -0.804207754135132
64 -1.37337713241577
128 -2.64859750270843
256 -4.57146091461182
};
\addplot [semithick, color0]
table {%
2 -0.283599172532558
4 -0.4598130017519
8 -0.734700673818588
16 -1.26462970972061
32 -1.96412897109985
64 -3.07790429592133
128 -4.46276597976685
256 -6.17630791664124
};
\addplot [semithick, color1]
table {%
2 -0.415940457582474
4 -0.795802134275436
8 -1.32706592082977
16 -2.09019047021866
32 -3.06663398742676
64 -4.1565633058548
128 -5.2149836063385
256 -6.16326332092285
};
\addplot [semithick, color2]
table {%
2 -0.447383272647858
4 -0.962460505962372
8 -1.56091910600662
16 -2.32489531040192
32 -3.15387749671936
64 -4.1147225856781
128 -4.99690537452698
256 -5.85328159332275
};
\addplot [semithick, color3]
table {%
2 -0.553802073001862
4 -1.17599992752075
8 -1.76285382509232
16 -2.48667759895325
32 -3.23963642120361
64 -4.00702676773071
128 -4.76612696647644
256 -5.45139951705933
};
\addplot [semithick, gray]
table {%
2 -0.599955654144287
4 -1.19131451845169
8 -1.81643961668015
16 -2.4948212146759
32 -3.24318771362305
64 -3.98443510532379
128 -4.75521750450134
256 -5.49486880302429
};
\end{axis}

\end{tikzpicture}}
        \hspace{.2cm}
        \begin{tikzpicture}

\newenvironment{customlegend}[1][]{%
    \begingroup
    \csname pgfplots@init@cleared@structures\endcsname
    \pgfplotsset{#1}%
}{%
    \csname pgfplots@createlegend\endcsname
    \endgroup
}%

\def\addlegendimage{\csname pgfplots@addlegendimage\endcsname}

\definecolor{color0}{rgb}{0.9,0.6,0}
\definecolor{color1}{rgb}{0.35,0.7,0.9}
\definecolor{color2}{rgb}{0,0.6,0.5}
\definecolor{color3}{rgb}{0.95,0.9,0.25}

\begin{customlegend}[legend entries={MAP, $\beta_k = .1$, $\beta_k = .5$, $\beta_k = 1$, $\beta_k = 5$, $\beta_k = 10$}, legend columns={1}, legend cell align=left, legend style={draw=none, font=\footnotesize, column sep=.2cm}]
\addlegendimage{black,semithick,dashed}
\addlegendimage{color0,semithick}
\addlegendimage{color1,semithick}
\addlegendimage{color2,semithick}
\addlegendimage{color3,semithick}
\addlegendimage{gray,semithick}
\end{customlegend}
\end{tikzpicture}
    \caption{Test log-likelihood values for the hypercube toy problem with prior parameters $\beta_k$ larger or smaller than 1.}
    \label{fig:hypercube_betas}
\end{figure}

Figure~\ref{fig:hypercube_betas} show the test log-likelihood of the MAP fVI model without fKL scaling after training with Dirichlet priors using varying prior concentration parameters $\beta_k$.
However, there are no significant differences when using different $\beta_k$ and no particular $\beta_k$ achieves test log-likelihoods which would be comparable to the improvements due to fKL scaling discussed in the previous subsection.

\begin{figure}[tb]
    \centering
        \begin{tabular}{c c c c c}
            $\beta_k = .1$ & $\beta_k = .5$ & $\beta_k = 1$ & $\beta_k = 5$ & $\beta_k = 10$ \\
            \includegraphics[width=.17\textwidth]{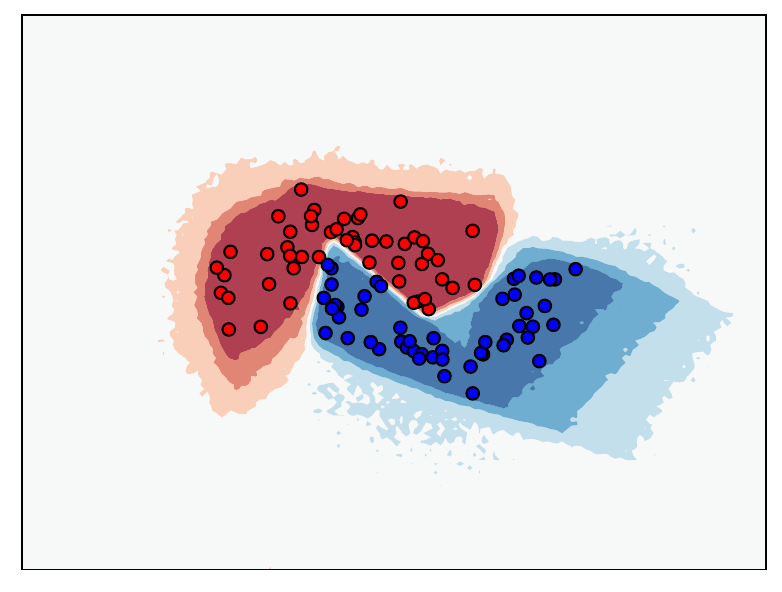} &
            \includegraphics[width=.17\textwidth]{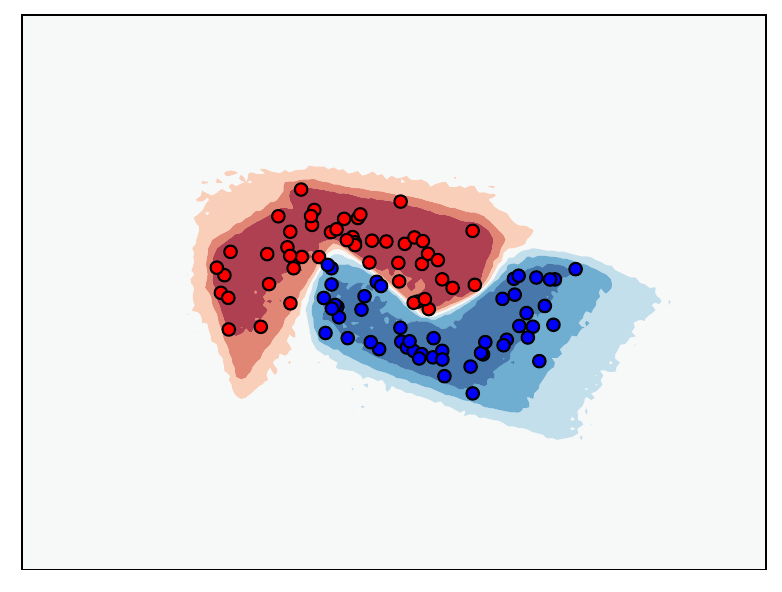} &
            \includegraphics[width=.17\textwidth]{figures/two_moons_mlp_dropout_fvi_prior_1_1_pred.pdf} &
            \includegraphics[width=.17\textwidth]{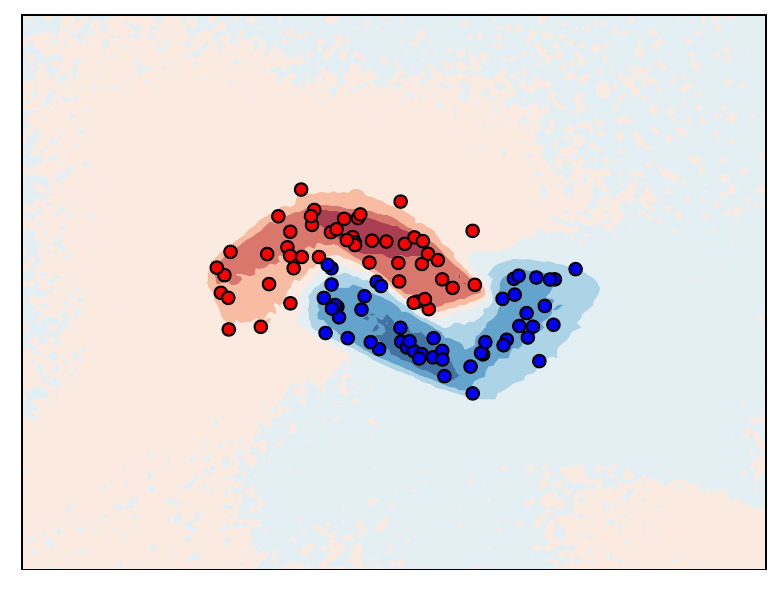} &
            \includegraphics[width=.17\textwidth]{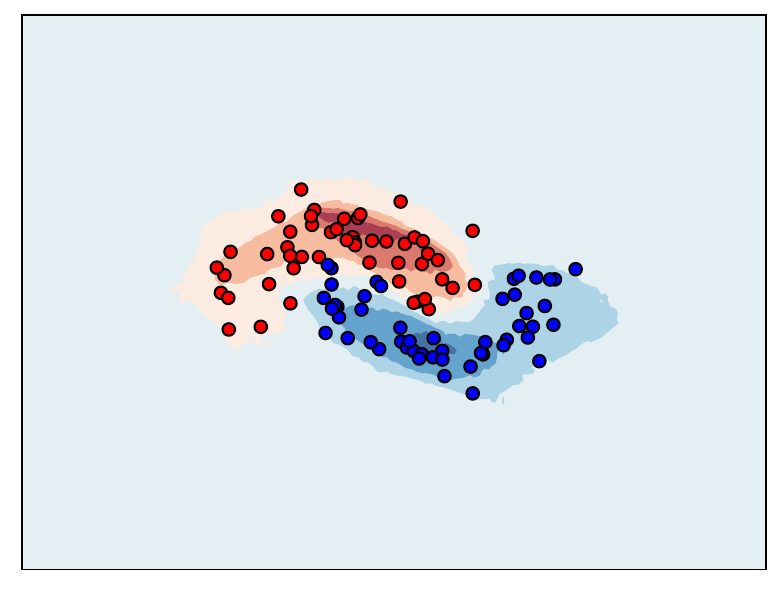}
        \end{tabular}
    \caption{Reproduction of the Two Moons toy problem (Figure \ref{fig:two_moons}) with varying uniform prior precision.}
    \label{fig:two_moons_betas}
\end{figure}

To further investigate the Dirichlet prior with different concentration parameters, we repeated the visualizable Two Moons toy problem using the MC Dropout fVI model with different $\beta_k$.
Figure~\ref{fig:two_moons_betas} depicts the predicted class probabilities of the toy problem with $K = 2$ classes.
For $\beta_k < 1$, the areas of confident prediction enlarge but quickly fall back to uniformity, whereas for $\beta_k > 1$, the confident predictions or more locally concentrated, slowly tapering towards uniformity.

\section{Approximate Gradient Computation}
\label{sec:gradient_estimation}
\begin{figure}[t]
    \centering
    \includegraphics{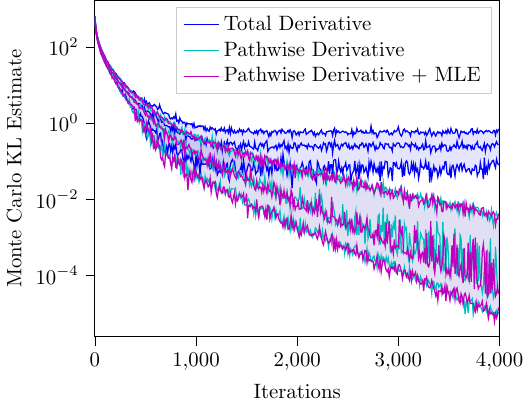}
    \caption{Comparing total and pathwise gradients for fitting a 100-dimensional Dirichlet using a Monte Carlo KL estimate, inspired by Figure 1 of \citep{roeder2017sticking}.
    The total gradient has visibly higher variance when viewed in log-space, leading the premature convergence. 
    Moreover, using an MLE fit from 5 samples for the pathwise estimator has no visible difference to the quality of the gradient estimate. 
    Results are over 50 seeds, showing the 10th, 50th and 90th percentiles.
    }
    \label{fig:pathwise_gradient}
\end{figure}
As stated in the main paper, we do not compute the total derivative of the approximate fKL divergence
\begin{align}
\KL[q || p] &\approx \frac{1}{M} \sum_{l,m=1}^{L,M} \left( \log q\left(\vf_{\vs_l}^{(m)} \middle| \boldsymbol{\theta} \right) - \log p\left(\vf_{\vs_l}^{(m)} \right) \right), \\
&= \frac{1}{M} \sum_{l,m=1}^{L,M} \left( \log q\left(\vf_{\vs_l}^{(m)}(\boldsymbol{\theta}) \middle| \boldsymbol{\alpha}\left(\vf_{\vs_l}^{(1:M)}(\boldsymbol{\theta})\right) \right) - \log p\left(\vf_{\vs_l}^{(m)}(\boldsymbol{\theta}) \right) \right), \label{eq:approx-fkl-divergence}
\end{align}
but only a partial one. Note that we have introduced the symbols $\boldsymbol{\alpha}\left(\vf_{\vs_l}^{(1:M)}(\boldsymbol{\theta})\right)$ and $\vf_{\vs_l}^{(m)}(\boldsymbol{\theta})$ to highlight the dependence of the Dirichlet posterior estimates on the $M$ implicit functions $\vf_{\vs_l}^{(m)}(\boldsymbol{\theta}),\ m \in [1, M]$ and the dependence of the individual implicit functions on the network parameters $\boldsymbol{\theta}$. The total derivative of above approximate KL divergence corresponds to
\begin{align}
\nabla_{\boldsymbol{\theta}} \KL[q || p]
&\approx
\frac{1}{M} \sum_{l,m=1}^{L,M} \left( \nabla_{\boldsymbol{\theta}} \log q\left(\vf_{\vs_l}^{(m)}(\boldsymbol{\theta}) \middle| \boldsymbol{\alpha}\left(\vf_{\vs_l}^{(1:M)}(\boldsymbol{\theta})\right) \right) - \nabla_{\boldsymbol{\theta}} \log p\left(\vf_{\vs_l}^{(m)}(\boldsymbol{\theta}) \right) \right) \nonumber \\
&=
\frac{1}{M} \sum_{l,m=1}^{L,M} \partial_{\boldsymbol{f}} \left( \log q\left(\boldsymbol{f} \middle| \boldsymbol{\alpha}\left(\vf_{\vs_l}^{(1:M)}(\boldsymbol{\theta})\right) \right) - \log p\left(\boldsymbol{f} \right) \right) \Big|_{\boldsymbol{f} =  \vf_{\vs_l}^{(m)}(\boldsymbol{\theta})} \nabla_{\boldsymbol{\theta}} \vf_{\vs_l}^{(m)}(\boldsymbol{\theta}) \nonumber \\
&\quad + \textcolor{blue}{\frac{1}{M} \sum_{l,m=1}^{L,M} \partial_{\boldsymbol{\alpha}} \log q\left(\vf_{\vs_l}^{(m)}(\boldsymbol{\theta}) \middle| \boldsymbol{\alpha} \right) \Big|_{\boldsymbol{\alpha} = \boldsymbol{\alpha}\left(\vf_{\vs_l}^{(1:M)}(\boldsymbol{\theta})\right)} \nabla_{\boldsymbol{\theta}} \boldsymbol{\alpha}\left(\vf_{\vs_l}^{(1:M)}(\boldsymbol{\theta})\right)}. \label{eq:approx-fkl-gradient}
\end{align}
The partial derivative with which we optimize the fKL divergence in our algorithm omits the blue term in (\ref{eq:approx-fkl-gradient}). This simplifies the computation graph, as $\boldsymbol{\alpha}\left(\vf_{\vs_l}^{(1:M)}(\boldsymbol{\theta})\right)$ is a maximum-likelihood estimate computed from the implicit function samples $\vf_{\vs_l}^{(m)}(\boldsymbol{\theta})$.
Computing $\nabla_{\boldsymbol{\theta}} \boldsymbol{\alpha}\left(\vf_{\vs_l}^{(1:M)}(\boldsymbol{\theta})\right)$ requires to compute the gradient of a maximum-likelihood estimate (MLE)  $\boldsymbol{\alpha}$ w.r.t. the implicit functions $\boldsymbol{f}$.
Given that we use an iterative scheme to compute an approximate MLE, this would require us to differentiate through each iteration of the MLE computation.

We now want to provide evidence that using this partial derivative of the approximate fKL divergence is still a reasonable choice.
As argued in \citep{roeder2017sticking}, terms of the form $\mathbb{E}_{p(x | \boldsymbol{\theta})} \left[ \partial_{\boldsymbol{\theta}} \log p(x | \boldsymbol{\theta}) \right] = 0$ tend to introduce high variance into the gradient of variational inference objectives due to the Monte Carlo expectation approximation.
Therefore, omitting that term from the gradient estimate can actually benefit the convergence of gradient-based variational inference methods in certain cases.
We visualize this phenomena in Figure \ref{fig:pathwise_gradient} for fitting a high-dimensional Dirichlet, in which the gradient-based optimization of a KL divergence objective using the total derivative prematurely converges due to high variance, while an optimization that ignores the variance-inducing terms does not face this problem.
Moreover, the MLE fit of the variational density using 5 samples has no visible effect of the gradient estimation quality, despite fitting a 100-dimensional distribution. 
This result indicates that as long as the predictive distribution is approximately Dirichlet, which is a central assumption of this approach, the gradient assumption is reasonable. 

\begin{figure}[tb]
    \centering
    \includegraphics[width=.49\textwidth]{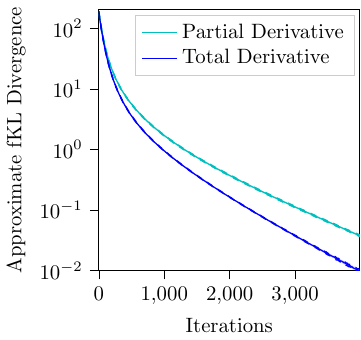}
    \includegraphics[width=.49\textwidth]{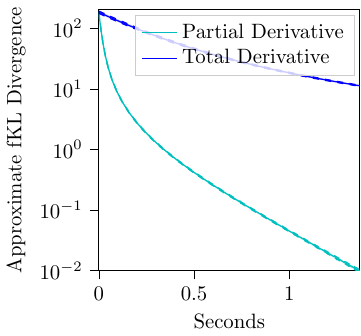}
    \caption{A comparison of the total- and partial derivative for minimizing approximate fKL divergence (\ref{eq:approx-fkl-divergence}) of an empirical distribution $q(\boldsymbol{f}) \approx \frac{1}{M} \sum_{m=1}^M \delta_{\boldsymbol{f}^{(m)}}(\boldsymbol{f}),\ \boldsymbol{f}^{(m)} \sim q(\boldsymbol{f})$ represented by $M{\,=\,}5$ samples to a $100$-dimensional Dirichlet prior $p$ w.r.t. the samples $\boldsymbol{f}^{(m)}$. The curves show the fKL divergence over gradient steps (left) and time (right) using the total and partial derivative. The partial derivative (that ignores the blue term in (\ref{eq:approx-fkl-gradient})) leads to a slower decline in fKL divergence per gradient step. However, given the much cheaper computational cost of the partial derivative, it leads to a faster optimization of the fKL divergence w.r.t.\ time. Results are computed over 50 seeds, showing the 10th, 50th and 90th percentiles. For each seed, the samples $\boldsymbol{f}^{(m)}$ were intialized differently.}
    \label{fig:approximate_gradient}
\end{figure}

However, our method does not exactly match the pathwise gradient of \cite{roeder2017sticking}.
The ignored term of the total approximate fKL divergence derivative (\ref{eq:approx-fkl-gradient}) resembles the variance-inducing term, as the $M$ implicit functions $\vf^{(m)}(\boldsymbol{\theta})$ evaluated at the different elements $\vs$ of the measurement set $\mathcal{S}$ approximate the expectation over $q\left(\vf \middle| \boldsymbol{\alpha} \right)$
\begin{align}
    \frac{1}{M} \sum_{m=1}^{M} \partial_{\boldsymbol{\alpha}} \log q\left(\vf_{\vs}^{(m)}(\boldsymbol{\theta}) \middle| \boldsymbol{\alpha} \right) \approx \mathbb{E}_{q\left(\vf \middle| \boldsymbol{\alpha} \right)} \left[ \partial_{\boldsymbol{\alpha}} \log q\left(\vf \middle| \boldsymbol{\alpha} \right) \right] = 0.
\end{align}
However, the implicit function samples $\vf_{\vs}^{(m)}(\boldsymbol{\theta})$ are clearly not i.i.d. samples
as there is a tight correlation between the parameter $\boldsymbol{\alpha}$ of the Dirichlet distribution $q$ and the implicit function samples. 
Moreover, our ensemble model does not use reparameterized gradients, optimizing a set of network weight `particles' instead.
Therefore, we also compared the optimization of the approximate fKL divergence objective using the partial- and total derivative on a particle-based variational representation of the Dirichlet.
To compute the gradient of the MLE w.r.t. the implicit function samples required by the total derivative, we first compute the MLE $\boldsymbol{\alpha}$ by solving the underlying convex optimization problem \citep{minka2000estimating} using the \texttt{CVXOpt} library \citep{cvxopt}.
We then leverage the implicit function theorem \citep{dontchev2009implicit} to compute the gradient of $\boldsymbol{\alpha}$ w.r.t. the samples, leveraging that the gradient of the likelihood function vanishes for the MLE
\begin{align*}
    \mathbf{0} = \nabla_{\boldsymbol{f}^{(1:M)}} \log\left(p(\boldsymbol{f}^{(1:M)} \vert \boldsymbol{\alpha})\right).
\end{align*}
The results in Figure \ref{fig:approximate_gradient} indeed highlight that in the setting of this paper, the total derivative leads to a faster descent along the fKL divergence objective per gradient step when optimizing particles.
However, we see that the partial derivatives also minimize the fKL divergence.
Furthermore, due to the lower computational overhead of the partial derivatives, this optimization is carried out in significantly less time.

\newpage
\section{Experimental Results}
Numerical values of all our experimental results are reported here.
\begin{table}[h!]
\caption{Accuracies for the rotated MNIST experiment. Means and standard errors over ten seeds. Best results within archetype in boldface, best results overall in blue.}
\resizebox{\textwidth}{!}{%
\begin{tabular}{ l c c c c c c c }
\toprule
\multicolumn{1}{c}{MNIST} & \multicolumn{7}{c}{Angle \textdegree} \\
\multicolumn{1}{c}{Accuracy $\uparrow$} & 0 & 10 & 20 & 30 & 40 & 50 & 60 \\
\midrule
MAP & $96.82 \pm 0.06$ & $94.19 \pm 0.15$ & $87.17 \pm 0.43$ & $71.60 \pm 0.76$ & $53.11 \pm 0.76$ & $\mathbf{36.03 \pm 0.52}$ & $\mathbf{25.18 \pm 0.46}$ \\
MAP fVI & $\mathbf{97.09 \pm 0.05}$ & $\mathbf{95.01 \pm 0.13}$ & $\mathbf{88.51 \pm 0.32}$ & $\mathbf{73.53 \pm 0.51}$ & $\mathbf{54.32 \pm 0.57}$ & $36.03 \pm 0.40$ & $24.56 \pm 0.35$ \\
\midrule
MC Dropout & $\mathbf{96.23 \pm 0.05}$ & $93.63 \pm 0.16$ & $\mathbf{86.77 \pm 0.16}$ & $\mathbf{71.06 \pm 0.28}$ & $\mathbf{52.46 \pm 0.48}$ & $\mathbf{35.57 \pm 0.56}$ & $25.28 \pm 0.54$ \\
MC Dropout fVI & $96.10 \pm 0.08$ & $\mathbf{93.70 \pm 0.17}$ & $86.61 \pm 0.42$ & $71.04 \pm 0.74$ & $52.20 \pm 0.71$ & $34.94 \pm 0.55$ & $\mathbf{25.38 \pm 0.33}$ \\
\midrule
Ensemble & $97.97 \pm 0.02$ & $96.36 \pm 0.04$ & $91.02 \pm 0.18$ & $76.98 \pm 0.25$ & $57.96 \pm 0.29$ & $39.52 \pm 0.27$ & $27.94 \pm 0.32$ \\
Ensemble fVI & {\color{blue}$\mathbf{98.09 \pm 0.02}$} & {\color{blue}$\mathbf{96.71 \pm 0.02}$} & {\color{blue}$\mathbf{91.95 \pm 0.13}$} & {\color{blue}$\mathbf{78.87 \pm 0.23}$} & {\color{blue}$\mathbf{60.08 \pm 0.31}$} & {\color{blue}$\mathbf{40.81 \pm 0.22}$} & {\color{blue}$\mathbf{28.20 \pm 0.29}$} \\
\bottomrule
\end{tabular}}
\end{table}

\begin{table}[h!]
\caption{Accuracies for the rotated MNIST experiment. Means and standard errors over ten seeds. Best results within archetype in boldface, best results overall in blue.}
\resizebox{\textwidth}{!}{%
\begin{tabular}{ l c c c c c c }
\toprule
\multicolumn{1}{c}{MNIST} & \multicolumn{6}{c}{Angle \textdegree} \\
\multicolumn{1}{c}{Accuracy $\uparrow$} & 70 & 80 & 90 & 100 & 110 & 120 \\
\midrule
MAP & $\mathbf{18.60 \pm 0.39}$ & $\mathbf{15.01 \pm 0.34}$ & $12.81 \pm 0.46$ & $11.77 \pm 0.52$ & $11.66 \pm 0.49$ & $14.09 \pm 0.33$ \\
MAP fVI & $17.91 \pm 0.42$ & $14.72 \pm 0.47$ & $\mathbf{13.33 \pm 0.48}$ & $\mathbf{13.02 \pm 0.40}$ & $\mathbf{13.33 \pm 0.28}$ & $\mathbf{15.38 \pm 0.21}$ \\
\midrule
MC Dropout & $20.26 \pm 0.48$ & $17.22 \pm 0.47$ & $15.11 \pm 0.43$ & $13.41 \pm 0.38$ & $12.67 \pm 0.35$ & $14.32 \pm 0.23$ \\
MC Dropout fVI & {\color{blue}$\mathbf{20.99 \pm 0.39}$} & {\color{blue}$\mathbf{18.22 \pm 0.34}$} & {\color{blue}$\mathbf{16.26 \pm 0.39}$} & {\color{blue}$\mathbf{14.82 \pm 0.46}$} & {\color{blue}$\mathbf{13.69 \pm 0.38}$} & $\mathbf{14.44 \pm 0.29}$ \\
\midrule
Ensemble & $19.90 \pm 0.31$ & $16.11 \pm 0.32$ & $13.62 \pm 0.31$ & $12.09 \pm 0.26$ & $12.31 \pm 0.22$ & $14.92 \pm 0.21$ \\
Ensemble fVI & $\mathbf{20.08 \pm 0.38}$ & $\mathbf{16.43 \pm 0.37}$ & $\mathbf{14.28 \pm 0.30}$ & $\mathbf{13.07 \pm 0.21}$ & $\mathbf{13.19 \pm 0.16}$ & {\color{blue}$\mathbf{15.43 \pm 0.12}$} \\
\bottomrule
\end{tabular}}
\end{table}

\begin{table}[h!]
\caption{Accuracies for the rotated MNIST experiment. Means and standard errors over ten seeds. Best results within archetype in boldface, best results overall in blue.}
\resizebox{\textwidth}{!}{%
\begin{tabular}{ l c c c c c c }
\toprule
\multicolumn{1}{c}{MNIST} & \multicolumn{6}{c}{Angle \textdegree} \\
\multicolumn{1}{c}{Accuracy $\uparrow$} & 130 & 140 & 150 & 160 & 170 & 180 \\
\midrule
MAP & $16.92 \pm 0.43$ & $19.05 \pm 0.56$ & $22.36 \pm 0.55$ & $25.06 \pm 0.64$ & $26.95 \pm 0.58$ & $28.58 \pm 0.57$ \\
MAP fVI & $\mathbf{17.94 \pm 0.29}$ & $\mathbf{19.82 \pm 0.43}$ & $\mathbf{22.45 \pm 0.54}$ & $\mathbf{25.26 \pm 0.46}$ & $\mathbf{27.42 \pm 0.43}$ & $\mathbf{28.93 \pm 0.42}$ \\
\midrule
MC Dropout & $\mathbf{17.35 \pm 0.34}$ & $20.11 \pm 0.49$ & $22.82 \pm 0.43$ & $24.37 \pm 0.39$ & $26.03 \pm 0.43$ & $28.14 \pm 0.43$ \\
MC Dropout fVI & $17.12 \pm 0.35$ & $\mathbf{20.25 \pm 0.40}$ & $\mathbf{23.25 \pm 0.34}$ & $\mathbf{25.24 \pm 0.27}$ & $\mathbf{26.86 \pm 0.18}$ & $\mathbf{28.95 \pm 0.29}$ \\
\midrule
Ensemble & $18.02 \pm 0.32$ & $20.83 \pm 0.29$ & $24.49 \pm 0.26$ & $27.61 \pm 0.20$ & $30.03 \pm 0.24$ & $31.27 \pm 0.32$ \\
Ensemble fVI & {\color{blue}$\mathbf{18.47 \pm 0.19}$} & {\color{blue}$\mathbf{21.49 \pm 0.20}$} & {\color{blue}$\mathbf{25.21 \pm 0.21}$} & {\color{blue}$\mathbf{28.54 \pm 0.24}$} & {\color{blue}$\mathbf{31.31 \pm 0.29}$} & {\color{blue}$\mathbf{32.51 \pm 0.31}$} \\
\bottomrule
\end{tabular}}
\end{table}

\begin{table}[h!]
\caption{Log-likelihoods for the rotated MNIST experiment. Means and standard errors over ten seeds. Best results within archetype in boldface, best results overall in blue.}
\resizebox{\textwidth}{!}{%
\begin{tabular}{ l c c c c c c c }
\toprule
\multicolumn{1}{c}{MNIST} & \multicolumn{7}{c}{Angle \textdegree} \\
\multicolumn{1}{c}{Log-Likelihood $\uparrow$} & 0 & 10 & 20 & 30 & 40 & 50 & 60 \\
\midrule
MAP & $\mathbf{-0.11 \pm 0.00}$ & $\mathbf{-0.20 \pm 0.00}$ & $-0.48 \pm 0.02$ & $-1.20 \pm 0.03$ & $-2.36 \pm 0.04$ & $-3.80 \pm 0.04$ & $-5.05 \pm 0.05$ \\
MAP fVI & $-0.13 \pm 0.00$ & $-0.20 \pm 0.00$ & $\mathbf{-0.39 \pm 0.01}$ & $\mathbf{-0.84 \pm 0.02}$ & $\mathbf{-1.53 \pm 0.02}$ & $\mathbf{-2.34 \pm 0.03}$ & $\mathbf{-3.02 \pm 0.03}$ \\
\midrule
MC Dropout & $\mathbf{-0.15 \pm 0.00}$ & $\mathbf{-0.23 \pm 0.00}$ & $\mathbf{-0.43 \pm 0.00}$ & $\mathbf{-0.88 \pm 0.01}$ & $-1.50 \pm 0.02$ & $-2.28 \pm 0.03$ & $-2.97 \pm 0.04$ \\
MC Dropout fVI & $-0.20 \pm 0.00$ & $-0.28 \pm 0.01$ & $-0.48 \pm 0.01$ & $-0.91 \pm 0.02$ & $\mathbf{-1.46 \pm 0.02}$ & $\mathbf{-2.06 \pm 0.02}$ & $\mathbf{-2.50 \pm 0.02}$ \\
\midrule
Ensemble & {\color{blue}$\mathbf{-0.07 \pm 0.00}$} & {\color{blue}$\mathbf{-0.12 \pm 0.00}$} & {\color{blue}$\mathbf{-0.29 \pm 0.00}$} & $-0.73 \pm 0.01$ & $-1.50 \pm 0.02$ & $-2.56 \pm 0.03$ & $-3.60 \pm 0.04$ \\
Ensemble fVI & $-0.11 \pm 0.00$ & $-0.17 \pm 0.00$ & $-0.32 \pm 0.00$ & {\color{blue}$\mathbf{-0.67 \pm 0.01}$} & {\color{blue}$\mathbf{-1.20 \pm 0.01}$} & {\color{blue}$\mathbf{-1.85 \pm 0.01}$} & {\color{blue}$\mathbf{-2.43 \pm 0.01}$} \\
\bottomrule
\end{tabular}}
\end{table}

\begin{table}[h!]
\caption{Log-likelihoods for the rotated MNIST experiment. Means and standard errors over ten seeds. Best results within archetype in boldface, best results overall in blue.}
\resizebox{\textwidth}{!}{%
\begin{tabular}{ l c c c c c c }
\toprule
\multicolumn{1}{c}{MNIST} & \multicolumn{6}{c}{Angle \textdegree} \\
\multicolumn{1}{c}{Log-Likelihood $\uparrow$} & 70 & 80 & 90 & 100 & 110 & 120 \\
\midrule
MAP & $-6.05 \pm 0.06$ & $-6.79 \pm 0.07$ & $-7.22 \pm 0.09$ & $-7.34 \pm 0.09$ & $-7.27 \pm 0.10$ & $-6.95 \pm 0.09$ \\
MAP fVI & $\mathbf{-3.54 \pm 0.06}$ & $\mathbf{-3.94 \pm 0.07}$ & $\mathbf{-4.14 \pm 0.08}$ & $\mathbf{-4.18 \pm 0.08}$ & $\mathbf{-4.16 \pm 0.07}$ & $\mathbf{-4.05 \pm 0.06}$ \\
\midrule
MC Dropout & $-3.47 \pm 0.04$ & $-3.88 \pm 0.04$ & $-4.17 \pm 0.05$ & $-4.28 \pm 0.05$ & $-4.32 \pm 0.05$ & $-4.15 \pm 0.05$ \\
MC Dropout fVI & {\color{blue}$\mathbf{-2.81 \pm 0.02}$} & {\color{blue}$\mathbf{-3.04 \pm 0.03}$} & {\color{blue}$\mathbf{-3.20 \pm 0.04}$} & {\color{blue}$\mathbf{-3.25 \pm 0.04}$} & {\color{blue}$\mathbf{-3.28 \pm 0.04}$} & {\color{blue}$\mathbf{-3.17 \pm 0.03}$} \\
\midrule
Ensemble & $-4.52 \pm 0.04$ & $-5.31 \pm 0.04$ & $-5.77 \pm 0.05$ & $-5.90 \pm 0.06$ & $-5.86 \pm 0.06$ & $-5.68 \pm 0.06$ \\
Ensemble fVI & $\mathbf{-2.89 \pm 0.02}$ & $\mathbf{-3.27 \pm 0.02}$ & $\mathbf{-3.49 \pm 0.02}$ & $\mathbf{-3.57 \pm 0.02}$ & $\mathbf{-3.57 \pm 0.02}$ & $\mathbf{-3.47 \pm 0.02}$ \\
\bottomrule
\end{tabular}}
\end{table}

\begin{table}[h!]
\caption{Log-likelihoods for the rotated MNIST experiment. Means and standard errors over ten seeds. Best results within archetype in boldface, best results overall in blue.}
\resizebox{\textwidth}{!}{%
\begin{tabular}{ l c c c c c c }
\toprule
\multicolumn{1}{c}{MNIST} & \multicolumn{6}{c}{Angle \textdegree} \\
\multicolumn{1}{c}{Log-Likelihood $\uparrow$} & 130 & 140 & 150 & 160 & 170 & 180 \\
\midrule
MAP & $-6.57 \pm 0.09$ & $-6.36 \pm 0.08$ & $-6.20 \pm 0.07$ & $-6.16 \pm 0.06$ & $-6.13 \pm 0.05$ & $-6.00 \pm 0.07$ \\
MAP fVI & $\mathbf{-3.87 \pm 0.05}$ & $\mathbf{-3.78 \pm 0.04}$ & $\mathbf{-3.68 \pm 0.04}$ & $\mathbf{-3.63 \pm 0.03}$ & $\mathbf{-3.61 \pm 0.04}$ & $\mathbf{-3.50 \pm 0.04}$ \\
\midrule
MC Dropout & $-3.93 \pm 0.05$ & $-3.85 \pm 0.06$ & $-3.87 \pm 0.05$ & $-4.04 \pm 0.04$ & $-4.15 \pm 0.04$ & $-4.19 \pm 0.04$ \\
MC Dropout fVI & {\color{blue}$\mathbf{-3.03 \pm 0.02}$} & {\color{blue}$\mathbf{-2.94 \pm 0.02}$} & {\color{blue}$\mathbf{-2.91 \pm 0.02}$} & {\color{blue}$\mathbf{-2.97 \pm 0.02}$} & {\color{blue}$\mathbf{-3.00 \pm 0.01}$} & {\color{blue}$\mathbf{-2.98 \pm 0.01}$} \\
\midrule
Ensemble & $-5.37 \pm 0.05$ & $-5.24 \pm 0.04$ & $-5.18 \pm 0.02$ & $-5.31 \pm 0.01$ & $-5.37 \pm 0.02$ & $-5.25 \pm 0.02$ \\
Ensemble fVI & $\mathbf{-3.29 \pm 0.02}$ & $\mathbf{-3.22 \pm 0.01}$ & $\mathbf{-3.17 \pm 0.01}$ & $\mathbf{-3.21 \pm 0.02}$ & $\mathbf{-3.23 \pm 0.02}$ & $\mathbf{-3.16 \pm 0.02}$ \\
\bottomrule
\end{tabular}}
\end{table}

\begin{table}[h!]
\caption{Expected calibration errors for the rotated MNIST experiment. Means and standard errors over ten seeds. Best results within archetype in boldface, best results overall in blue.}
\resizebox{\textwidth}{!}{%
\begin{tabular}{ l c c c c c c c }
\toprule
\multicolumn{1}{c}{MNIST} & \multicolumn{7}{c}{Angle \textdegree} \\
\multicolumn{1}{c}{ECE $\downarrow$} & 0 & 10 & 20 & 30 & 40 & 50 & 60 \\
\midrule
MAP & {\color{blue}$\mathbf{0.01 \pm 0.00}$} & $\mathbf{0.02 \pm 0.00}$ & $0.06 \pm 0.00$ & $0.17 \pm 0.01$ & $0.32 \pm 0.01$ & $0.46 \pm 0.00$ & $0.56 \pm 0.00$ \\
MAP fVI & $0.05 \pm 0.00$ & $0.05 \pm 0.00$ & $\mathbf{0.04 \pm 0.00}$ & $\mathbf{0.03 \pm 0.00}$ & $\mathbf{0.15 \pm 0.01}$ & $\mathbf{0.28 \pm 0.01}$ & $\mathbf{0.38 \pm 0.01}$ \\
\midrule
MC Dropout & $\mathbf{0.04 \pm 0.00}$ & $\mathbf{0.05 \pm 0.00}$ & $\mathbf{0.05 \pm 0.00}$ & {\color{blue}$\mathbf{0.02 \pm 0.00}$} & $0.13 \pm 0.01$ & $0.25 \pm 0.01$ & $0.33 \pm 0.01$ \\
MC Dropout fVI & $0.09 \pm 0.00$ & $0.10 \pm 0.00$ & $0.10 \pm 0.00$ & $0.04 \pm 0.01$ & $\mathbf{0.07 \pm 0.01}$ & $\mathbf{0.19 \pm 0.01}$ & $\mathbf{0.27 \pm 0.01}$ \\
\midrule
Ensemble & $\mathbf{0.01 \pm 0.00}$ & {\color{blue}$\mathbf{0.02 \pm 0.00}$} & {\color{blue}$\mathbf{0.01 \pm 0.00}$} & $\mathbf{0.04 \pm 0.00}$ & $0.15 \pm 0.00$ & $0.27 \pm 0.00$ & $0.36 \pm 0.00$ \\
Ensemble fVI & $0.06 \pm 0.00$ & $0.08 \pm 0.00$ & $0.10 \pm 0.00$ & $0.08 \pm 0.00$ & {\color{blue}$\mathbf{0.02 \pm 0.00}$} & {\color{blue}$\mathbf{0.14 \pm 0.00}$} & {\color{blue}$\mathbf{0.23 \pm 0.00}$} \\
\bottomrule
\end{tabular}}
\end{table}

\begin{table}[h!]
\caption{Expected calibration errors for the rotated MNIST experiment. Means and standard errors over ten seeds. Best results within archetype in boldface, best results overall in blue.}
\resizebox{\textwidth}{!}{%
\begin{tabular}{ l c c c c c c }
\toprule
\multicolumn{1}{c}{MNIST} & \multicolumn{6}{c}{Angle \textdegree} \\
\multicolumn{1}{c}{ECE $\downarrow$} & 70 & 80 & 90 & 100 & 110 & 120 \\
\midrule
MAP & $0.62 \pm 0.00$ & $0.65 \pm 0.01$ & $0.68 \pm 0.01$ & $0.69 \pm 0.01$ & $0.69 \pm 0.01$ & $0.67 \pm 0.01$ \\
MAP fVI & $\mathbf{0.44 \pm 0.01}$ & $\mathbf{0.48 \pm 0.01}$ & $\mathbf{0.50 \pm 0.01}$ & $\mathbf{0.50 \pm 0.01}$ & $\mathbf{0.51 \pm 0.01}$ & $\mathbf{0.50 \pm 0.00}$ \\
\midrule
MC Dropout & $0.39 \pm 0.01$ & $0.43 \pm 0.01$ & $0.46 \pm 0.01$ & $0.48 \pm 0.01$ & $0.50 \pm 0.01$ & $0.48 \pm 0.01$ \\
MC Dropout fVI & $\mathbf{0.32 \pm 0.01}$ & $\mathbf{0.36 \pm 0.01}$ & $\mathbf{0.38 \pm 0.01}$ & {\color{blue}$\mathbf{0.40 \pm 0.01}$} & $\mathbf{0.42 \pm 0.01}$ & $\mathbf{0.40 \pm 0.01}$ \\
\midrule
Ensemble & $0.43 \pm 0.00$ & $0.46 \pm 0.00$ & $0.49 \pm 0.00$ & $0.51 \pm 0.01$ & $0.52 \pm 0.00$ & $0.50 \pm 0.00$ \\
Ensemble fVI & {\color{blue}$\mathbf{0.31 \pm 0.00}$} & {\color{blue}$\mathbf{0.35 \pm 0.00}$} & {\color{blue}$\mathbf{0.38 \pm 0.00}$} & $\mathbf{0.40 \pm 0.00}$ & {\color{blue}$\mathbf{0.41 \pm 0.00}$} & {\color{blue}$\mathbf{0.40 \pm 0.00}$} \\
\bottomrule
\end{tabular}}
\end{table}

\begin{table}[h!]
\caption{Expected calibration errors for the rotated MNIST experiment. Means and standard errors over ten seeds. Best results within archetype in boldface, best results overall in blue.}
\resizebox{\textwidth}{!}{%
\begin{tabular}{ l c c c c c c }
\toprule
\multicolumn{1}{c}{MNIST} & \multicolumn{6}{c}{Angle \textdegree} \\
\multicolumn{1}{c}{ECE $\downarrow$} & 130 & 140 & 150 & 160 & 170 & 180 \\
\midrule
MAP & $0.65 \pm 0.01$ & $0.63 \pm 0.01$ & $0.60 \pm 0.01$ & $0.58 \pm 0.01$ & $0.57 \pm 0.01$ & $0.56 \pm 0.01$ \\
MAP fVI & $\mathbf{0.48 \pm 0.00}$ & $\mathbf{0.47 \pm 0.01}$ & $\mathbf{0.45 \pm 0.01}$ & $\mathbf{0.43 \pm 0.01}$ & $\mathbf{0.41 \pm 0.01}$ & $\mathbf{0.40 \pm 0.01}$ \\
\midrule
MC Dropout & $0.45 \pm 0.01$ & $0.42 \pm 0.01$ & $0.41 \pm 0.01$ & $0.41 \pm 0.01$ & $0.41 \pm 0.01$ & $0.40 \pm 0.00$ \\
MC Dropout fVI & {\color{blue}$\mathbf{0.37 \pm 0.00}$} & {\color{blue}$\mathbf{0.35 \pm 0.00}$} & $\mathbf{0.33 \pm 0.00}$ & $\mathbf{0.34 \pm 0.00}$ & $\mathbf{0.33 \pm 0.00}$ & $\mathbf{0.32 \pm 0.00}$ \\
\midrule
Ensemble & $0.48 \pm 0.00$ & $0.46 \pm 0.00$ & $0.44 \pm 0.00$ & $0.43 \pm 0.00$ & $0.42 \pm 0.00$ & $0.42 \pm 0.00$ \\
Ensemble fVI & $\mathbf{0.38 \pm 0.00}$ & $\mathbf{0.35 \pm 0.00}$ & {\color{blue}$\mathbf{0.33 \pm 0.00}$} & {\color{blue}$\mathbf{0.32 \pm 0.00}$} & {\color{blue}$\mathbf{0.31 \pm 0.00}$} & {\color{blue}$\mathbf{0.30 \pm 0.00}$} \\
\bottomrule
\end{tabular}}
\end{table}

\begin{table}[h!]
\caption{Accuracies for the corrupted CIFAR10 experiment. Means and standard errors over ten seeds. Best results within archetype in boldface, best results overall in blue.}
\resizebox{\textwidth}{!}{%
\begin{tabular}{ l c c c c c c }
\toprule
\multicolumn{1}{c}{CIFAR10} & \multicolumn{6}{c}{Corruption Severity} \\
\multicolumn{1}{c}{Accuracy $\uparrow$} & 0 & 1 & 2 & 3 & 4 & 5 \\
\midrule
MAP & $94.32 \pm 0.05$ & $87.65 \pm 0.08$ & $\mathbf{81.76 \pm 0.09}$ & $\mathbf{75.97 \pm 0.13}$ & $\mathbf{68.86 \pm 0.20}$ & $\mathbf{57.28 \pm 0.21}$ \\
MAP fVI & $\mathbf{94.40 \pm 0.08}$ & $\mathbf{87.66 \pm 0.06}$ & $81.66 \pm 0.10$ & $75.81 \pm 0.16$ & $68.77 \pm 0.17$ & $57.06 \pm 0.18$ \\
\midrule
MC Dropout & $\mathbf{94.32 \pm 0.04}$ & $\mathbf{88.21 \pm 0.07}$ & $\mathbf{82.13 \pm 0.13}$ & $\mathbf{75.81 \pm 0.19}$ & $\mathbf{67.59 \pm 0.21}$ & $\mathbf{55.72 \pm 0.26}$ \\
MC Dropout fVI & $93.38 \pm 0.03$ & $87.01 \pm 0.09$ & $80.64 \pm 0.14$ & $74.36 \pm 0.18$ & $66.33 \pm 0.18$ & $54.87 \pm 0.19$ \\
\midrule
Ensemble & {\color{blue}$\mathbf{95.30 \pm 0.04}$} & $89.37 \pm 0.03$ & $83.77 \pm 0.06$ & $78.21 \pm 0.07$ & $71.05 \pm 0.11$ & $59.33 \pm 0.16$ \\
Ensemble fVI & $95.26 \pm 0.03$ & {\color{blue}$\mathbf{89.44 \pm 0.03}$} & {\color{blue}$\mathbf{83.94 \pm 0.07}$} & {\color{blue}$\mathbf{78.49 \pm 0.08}$} & {\color{blue}$\mathbf{71.42 \pm 0.12}$} & {\color{blue}$\mathbf{59.73 \pm 0.16}$} \\
\midrule
Radial & $\mathbf{95.05 \pm 0.04}$ & $\mathbf{87.98 \pm 0.07}$ & $\mathbf{81.82 \pm 0.10}$ & $75.93 \pm 0.12$ & $\mathbf{68.93 \pm 0.16}$ & $57.24 \pm 0.21$ \\
Radial fVI & $93.73 \pm 0.03$ & $87.43 \pm 0.07$ & $81.75 \pm 0.14$ & $\mathbf{76.09 \pm 0.21}$ & $68.84 \pm 0.29$ & $\mathbf{57.42 \pm 0.35}$ \\
\midrule
Rank1 & $93.68 \pm 0.05$ & $87.60 \pm 0.05$ & $82.32 \pm 0.08$ & $\mathbf{76.90 \pm 0.08}$ & $\mathbf{69.92 \pm 0.13}$ & $\mathbf{58.53 \pm 0.17}$ \\
Rank1 fVI & $\mathbf{93.91 \pm 0.04}$ & $\mathbf{87.75 \pm 0.05}$ & $\mathbf{82.38 \pm 0.09}$ & $76.77 \pm 0.14$ & $69.69 \pm 0.17$ & $58.23 \pm 0.19$ \\
\midrule
Subnetwork & $91.00 \pm 0.00$ & $83.00 \pm 1.00$ & $77.00 \pm 0.00$ & $68.00 \pm 1.00$ & $64.00 \pm 1.00$ & $59.00 \pm 0.00$ \\
Belief Matching & $94.52 \pm 0.03$ & $86.98 \pm 0.12$ & $80.39 \pm 0.21$ & $73.62 \pm 0.29$ & $65.74 \pm 0.35$ & $53.67 \pm 0.36$ \\
Prior Networks & $66.65 \pm 0.61$ & $61.28 \pm 0.43$ & $57.87 \pm 0.37$ & $54.71 \pm 0.35$ & $50.24 \pm 0.33$ & $42.29 \pm 0.31$ \\
\bottomrule
\end{tabular}}
\end{table}

\begin{table}[h!]
\caption{Log-likelihoods for the corrupted CIFAR10 experiment. Means and standard errors over ten seeds. Best results within archetype in boldface, best results overall in blue.}
\resizebox{\textwidth}{!}{%
\begin{tabular}{ l c c c c c c }
\toprule
\multicolumn{1}{c}{CIFAR10} & \multicolumn{6}{c}{Corruption Severity} \\
\multicolumn{1}{c}{Log-Likelihood $\uparrow$} & 0 & 1 & 2 & 3 & 4 & 5 \\
\midrule
MAP & $\mathbf{-0.22 \pm 0.00}$ & $-0.52 \pm 0.00$ & $-0.80 \pm 0.01$ & $-1.12 \pm 0.01$ & $-1.52 \pm 0.01$ & $-2.20 \pm 0.02$ \\
MAP fVI & $-0.25 \pm 0.00$ & $\mathbf{-0.48 \pm 0.00}$ & $\mathbf{-0.69 \pm 0.00}$ & $\mathbf{-0.90 \pm 0.01}$ & $\mathbf{-1.16 \pm 0.01}$ & $\mathbf{-1.60 \pm 0.01}$ \\
\midrule
MC Dropout & $\mathbf{-0.17 \pm 0.00}$ & $\mathbf{-0.39 \pm 0.00}$ & $\mathbf{-0.63 \pm 0.01}$ & $-0.93 \pm 0.01$ & $-1.36 \pm 0.02$ & $-2.09 \pm 0.02$ \\
MC Dropout fVI & $-0.25 \pm 0.00$ & $-0.44 \pm 0.00$ & $-0.64 \pm 0.01$ & $\mathbf{-0.85 \pm 0.01}$ & $\mathbf{-1.12 \pm 0.01}$ & $\mathbf{-1.56 \pm 0.01}$ \\
\midrule
Ensemble & {\color{blue}$\mathbf{-0.15 \pm 0.00}$} & {\color{blue}$\mathbf{-0.35 \pm 0.00}$} & {\color{blue}$\mathbf{-0.54 \pm 0.00}$} & $-0.76 \pm 0.00$ & $-1.03 \pm 0.01$ & $-1.51 \pm 0.01$ \\
Ensemble fVI & $-0.21 \pm 0.00$ & $-0.38 \pm 0.00$ & $-0.55 \pm 0.00$ & {\color{blue}$\mathbf{-0.72 \pm 0.00}$} & {\color{blue}$\mathbf{-0.94 \pm 0.00}$} & {\color{blue}$\mathbf{-1.33 \pm 0.01}$} \\
\midrule
Radial & $\mathbf{-0.21 \pm 0.00}$ & $-0.58 \pm 0.00$ & $-0.93 \pm 0.01$ & $-1.32 \pm 0.01$ & $-1.79 \pm 0.01$ & $-2.61 \pm 0.02$ \\
Radial fVI & $-0.28 \pm 0.00$ & $\mathbf{-0.49 \pm 0.00}$ & $\mathbf{-0.69 \pm 0.01}$ & $\mathbf{-0.90 \pm 0.01}$ & $\mathbf{-1.17 \pm 0.01}$ & $\mathbf{-1.61 \pm 0.02}$ \\
\midrule
Rank1 & $-0.33 \pm 0.00$ & $-0.71 \pm 0.01$ & $-1.06 \pm 0.01$ & $-1.46 \pm 0.01$ & $-2.02 \pm 0.02$ & $-3.00 \pm 0.03$ \\
Rank1 fVI & $\mathbf{-0.27 \pm 0.00}$ & $\mathbf{-0.47 \pm 0.00}$ & $\mathbf{-0.65 \pm 0.00}$ & $\mathbf{-0.85 \pm 0.01}$ & $\mathbf{-1.10 \pm 0.01}$ & $\mathbf{-1.53 \pm 0.01}$ \\
\midrule
Subnetwork & $-0.27 \pm 0.00$ & $-0.51 \pm 0.01$ & $-0.73 \pm 0.01$ & $-1.06 \pm 0.02$ & $-1.25 \pm 0.03$ & $-1.47 \pm 0.03$ \\
Belief Matching & $-0.26 \pm 0.00$ & $-0.51 \pm 0.00$ & $-0.73 \pm 0.01$ & $-0.97 \pm 0.01$ & $-1.26 \pm 0.01$ & $-1.70 \pm 0.02$ \\
Prior Networks & $-1.45 \pm 0.02$ & $-1.59 \pm 0.01$ & $-1.66 \pm 0.01$ & $-1.73 \pm 0.01$ & $-1.85 \pm 0.01$ & $-2.06 \pm 0.01$ \\
\bottomrule
\end{tabular}}
\end{table}

\begin{table}[h!]
\caption{Expected calibration errors for the corrupted CIFAR10 experiment. Means and standard errors over ten seeds. Best results within archetype in boldface, best results overall in blue.}
\resizebox{\textwidth}{!}{%
\begin{tabular}{ l c c c c c c }
\toprule
\multicolumn{1}{c}{CIFAR10} & \multicolumn{6}{c}{Corruption Severity} \\
\multicolumn{1}{c}{ECE $\downarrow$} & 0 & 1 & 2 & 3 & 4 & 5 \\
\midrule
MAP & $\mathbf{0.03 \pm 0.00}$ & $0.08 \pm 0.00$ & $0.12 \pm 0.00$ & $0.16 \pm 0.00$ & $0.22 \pm 0.00$ & $0.30 \pm 0.00$ \\
MAP fVI & $0.05 \pm 0.00$ & $\mathbf{0.05 \pm 0.00}$ & $\mathbf{0.05 \pm 0.00}$ & $\mathbf{0.09 \pm 0.00}$ & $\mathbf{0.14 \pm 0.00}$ & $\mathbf{0.22 \pm 0.00}$ \\
\midrule
MC Dropout & $\mathbf{0.01 \pm 0.00}$ & $\mathbf{0.03 \pm 0.00}$ & $0.06 \pm 0.00$ & $0.09 \pm 0.00$ & $0.15 \pm 0.00$ & $0.23 \pm 0.00$ \\
MC Dropout fVI & $0.06 \pm 0.00$ & $0.03 \pm 0.00$ & $\mathbf{0.03 \pm 0.00}$ & $\mathbf{0.05 \pm 0.00}$ & $\mathbf{0.09 \pm 0.00}$ & $\mathbf{0.17 \pm 0.00}$ \\
\midrule
Ensemble & {\color{blue}$\mathbf{0.01 \pm 0.00}$} & {\color{blue}$\mathbf{0.02 \pm 0.00}$} & $0.05 \pm 0.00$ & $0.08 \pm 0.00$ & $0.12 \pm 0.00$ & $0.19 \pm 0.00$ \\
Ensemble fVI & $0.07 \pm 0.00$ & $0.05 \pm 0.00$ & {\color{blue}$\mathbf{0.03 \pm 0.00}$} & {\color{blue}$\mathbf{0.04 \pm 0.00}$} & {\color{blue}$\mathbf{0.06 \pm 0.00}$} & {\color{blue}$\mathbf{0.11 \pm 0.00}$} \\
\midrule
Radial & $\mathbf{0.03 \pm 0.00}$ & $0.08 \pm 0.00$ & $0.13 \pm 0.00$ & $0.17 \pm 0.00$ & $0.23 \pm 0.00$ & $0.32 \pm 0.00$ \\
Radial fVI & $0.05 \pm 0.00$ & $\mathbf{0.05 \pm 0.00}$ & $\mathbf{0.05 \pm 0.00}$ & $\mathbf{0.09 \pm 0.00}$ & $\mathbf{0.14 \pm 0.00}$ & $\mathbf{0.23 \pm 0.00}$ \\
\midrule
Rank1 & $\mathbf{0.04 \pm 0.00}$ & $0.09 \pm 0.00$ & $0.13 \pm 0.00$ & $0.17 \pm 0.00$ & $0.23 \pm 0.00$ & $0.32 \pm 0.00$ \\
Rank1 fVI & $0.05 \pm 0.00$ & $\mathbf{0.05 \pm 0.00}$ & $\mathbf{0.04 \pm 0.00}$ & $\mathbf{0.07 \pm 0.00}$ & $\mathbf{0.12 \pm 0.00}$ & $\mathbf{0.20 \pm 0.00}$ \\
\midrule
Subnetwork & $0.01 \pm 0.00$ & $0.03 \pm 0.00$ & $0.06 \pm 0.00$ & $0.11 \pm 0.01$ & $0.13 \pm 0.01$ & $0.16 \pm 0.01$ \\
Belief Matching & $0.07 \pm 0.00$ & $0.07 \pm 0.00$ & $0.07 \pm 0.00$ & $0.09 \pm 0.00$ & $0.14 \pm 0.00$ & $0.24 \pm 0.00$ \\
Prior Networks & $0.20 \pm 0.00$ & $0.21 \pm 0.00$ & $0.21 \pm 0.00$ & $0.22 \pm 0.00$ & $0.23 \pm 0.00$ & $0.24 \pm 0.00$ \\
\bottomrule
\end{tabular}}
\end{table}

\begin{table}[h!]
\caption{Accuracies for the CIFAR10 adversarial attack experiment. Means and standard errors over ten seeds. Best results within archetype in boldface, best results overall in blue.}
\resizebox{\textwidth}{!}{%
\begin{tabular}{ l c c c c c c c }
\toprule
\multicolumn{1}{c}{CIFAR10} & \multicolumn{7}{c}{Adversarial Attack Epsilon} \\
\multicolumn{1}{c}{Accuracy $\uparrow$} & 0.00 & 0.05 & 0.10 & 0.15 & 0.20 & 0.25 & 0.30 \\
\midrule
MAP & $94.32 \pm 0.05$ & $42.35 \pm 0.23$ & $35.61 \pm 0.31$ & $31.93 \pm 0.38$ & $28.04 \pm 0.41$ & $23.60 \pm 0.40$ & $19.65 \pm 0.34$ \\
MAP fVI & $\mathbf{94.40 \pm 0.08}$ & {\color{blue}$\mathbf{70.04 \pm 0.17}$} & {\color{blue}$\mathbf{61.10 \pm 0.20}$} & $\mathbf{51.32 \pm 0.47}$ & $\mathbf{40.86 \pm 0.68}$ & $\mathbf{31.60 \pm 0.77}$ & $\mathbf{24.80 \pm 0.63}$ \\
\midrule
MC Dropout & $\mathbf{94.32 \pm 0.02}$ & $43.30 \pm 0.15$ & $32.76 \pm 0.24$ & $29.39 \pm 0.25$ & $27.08 \pm 0.31$ & $23.83 \pm 0.38$ & $19.99 \pm 0.44$ \\
MC Dropout fVI & $93.43 \pm 0.04$ & $\mathbf{56.99 \pm 0.16}$ & $\mathbf{49.91 \pm 0.29}$ & $\mathbf{43.73 \pm 0.39}$ & $\mathbf{36.97 \pm 0.57}$ & $\mathbf{29.86 \pm 0.68}$ & $\mathbf{23.63 \pm 0.68}$ \\
\midrule
Ensemble & {\color{blue}$\mathbf{95.30 \pm 0.04}$} & $43.99 \pm 0.10$ & $29.06 \pm 0.13$ & $22.55 \pm 0.13$ & $18.58 \pm 0.15$ & $15.68 \pm 0.15$ & $13.54 \pm 0.15$ \\
Ensemble fVI & $95.26 \pm 0.03$ & $\mathbf{60.12 \pm 0.10}$ & $\mathbf{49.34 \pm 0.21}$ & $\mathbf{39.38 \pm 0.36}$ & $\mathbf{30.59 \pm 0.42}$ & $\mathbf{23.73 \pm 0.43}$ & $\mathbf{18.92 \pm 0.39}$ \\
\midrule
Radial & $\mathbf{94.84 \pm 0.04}$ & $27.92 \pm 0.17$ & $18.50 \pm 0.18$ & $14.98 \pm 0.22$ & $12.69 \pm 0.25$ & $11.28 \pm 0.22$ & $10.47 \pm 0.20$ \\
Radial fVI & $93.72 \pm 0.03$ & $\mathbf{67.33 \pm 0.13}$ & $\mathbf{59.92 \pm 0.25}$ & {\color{blue}$\mathbf{52.07 \pm 0.40}$} & {\color{blue}$\mathbf{43.46 \pm 0.60}$} & {\color{blue}$\mathbf{35.32 \pm 0.64}$} & {\color{blue}$\mathbf{28.44 \pm 0.66}$} \\
\midrule
Rank1 & $93.55 \pm 0.05$ & $19.65 \pm 0.21$ & $9.66 \pm 0.18$ & $8.34 \pm 0.16$ & $8.47 \pm 0.17$ & $8.77 \pm 0.20$ & $9.11 \pm 0.19$ \\
Rank1 fVI & $\mathbf{93.86 \pm 0.04}$ & $\mathbf{67.89 \pm 0.14}$ & $\mathbf{58.99 \pm 0.12}$ & $\mathbf{49.88 \pm 0.24}$ & $\mathbf{40.48 \pm 0.36}$ & $\mathbf{31.82 \pm 0.51}$ & $\mathbf{24.89 \pm 0.57}$ \\
\bottomrule
\end{tabular}}
\end{table}

\begin{table}[h!]
\caption{Log-likelihoods for the CIFAR10 adversarial attack experiment. Means and standard errors over ten seeds. Best results within archetype in boldface, best results overall in blue.}
\resizebox{\textwidth}{!}{%
\begin{tabular}{ l c c c c c c c }
\toprule
\multicolumn{1}{c}{CIFAR10} & \multicolumn{7}{c}{Adversarial Attack Epsilon} \\
\multicolumn{1}{c}{Log-Likelihood $\uparrow$} & 0.00 & 0.05 & 0.10 & 0.15 & 0.20 & 0.25 & 0.30 \\
\midrule
MAP & $\mathbf{-0.22 \pm 0.00}$ & $-4.09 \pm 0.01$ & $-4.58 \pm 0.02$ & $-4.71 \pm 0.03$ & $-4.87 \pm 0.04$ & $-5.16 \pm 0.05$ & $-5.53 \pm 0.06$ \\
MAP fVI & $-0.25 \pm 0.00$ & {\color{blue}$\mathbf{-1.36 \pm 0.01}$} & {\color{blue}$\mathbf{-1.66 \pm 0.01}$} & {\color{blue}$\mathbf{-1.97 \pm 0.02}$} & $\mathbf{-2.34 \pm 0.03}$ & $\mathbf{-2.71 \pm 0.04}$ & $\mathbf{-3.01 \pm 0.04}$ \\
\midrule
MC Dropout & $\mathbf{-0.17 \pm 0.00}$ & $-3.26 \pm 0.01$ & $-4.42 \pm 0.02$ & $-4.61 \pm 0.02$ & $-4.60 \pm 0.02$ & $-4.69 \pm 0.05$ & $-4.96 \pm 0.07$ \\
MC Dropout fVI & $-0.25 \pm 0.00$ & $\mathbf{-1.82 \pm 0.01}$ & $\mathbf{-2.19 \pm 0.01}$ & $\mathbf{-2.37 \pm 0.02}$ & $\mathbf{-2.52 \pm 0.02}$ & $\mathbf{-2.70 \pm 0.03}$ & $\mathbf{-2.90 \pm 0.04}$ \\
\midrule
Ensemble & {\color{blue}$\mathbf{-0.15 \pm 0.00}$} & $-2.58 \pm 0.01$ & $-3.61 \pm 0.01$ & $-3.91 \pm 0.01$ & $-4.07 \pm 0.01$ & $-4.21 \pm 0.02$ & $-4.32 \pm 0.03$ \\
Ensemble fVI & $-0.21 \pm 0.00$ & $\mathbf{-1.49 \pm 0.00}$ & $\mathbf{-1.98 \pm 0.01}$ & $\mathbf{-2.30 \pm 0.01}$ & $\mathbf{-2.59 \pm 0.01}$ & $\mathbf{-2.82 \pm 0.02}$ & $\mathbf{-3.00 \pm 0.02}$ \\
\midrule
Radial & $\mathbf{-0.21 \pm 0.00}$ & $-5.52 \pm 0.01$ & $-6.49 \pm 0.02$ & $-6.78 \pm 0.04$ & $-7.01 \pm 0.05$ & $-7.30 \pm 0.07$ & $-7.56 \pm 0.09$ \\
Radial fVI & $-0.28 \pm 0.00$ & $\mathbf{-1.50 \pm 0.01}$ & $\mathbf{-1.76 \pm 0.01}$ & $\mathbf{-2.01 \pm 0.01}$ & {\color{blue}$\mathbf{-2.29 \pm 0.02}$} & {\color{blue}$\mathbf{-2.57 \pm 0.03}$} & {\color{blue}$\mathbf{-2.83 \pm 0.03}$} \\
\midrule
Rank1 & $-0.33 \pm 0.00$ & $-7.84 \pm 0.02$ & $-9.30 \pm 0.03$ & $-9.39 \pm 0.03$ & $-9.23 \pm 0.04$ & $-9.08 \pm 0.06$ & $-9.00 \pm 0.07$ \\
Rank1 fVI & $\mathbf{-0.27 \pm 0.00}$ & $\mathbf{-1.46 \pm 0.01}$ & $\mathbf{-1.75 \pm 0.00}$ & $\mathbf{-2.04 \pm 0.01}$ & $\mathbf{-2.34 \pm 0.02}$ & $\mathbf{-2.66 \pm 0.03}$ & $\mathbf{-2.93 \pm 0.04}$ \\
\bottomrule
\end{tabular}}
\end{table}

\begin{table}[h!]
\caption{Expected calibration errors for the CIFAR10 adversarial attack experiment. Means and standard errors over ten seeds. Best results within archetype in boldface, best results overall in blue.}
\resizebox{\textwidth}{!}{%
\begin{tabular}{ l c c c c c c c }
\toprule
\multicolumn{1}{c}{CIFAR10} & \multicolumn{7}{c}{Adversarial Attack Epsilon} \\
\multicolumn{1}{c}{ECE $\downarrow$} & 0.00 & 0.05 & 0.10 & 0.15 & 0.20 & 0.25 & 0.30 \\
\midrule
MAP & $\mathbf{0.03 \pm 0.00}$ & $0.48 \pm 0.00$ & $0.53 \pm 0.00$ & $0.55 \pm 0.00$ & $0.58 \pm 0.00$ & $0.61 \pm 0.00$ & $0.65 \pm 0.01$ \\
MAP fVI & $0.05 \pm 0.00$ & {\color{blue}$\mathbf{0.19 \pm 0.00}$} & {\color{blue}$\mathbf{0.24 \pm 0.00}$} & {\color{blue}$\mathbf{0.30 \pm 0.00}$} & $\mathbf{0.38 \pm 0.01}$ & $\mathbf{0.45 \pm 0.01}$ & $\mathbf{0.52 \pm 0.01}$ \\
\midrule
MC Dropout & $\mathbf{0.01 \pm 0.00}$ & $0.43 \pm 0.00$ & $0.53 \pm 0.00$ & $0.53 \pm 0.00$ & $0.52 \pm 0.00$ & $0.52 \pm 0.00$ & $0.54 \pm 0.01$ \\
MC Dropout fVI & $0.06 \pm 0.00$ & $\mathbf{0.28 \pm 0.00}$ & $\mathbf{0.32 \pm 0.00}$ & $\mathbf{0.34 \pm 0.00}$ & {\color{blue}$\mathbf{0.35 \pm 0.00}$} & {\color{blue}$\mathbf{0.38 \pm 0.01}$} & {\color{blue}$\mathbf{0.41 \pm 0.01}$} \\
\midrule
Ensemble & {\color{blue}$\mathbf{0.01 \pm 0.00}$} & $0.39 \pm 0.00$ & $0.51 \pm 0.00$ & $0.54 \pm 0.00$ & $0.56 \pm 0.00$ & $0.57 \pm 0.00$ & $0.58 \pm 0.00$ \\
Ensemble fVI & $0.07 \pm 0.00$ & $\mathbf{0.20 \pm 0.00}$ & $\mathbf{0.27 \pm 0.00}$ & $\mathbf{0.32 \pm 0.00}$ & $\mathbf{0.37 \pm 0.00}$ & $\mathbf{0.41 \pm 0.00}$ & $\mathbf{0.45 \pm 0.00}$ \\
\midrule
Radial & $\mathbf{0.03 \pm 0.00}$ & $0.62 \pm 0.00$ & $0.70 \pm 0.00$ & $0.71 \pm 0.00$ & $0.73 \pm 0.00$ & $0.74 \pm 0.00$ & $0.75 \pm 0.01$ \\
Radial fVI & $0.05 \pm 0.00$ & $\mathbf{0.21 \pm 0.00}$ & $\mathbf{0.26 \pm 0.00}$ & $\mathbf{0.30 \pm 0.00}$ & $\mathbf{0.36 \pm 0.00}$ & $\mathbf{0.42 \pm 0.01}$ & $\mathbf{0.47 \pm 0.01}$ \\
\midrule
Rank1 & $\mathbf{0.04 \pm 0.00}$ & $0.76 \pm 0.00$ & $0.86 \pm 0.00$ & $0.86 \pm 0.00$ & $0.83 \pm 0.00$ & $0.81 \pm 0.00$ & $0.80 \pm 0.00$ \\
Rank1 fVI & $0.05 \pm 0.00$ & $\mathbf{0.20 \pm 0.00}$ & $\mathbf{0.25 \pm 0.00}$ & $\mathbf{0.30 \pm 0.00}$ & $\mathbf{0.36 \pm 0.00}$ & $\mathbf{0.43 \pm 0.01}$ & $\mathbf{0.49 \pm 0.01}$ \\
\bottomrule
\end{tabular}}
\end{table}

\begin{table}[h!]
\caption{Accuracies for the corrupted CIFAR100 experiment. Means and standard errors over ten seeds. Best results within archetype in boldface, best results overall in blue.}
\resizebox{\textwidth}{!}{%
\begin{tabular}{ l c c c c c c }
\toprule
\multicolumn{1}{c}{CIFAR100} & \multicolumn{6}{c}{Corruption Severity} \\
\multicolumn{1}{c}{Accuracy $\uparrow$} & 0 & 1 & 2 & 3 & 4 & 5 \\
\midrule
MAP & $\mathbf{75.68 \pm 0.07}$ & $64.17 \pm 0.06$ & $55.41 \pm 0.07$ & $49.78 \pm 0.07$ & $43.11 \pm 0.08$ & $33.03 \pm 0.08$ \\
MAP fVI & $74.77 \pm 0.09$ & $\mathbf{64.26 \pm 0.07}$ & $\mathbf{55.82 \pm 0.09}$ & $\mathbf{50.31 \pm 0.11}$ & $\mathbf{43.56 \pm 0.12}$ & $\mathbf{33.81 \pm 0.11}$ \\
\midrule
MC Dropout & $\mathbf{74.15 \pm 0.07}$ & $\mathbf{63.23 \pm 0.06}$ & $\mathbf{54.04 \pm 0.09}$ & $\mathbf{48.33 \pm 0.08}$ & $\mathbf{41.63 \pm 0.08}$ & $\mathbf{32.02 \pm 0.09}$ \\
MC Dropout fVI & $71.53 \pm 0.12$ & $61.03 \pm 0.10$ & $51.94 \pm 0.11$ & $46.46 \pm 0.10$ & $39.88 \pm 0.09$ & $30.87 \pm 0.10$ \\
\midrule
Ensemble & {\color{blue}$\mathbf{79.13 \pm 0.05}$} & {\color{blue}$\mathbf{68.00 \pm 0.05}$} & {\color{blue}$\mathbf{59.19 \pm 0.06}$} & {\color{blue}$\mathbf{53.42 \pm 0.07}$} & {\color{blue}$\mathbf{46.45 \pm 0.06}$} & $\mathbf{35.69 \pm 0.06}$ \\
Ensemble fVI & $75.89 \pm 0.06$ & $66.38 \pm 0.07$ & $57.97 \pm 0.09$ & $52.23 \pm 0.09$ & $45.14 \pm 0.09$ & $35.19 \pm 0.11$ \\
\midrule
Radial & $\mathbf{76.40 \pm 0.08}$ & $63.76 \pm 0.07$ & $54.68 \pm 0.06$ & $49.02 \pm 0.05$ & $42.29 \pm 0.07$ & $31.89 \pm 0.07$ \\
Radial fVI & $75.29 \pm 0.10$ & $\mathbf{64.84 \pm 0.11}$ & $\mathbf{56.49 \pm 0.12}$ & $\mathbf{50.96 \pm 0.11}$ & $\mathbf{44.22 \pm 0.10}$ & $\mathbf{34.43 \pm 0.09}$ \\
\midrule
Rank1 & $73.68 \pm 0.10$ & $63.48 \pm 0.09$ & $55.34 \pm 0.12$ & $49.92 \pm 0.11$ & $43.45 \pm 0.10$ & $33.87 \pm 0.11$ \\
Rank1 fVI & $\mathbf{75.56 \pm 0.10}$ & $\mathbf{65.49 \pm 0.08}$ & $\mathbf{57.55 \pm 0.11}$ & $\mathbf{52.24 \pm 0.10}$ & $\mathbf{45.63 \pm 0.11}$ & {\color{blue}$\mathbf{35.73 \pm 0.11}$} \\
\bottomrule
\end{tabular}}
\end{table}

\begin{table}[h!]
\caption{Log-likelihoods for the corrupted CIFAR100 experiment. Means and standard errors over ten seeds. Best results within archetype in boldface, best results overall in blue.}
\resizebox{\textwidth}{!}{%
\begin{tabular}{ l c c c c c c }
\toprule
\multicolumn{1}{c}{CIFAR100} & \multicolumn{6}{c}{Corruption Severity} \\
\multicolumn{1}{c}{Log-Likelihood $\uparrow$} & 0 & 1 & 2 & 3 & 4 & 5 \\
\midrule
MAP & $\mathbf{-1.00 \pm 0.00}$ & $\mathbf{-1.59 \pm 0.00}$ & $-2.09 \pm 0.01$ & $-2.48 \pm 0.01$ & $-2.99 \pm 0.01$ & $-3.73 \pm 0.01$ \\
MAP fVI & $-1.20 \pm 0.00$ & $-1.69 \pm 0.00$ & $\mathbf{-2.08 \pm 0.01}$ & $\mathbf{-2.35 \pm 0.01}$ & $\mathbf{-2.69 \pm 0.01}$ & $\mathbf{-3.19 \pm 0.01}$ \\
\midrule
MC Dropout & $\mathbf{-0.97 \pm 0.00}$ & $\mathbf{-1.51 \pm 0.00}$ & $\mathbf{-2.02 \pm 0.01}$ & $-2.42 \pm 0.01$ & $-2.96 \pm 0.01$ & $-3.76 \pm 0.02$ \\
MC Dropout fVI & $-1.17 \pm 0.00$ & $-1.65 \pm 0.01$ & $-2.09 \pm 0.01$ & $\mathbf{-2.39 \pm 0.01}$ & $\mathbf{-2.79 \pm 0.01}$ & $\mathbf{-3.35 \pm 0.01}$ \\
\midrule
Ensemble & {\color{blue}$\mathbf{-0.81 \pm 0.00}$} & {\color{blue}$\mathbf{-1.30 \pm 0.00}$} & {\color{blue}$\mathbf{-1.70 \pm 0.00}$} & {\color{blue}$\mathbf{-1.98 \pm 0.00}$} & {\color{blue}$\mathbf{-2.37 \pm 0.01}$} & {\color{blue}$\mathbf{-2.93 \pm 0.01}$} \\
Ensemble fVI & $-1.18 \pm 0.00$ & $-1.61 \pm 0.00$ & $-1.98 \pm 0.00$ & $-2.24 \pm 0.00$ & $-2.58 \pm 0.00$ & $-3.06 \pm 0.01$ \\
\midrule
Radial & $\mathbf{-0.98 \pm 0.00}$ & $\mathbf{-1.66 \pm 0.01}$ & $-2.21 \pm 0.01$ & $-2.65 \pm 0.02$ & $-3.21 \pm 0.02$ & $-4.09 \pm 0.03$ \\
Radial fVI & $-1.21 \pm 0.00$ & $-1.69 \pm 0.00$ & $\mathbf{-2.08 \pm 0.01}$ & $\mathbf{-2.34 \pm 0.01}$ & $\mathbf{-2.67 \pm 0.01}$ & $\mathbf{-3.16 \pm 0.00}$ \\
\midrule
Rank1 & $-1.48 \pm 0.01$ & $-2.29 \pm 0.01$ & $-3.01 \pm 0.01$ & $-3.59 \pm 0.02$ & $-4.38 \pm 0.02$ & $-5.57 \pm 0.02$ \\
Rank1 fVI & $\mathbf{-1.17 \pm 0.00}$ & $\mathbf{-1.64 \pm 0.00}$ & $\mathbf{-2.00 \pm 0.01}$ & $\mathbf{-2.25 \pm 0.01}$ & $\mathbf{-2.58 \pm 0.01}$ & $\mathbf{-3.06 \pm 0.01}$ \\
\bottomrule
\end{tabular}}
\end{table}

\begin{table}[h!]
\caption{Expected calibration errors for the corrupted CIFAR100 experiment. Means and standard errors over ten seeds. Best results within archetype in boldface, best results overall in blue.}
\resizebox{\textwidth}{!}{%
\begin{tabular}{ l c c c c c c }
\toprule
\multicolumn{1}{c}{CIFAR100} & \multicolumn{6}{c}{Corruption Severity} \\
\multicolumn{1}{c}{ECE $\downarrow$} & 0 & 1 & 2 & 3 & 4 & 5 \\
\midrule
MAP & $\mathbf{0.08 \pm 0.00}$ & $0.12 \pm 0.00$ & $0.16 \pm 0.00$ & $0.20 \pm 0.00$ & $0.23 \pm 0.00$ & $0.30 \pm 0.00$ \\
MAP fVI & $0.12 \pm 0.00$ & $\mathbf{0.11 \pm 0.00}$ & $\mathbf{0.09 \pm 0.00}$ & $\mathbf{0.06 \pm 0.00}$ & {\color{blue}$\mathbf{0.04 \pm 0.00}$} & $\mathbf{0.04 \pm 0.00}$ \\
\midrule
MC Dropout & {\color{blue}$\mathbf{0.02 \pm 0.00}$} & $0.06 \pm 0.00$ & $0.10 \pm 0.00$ & $0.13 \pm 0.00$ & $0.17 \pm 0.00$ & $0.24 \pm 0.00$ \\
MC Dropout fVI & $0.08 \pm 0.00$ & $\mathbf{0.06 \pm 0.00}$ & {\color{blue}$\mathbf{0.02 \pm 0.00}$} & {\color{blue}$\mathbf{0.01 \pm 0.00}$} & $\mathbf{0.05 \pm 0.00}$ & $\mathbf{0.10 \pm 0.00}$ \\
\midrule
Ensemble & $\mathbf{0.05 \pm 0.00}$ & {\color{blue}$\mathbf{0.04 \pm 0.00}$} & $\mathbf{0.02 \pm 0.00}$ & $\mathbf{0.02 \pm 0.00}$ & $\mathbf{0.04 \pm 0.00}$ & $0.10 \pm 0.00$ \\
Ensemble fVI & $0.18 \pm 0.00$ & $0.18 \pm 0.00$ & $0.16 \pm 0.00$ & $0.14 \pm 0.00$ & $0.10 \pm 0.00$ & $\mathbf{0.04 \pm 0.00}$ \\
\midrule
Radial & $\mathbf{0.09 \pm 0.00}$ & $0.14 \pm 0.00$ & $0.19 \pm 0.00$ & $0.23 \pm 0.00$ & $0.27 \pm 0.00$ & $0.34 \pm 0.00$ \\
Radial fVI & $0.13 \pm 0.00$ & $\mathbf{0.12 \pm 0.00}$ & $\mathbf{0.10 \pm 0.00}$ & $\mathbf{0.08 \pm 0.00}$ & $\mathbf{0.05 \pm 0.00}$ & $\mathbf{0.04 \pm 0.00}$ \\
\midrule
Rank1 & $0.16 \pm 0.00$ & $0.22 \pm 0.00$ & $0.28 \pm 0.00$ & $0.32 \pm 0.00$ & $0.37 \pm 0.00$ & $0.44 \pm 0.00$ \\
Rank1 fVI & $\mathbf{0.13 \pm 0.00}$ & $\mathbf{0.13 \pm 0.00}$ & $\mathbf{0.11 \pm 0.00}$ & $\mathbf{0.09 \pm 0.00}$ & $\mathbf{0.06 \pm 0.00}$ & {\color{blue}$\mathbf{0.03 \pm 0.00}$} \\
\bottomrule
\end{tabular}}
\end{table}

\begin{table}[h!]
\caption{Accuracies for the CIFAR100 adversarial attack experiment. Means and standard errors over ten seeds. Best results within archetype in boldface, best results overall in blue.}
\resizebox{\textwidth}{!}{%
\begin{tabular}{ l c c c c c c c }
\toprule
\multicolumn{1}{c}{CIFAR100} & \multicolumn{7}{c}{Adversarial Attack Epsilon} \\
\multicolumn{1}{c}{Accuracy $\uparrow$} & 0.00 & 0.05 & 0.10 & 0.15 & 0.20 & 0.25 & 0.30 \\
\midrule
MAP & $\mathbf{75.68 \pm 0.07}$ & $11.84 \pm 0.13$ & $6.70 \pm 0.08$ & $4.98 \pm 0.08$ & $4.18 \pm 0.07$ & $3.62 \pm 0.09$ & $3.18 \pm 0.09$ \\
MAP fVI & $74.77 \pm 0.09$ & $\mathbf{15.83 \pm 0.11}$ & $\mathbf{9.79 \pm 0.12}$ & $\mathbf{7.08 \pm 0.09}$ & $\mathbf{5.38 \pm 0.05}$ & $\mathbf{4.30 \pm 0.08}$ & $\mathbf{3.46 \pm 0.07}$ \\
\midrule
MC Dropout & $\mathbf{74.05 \pm 0.08}$ & $19.55 \pm 0.10$ & $10.56 \pm 0.08$ & $7.45 \pm 0.09$ & $5.90 \pm 0.09$ & $4.91 \pm 0.06$ & $4.01 \pm 0.07$ \\
MC Dropout fVI & $71.61 \pm 0.10$ & $\mathbf{21.60 \pm 0.07}$ & $\mathbf{13.22 \pm 0.09}$ & $\mathbf{9.52 \pm 0.09}$ & {\color{blue}$\mathbf{7.19 \pm 0.12}$} & {\color{blue}$\mathbf{5.62 \pm 0.11}$} & {\color{blue}$\mathbf{4.49 \pm 0.11}$} \\
\midrule
Ensemble & {\color{blue}$\mathbf{79.13 \pm 0.05}$} & $26.04 \pm 0.05$ & $13.51 \pm 0.09$ & $8.64 \pm 0.07$ & $6.32 \pm 0.06$ & $\mathbf{4.92 \pm 0.07}$ & $\mathbf{3.97 \pm 0.07}$ \\
Ensemble fVI & $75.89 \pm 0.06$ & {\color{blue}$\mathbf{29.32 \pm 0.06}$} & {\color{blue}$\mathbf{16.00 \pm 0.08}$} & {\color{blue}$\mathbf{10.09 \pm 0.07}$} & $\mathbf{6.77 \pm 0.05}$ & $4.82 \pm 0.05$ & $3.66 \pm 0.06$ \\
\midrule
Radial & $\mathbf{76.42 \pm 0.08}$ & $12.05 \pm 0.08$ & $7.36 \pm 0.06$ & $5.57 \pm 0.05$ & $4.52 \pm 0.07$ & $3.77 \pm 0.07$ & $3.19 \pm 0.08$ \\
Radial fVI & $75.29 \pm 0.10$ & $\mathbf{16.85 \pm 0.12}$ & $\mathbf{10.97 \pm 0.09}$ & $\mathbf{7.91 \pm 0.09}$ & $\mathbf{6.06 \pm 0.10}$ & $\mathbf{4.83 \pm 0.12}$ & $\mathbf{3.99 \pm 0.10}$ \\
\midrule
Rank1 & $73.76 \pm 0.07$ & $14.01 \pm 0.09$ & $9.54 \pm 0.08$ & $7.63 \pm 0.12$ & $6.18 \pm 0.12$ & $5.09 \pm 0.14$ & $4.20 \pm 0.13$ \\
Rank1 fVI & $\mathbf{75.58 \pm 0.09}$ & $\mathbf{18.34 \pm 0.08}$ & $\mathbf{11.68 \pm 0.07}$ & $\mathbf{8.65 \pm 0.10}$ & $\mathbf{6.67 \pm 0.09}$ & $\mathbf{5.29 \pm 0.08}$ & $\mathbf{4.38 \pm 0.08}$ \\
\bottomrule
\end{tabular}}
\end{table}

\begin{table}[h!]
\caption{Log-likelihoods for the CIFAR100 adversarial attack experiment. Means and standard errors over ten seeds. Best results within archetype in boldface, best results overall in blue.}
\resizebox{\textwidth}{!}{%
\begin{tabular}{ l c c c c c c c }
\toprule
\multicolumn{1}{c}{CIFAR100} & \multicolumn{7}{c}{Adversarial Attack Epsilon} \\
\multicolumn{1}{c}{Log-Likelihood $\uparrow$} & 0.00 & 0.05 & 0.10 & 0.15 & 0.20 & 0.25 & 0.30 \\
\midrule
MAP & $\mathbf{-1.00 \pm 0.00}$ & $-6.64 \pm 0.03$ & $-6.72 \pm 0.03$ & $-6.53 \pm 0.03$ & $-6.52 \pm 0.05$ & $-6.64 \pm 0.06$ & $-6.76 \pm 0.08$ \\
MAP fVI & $-1.20 \pm 0.00$ & $\mathbf{-5.36 \pm 0.01}$ & $\mathbf{-5.62 \pm 0.02}$ & $\mathbf{-5.53 \pm 0.02}$ & $\mathbf{-5.45 \pm 0.02}$ & $\mathbf{-5.40 \pm 0.02}$ & $\mathbf{-5.36 \pm 0.02}$ \\
\midrule
MC Dropout & $\mathbf{-0.97 \pm 0.00}$ & $-5.02 \pm 0.01$ & $-5.93 \pm 0.02$ & $-6.08 \pm 0.02$ & $-6.23 \pm 0.02$ & $-6.46 \pm 0.03$ & $-6.71 \pm 0.05$ \\
MC Dropout fVI & $-1.17 \pm 0.00$ & $\mathbf{-4.36 \pm 0.01}$ & $\mathbf{-5.17 \pm 0.02}$ & $\mathbf{-5.35 \pm 0.01}$ & $\mathbf{-5.45 \pm 0.01}$ & $\mathbf{-5.55 \pm 0.01}$ & $\mathbf{-5.66 \pm 0.02}$ \\
\midrule
Ensemble & {\color{blue}$\mathbf{-0.81 \pm 0.00}$} & {\color{blue}$\mathbf{-3.68 \pm 0.00}$} & {\color{blue}$\mathbf{-4.41 \pm 0.00}$} & {\color{blue}$\mathbf{-4.65 \pm 0.01}$} & {\color{blue}$\mathbf{-4.81 \pm 0.01}$} & {\color{blue}$\mathbf{-4.93 \pm 0.02}$} & {\color{blue}$\mathbf{-5.01 \pm 0.02}$} \\
Ensemble fVI & $-1.18 \pm 0.00$ & $-3.97 \pm 0.00$ & $-4.86 \pm 0.01$ & $-5.10 \pm 0.01$ & $-5.15 \pm 0.01$ & $-5.17 \pm 0.01$ & $-5.18 \pm 0.01$ \\
\midrule
Radial & $\mathbf{-0.98 \pm 0.00}$ & $-6.73 \pm 0.02$ & $-6.80 \pm 0.03$ & $-6.69 \pm 0.03$ & $-6.79 \pm 0.05$ & $-7.04 \pm 0.07$ & $-7.28 \pm 0.10$ \\
Radial fVI & $-1.21 \pm 0.00$ & $\mathbf{-5.14 \pm 0.01}$ & $\mathbf{-5.36 \pm 0.02}$ & $\mathbf{-5.29 \pm 0.02}$ & $\mathbf{-5.22 \pm 0.02}$ & $\mathbf{-5.18 \pm 0.02}$ & $\mathbf{-5.16 \pm 0.02}$ \\
\midrule
Rank1 & $-1.48 \pm 0.00$ & $-9.99 \pm 0.02$ & $-10.69 \pm 0.02$ & $-10.76 \pm 0.02$ & $-10.81 \pm 0.04$ & $-10.93 \pm 0.05$ & $-11.12 \pm 0.07$ \\
Rank1 fVI & $\mathbf{-1.17 \pm 0.00}$ & $\mathbf{-4.95 \pm 0.01}$ & $\mathbf{-5.26 \pm 0.01}$ & $\mathbf{-5.22 \pm 0.01}$ & $\mathbf{-5.17 \pm 0.01}$ & $\mathbf{-5.16 \pm 0.01}$ & $\mathbf{-5.16 \pm 0.02}$ \\
\bottomrule
\end{tabular}}
\end{table}

\begin{table}[h!]
\caption{Expected calibration errors for the CIFAR100 adversarial attack experiment. Means and standard errors over ten seeds. Best results within archetype in boldface, best results overall in blue.}
\resizebox{\textwidth}{!}{%
\begin{tabular}{ l c c c c c c c }
\toprule
\multicolumn{1}{c}{CIFAR100} & \multicolumn{7}{c}{Adversarial Attack Epsilon} \\
\multicolumn{1}{c}{ECE $\downarrow$} & 0.00 & 0.05 & 0.10 & 0.15 & 0.20 & 0.25 & 0.30 \\
\midrule
MAP & $\mathbf{0.08 \pm 0.00}$ & $0.55 \pm 0.00$ & $0.52 \pm 0.00$ & $0.49 \pm 0.00$ & $0.49 \pm 0.01$ & $0.49 \pm 0.01$ & $0.51 \pm 0.01$ \\
MAP fVI & $0.12 \pm 0.00$ & $\mathbf{0.23 \pm 0.00}$ & $\mathbf{0.23 \pm 0.00}$ & $\mathbf{0.22 \pm 0.00}$ & $\mathbf{0.23 \pm 0.00}$ & $\mathbf{0.23 \pm 0.01}$ & $\mathbf{0.23 \pm 0.01}$ \\
\midrule
MC Dropout & {\color{blue}$\mathbf{0.03 \pm 0.00}$} & $0.45 \pm 0.00$ & $0.47 \pm 0.00$ & $0.44 \pm 0.00$ & $0.43 \pm 0.00$ & $0.43 \pm 0.00$ & $0.44 \pm 0.01$ \\
MC Dropout fVI & $0.09 \pm 0.00$ & $\mathbf{0.26 \pm 0.00}$ & $\mathbf{0.29 \pm 0.00}$ & $\mathbf{0.28 \pm 0.00}$ & $\mathbf{0.28 \pm 0.00}$ & $\mathbf{0.29 \pm 0.00}$ & $\mathbf{0.30 \pm 0.00}$ \\
\midrule
Ensemble & $\mathbf{0.05 \pm 0.00}$ & $0.26 \pm 0.00$ & $0.31 \pm 0.00$ & $0.30 \pm 0.00$ & $0.30 \pm 0.00$ & $0.30 \pm 0.01$ & $0.29 \pm 0.01$ \\
Ensemble fVI & $0.18 \pm 0.00$ & {\color{blue}$\mathbf{0.08 \pm 0.00}$} & {\color{blue}$\mathbf{0.14 \pm 0.00}$} & {\color{blue}$\mathbf{0.15 \pm 0.00}$} & {\color{blue}$\mathbf{0.16 \pm 0.00}$} & {\color{blue}$\mathbf{0.17 \pm 0.00}$} & {\color{blue}$\mathbf{0.18 \pm 0.00}$} \\
\midrule
Radial & $\mathbf{0.09 \pm 0.00}$ & $0.57 \pm 0.00$ & $0.54 \pm 0.00$ & $0.51 \pm 0.00$ & $0.51 \pm 0.01$ & $0.53 \pm 0.01$ & $0.55 \pm 0.01$ \\
Radial fVI & $0.13 \pm 0.00$ & $\mathbf{0.21 \pm 0.00}$ & $\mathbf{0.21 \pm 0.00}$ & $\mathbf{0.20 \pm 0.00}$ & $\mathbf{0.20 \pm 0.00}$ & $\mathbf{0.20 \pm 0.01}$ & $\mathbf{0.19 \pm 0.01}$ \\
\midrule
Rank1 & $0.16 \pm 0.00$ & $0.76 \pm 0.00$ & $0.76 \pm 0.00$ & $0.73 \pm 0.00$ & $0.71 \pm 0.00$ & $0.71 \pm 0.00$ & $0.70 \pm 0.01$ \\
Rank1 fVI & $\mathbf{0.13 \pm 0.00}$ & $\mathbf{0.20 \pm 0.00}$ & $\mathbf{0.21 \pm 0.00}$ & $\mathbf{0.20 \pm 0.00}$ & $\mathbf{0.20 \pm 0.00}$ & $\mathbf{0.20 \pm 0.00}$ & $\mathbf{0.20 \pm 0.01}$ \\
\bottomrule
\end{tabular}}
\end{table}

\end{document}